# Causal Value Attribution of Marketing Labels in LLM-Generated Ads: A Deconfounder Approach

Danhong You
Shanghai Jiao Tong University
dan1559187326@sjtu.edu.cn

Wei Jiang
Shanghai Jiao Tong University
jiangwei@sjtu.edu.cn

Zhiqiang (Eric) Zheng*
University of Texas at Dallas
ericz@utdallas.edu

**Abstract**

As AI-generated content becomes commoditized, economic value increasingly shits from final artifacts to the upstream expertise used to specify inputs that steer LLM generation. Yet organizations lack systematic methods for valuing this expertise: although they routinely record granular skill primitives, they face the challenge of determining which individual skills or configurations causally improve performance. We ground this emerging problem in generative marketing, where structured expert labels serve as explicit skill primitives that condition LLM-generated advertisements. Identifying their causal effects is hard because outcomes are shaped by stochastic LLM mediation, high-dimensional interactions among multiple labels, and multimodal confounding. To address these challenges, we propose D-VAE (Deconfounder Variational Autoencoder), a counterfactual-inference framework that integrates a multi-treatment deconfounder with causal representation learning. D-VAE recovers a causally disentangled latent representation of expert labeling logic and provides PAC-based theoretical bounds on omitted-variable bias arising from single-treatment confounders. Applied to 29,177 foundation-makeup ad creatives, D-VAE significantly outperforms state-of-the-art benchmarks in out-of-sample prediction and counterfactual inference, enabling causal attribution of both individual labels and label combinations to user engagement. A controlled field experiment provides direct evidence of its practical value. Label configurations recommended by D-VAE increase total engagement by 32.7% relative to control configurations and achieve performance statistically indistinguishable from that of human experts. A simulation study with a known ground-truth causal structure further demonstrates D-VAE's superior ability to causally attribute value to individual skills under complex confounding and treatment interactions. Overall, this study advocates shifting the analytical focus from valuing AI-generated products to valuing the skills that produce them. By converting expert know-how into scalable, causally grounded recommendations, the framework provides a methodological foundation for pricing and optimizing AI-augmented skill portfolios.

**Keywords:** Skill Valuation, Causal Inference, Marketing Label Attribution, Deconfounder Method, Large Language Model

## 1. Introduction

Large language models (LLMs) have drastically reduced the marginal cost of digital content generation (Stanford HAI 2025). Consequently, producing final artifacts is no longer scarce; what remains scarce is the skill—the structured know-how to effectively steer, condition, and combine model capabilities for reliable, high-value outcomes. This technological shift signals a broader business model transformation: moving from valuing end products to valuing the upstream skills that produce them.

This transition has already been materialized. The global AI prompt and skill marketplace was projected to reach $1.74 billion in 2025 (Grand View Research, 2026), with organizations increasingly licensing and optimizing domain-specific instructions and workflow configurations rather than merely consuming outputs. Yet, traditional pricing mechanisms and academic valuation frameworks remain anchored to final products. To value skills effectively, we must pivot methodologically: decomposing observed product value into the attributable contributions of constituent skills.

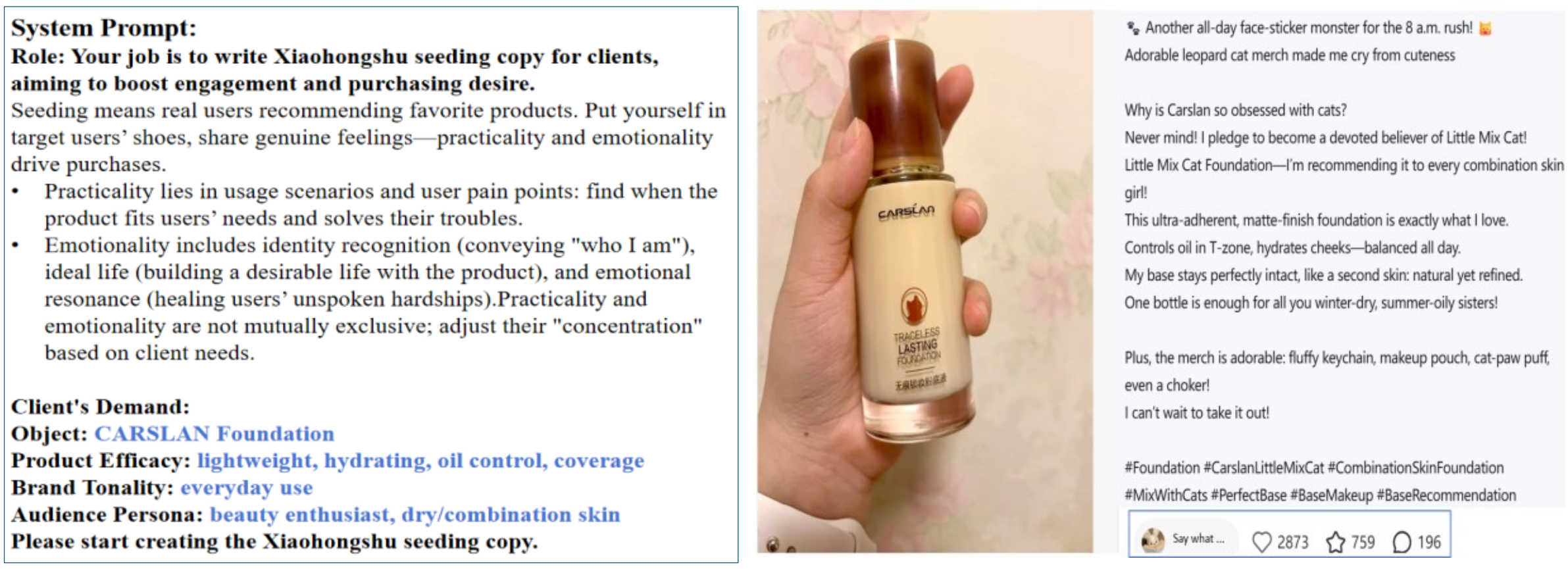

**Figure 1. Example of an ad creative (copy) and user engagement**

We ground this emerging problem in AI-augmented marketing content creation, where LLMs are increasingly embedded into daily creative workflows (Arora et al., 2025; Reisenbichler et al., 2026). A prevalent practice is "label-to-copy": marketers translate campaign intent into structured labels—compact representations of expert knowledge spanning product efficacy, brand tonality, and audience persona, which then steer LLM generation (Zhao et al., 2023; Cillo & Rubera, 2025). In our context, skill refers to the ability to select appropriate labels that guide an LLM in generating ad copies for a target audience. The selected labels serve as observable proxies — *skill primitives* — while the underlying logic governing their selection constitutes the latent skill of interest. Consider a campaign for the makeup brand Carslan Foundation (Figure 1): the creative direction is guided by a label vector $L$ specifying efficacy cues (e.g., lightweight, hydrating), tonality cues (e.g., tech-driven aesthetics), and persona cues (e.g., beauty enthusiasts). Conditioned on this configuration, the LLM generates a copy $X$ which leads to user engagement $Y$ (e.g., likes, comments, collects) that reflects the market resonance of the chosen label setup. Figure 2(a) formalizes this causal chain: observable label ($L$) → LLM-mediated product ($X$) → realized product value ($Y$). However, while individual labels are routinely archived in marketing databases, the strategic logic that binds them remains tacit and highly context dependent.

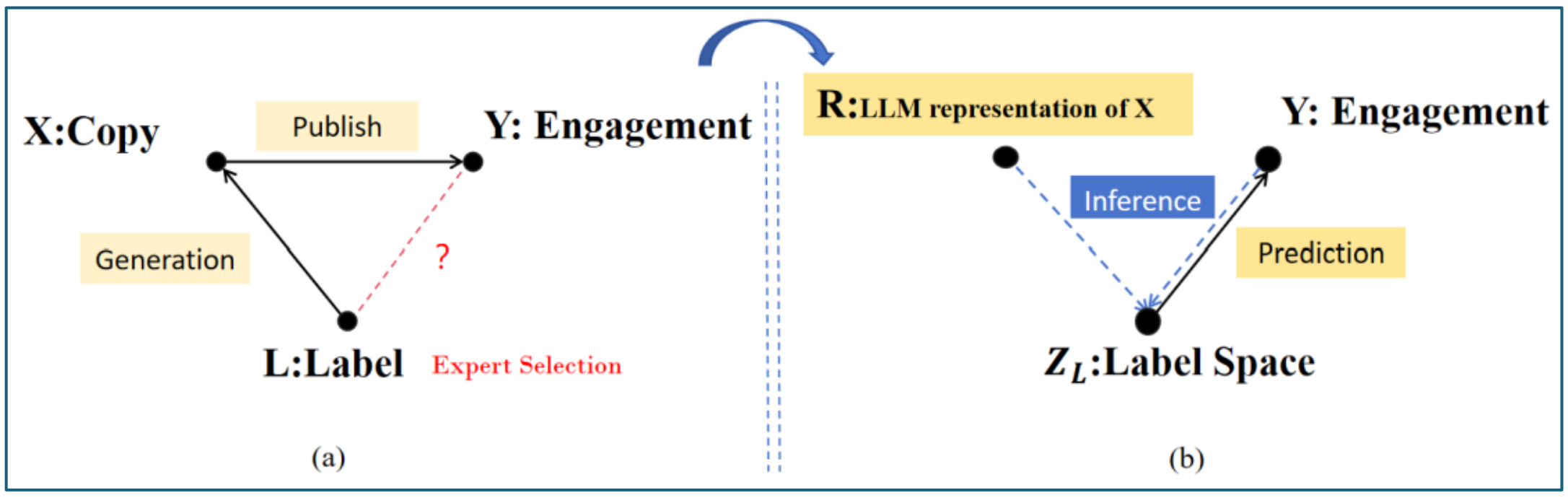

**Figure 2: Two Causal Structures** (a) the causal path of label, copy and user engagement, where copy is the mediator; (b) the latent label space inference and direct effect on user engagement.

This creates a critical valuation gap: organizations possess granular records of skill primitives (corresponding to chosen labels) but lack systematic insight into which combinatorial strategies actually drive performance. Should a student-targeting campaign prioritize “affordability” or a “co-branded IP” hook? How must the label combination adapt when shifting from everyday utility to holiday gifting? These decisions hinge on the latent skill: the ability to select and combine labels for a given target audience. This maps directly onto our core methodological quest: how do we measure the causal value of label configurations when only aggregate product outcomes (engagements) are observed? Attributing value to the underlying skill through observed label configurations requires decomposing aggregate performance into constituent contributions, a task complicated by high-dimensional label spaces, complex interactions, and the absence of ground-truth causal signals at the individual-label and label-configuration levels. In LLM-driven ad creation, this value attribution problem manifests as three identification challenges for estimating the causal effect of label configurations on engagement:

First, the label-engagement relationship is fully mediated by stochastic LLM generation. Labels do not directly drive outcomes (because customers do not see labels directly); they operate exclusively through the generated copy. But identical label sets can yield semantically aligned but stylistically divergent texts due to stochastic decoding and lexical variation (Renze & Guven, 2024). This uncontrolled mediator noise brought by LLM generation obscures the true causal contribution of the label combination.

Second, the treatment space is high-dimensional and inherently interactive, rendering experimentation economically prohibitive. Labels form sparse, discrete vectors whose effects are nonlinear and strongly dependent on combinatorial configurations (e.g., efficacy × tonality × persona). The resulting combinatorial dimensionality explosion makes exhaustive A/B or factorial testing operationally infeasible.

Third, latent and multimodal confounding systematically bias observational identification. Latent confounds are persistent and heterogeneous in scope: some affect multiple labels simultaneously (multi-treatment confounds), while others affect only a single label (single-treatment confounds). Real-world ads integrate text and imagery that jointly drive engagement, and these visual features are frequently co-determined with label choices (e.g., "tech-driven" positioning systematically paired with minimalist visuals). Failure to account for these latent and cross-modal dependencies creates structured confounding that masks the attributed value of individual skills.

These challenges motivate D-VAE (Deconfounder Variational Autoencoder), a counterfactual inference framework designed to estimate the causal value of expert-selected labels in LLM-mediated marketing. The central idea is to shift the object of intervention from realized copies to a latent skill space that faithfully represents the expert logic behind label selection. This shift is necessary because realized copies contain stochastic narrative and stylistic variation introduced by the LLM, while raw label vectors are sparse, high-dimensional, and confounded by unobserved campaign and platform conditions. D-VAE therefore learns an intervention-ready representation that preserves label-driven semantic intent (Yang et al., 2021), separates it from mediator-side noise, and adjusts it for observed and latent confounding. Specifically, multimodal LLM representations are used to capture content-related and visual confounders (Imai & Nakamura, 2025), the multi-label assignment structure is used to infer shared latent confounders (Wang & Blei, 2019), and the single label-specific confounders is theoretically bounded (Oster, 2019). Counterfactual label interventions are then performed in the deconfounded latent space, enabling causal inference under alternative label configurations (Figure 2b).

Empirical analyses of 29,177 paid ad creatives show that D-VAE learns high-fidelity label representations ($MSE_X$ = 0.1103; *mean squared error for ad copy reconstruction*), accurately predicts observed engagement outcomes ($MSE_Y$ = 0.0330; *mean squared error for engagement prediction* ), and produces causally coherent counterfactual estimates after adjusting for confounding, achieving an Interchange Intervention Accuracy (*IIA; percentage of latent interventions that reproduce the outcome of corresponding label interventions*) score of 96.7%. A supplementary controlled field experiment provides direct evidence of practical value: label configurations optimized by D-VAE increased engagement by 32.7% relative to the control configurations ($p$ = 0.0053) and achieved performance statistically indistinguishable from that of human experts.

Finally, we conduct a simulation study to evaluate the method more systematically and examine its generalizability. Synthetic data allow us to specify the ground-truth causal effects as well as the magnitude and structure of latent confounding. The simulation captures a general and increasingly common setting in which skills $L$ steers an LLM to generate a high-dimensional, unstructured artifact $X$, which subsequently affects a downstream outcome $Y$. We instantiate this setting in a labor market context, where job candidates select keywords to guide LLM in generating self-descriptions intended to attract recruiters' attentions. The results comprehensively demonstrate D-VAE's superior causal-attribution performance and identify important boundary conditions for its effective application.

Our contributions are threefold:

Theoretically, we formalize a skill-to-product causal value attribution framework that shifts the analytical lens from valuing final outputs to valuing the upstream skills that generate them. We model the causal pathway: Skill → LLM-mediated product → downstream value (e.g., L → X → Y). This reflects the operational reality of AI-augmented workflows: practitioners no longer control the final output directly; instead, they intervene at the skill level by designing, selecting, and combining input specifications to steer LLM generation. While prior literature predominantly estimates the effect of realized content on outcomes (X→Y) (Xiao et al., 2024; Kapoor & Kumar, 2025), we ground valuation at the actionable skill-intervention stage. By treating the marketing label-to-copy workflow as a canonical instantiation, we provide a generalizable structure for decomposing product-level value into constituent skill contributions.

Methodologically, we develop a counterfactual representation learning framework designed to identify and disentangle label contributions in high-dimensional generative workflows. Standard causal inference tools often struggle in high-dimensional, unstructured settings, while representation learning methods can yield biased effects when confounding is not addressed. Our approach learns a latent causal label space and adjusts for both observed and partially identifiable confounders, enabling more reliable counterfactual engagement prediction.

Practically, we translate skill valuation into actionable portfolio optimization and pricing insights for AI-augmented organizations. In complex generative workflows, multiple label components (e.g., efficacy cues, tonality frames, audience targeting) are combined, making it challenging for standard predictive models to disentangle causal drivers from spurious correlations. Our counterfactual attribution framework quantifies the causal value of labels, revealing which configurations genuinely drive performance. Applied to the marketing domain, this enables firms to audit, price, and optimize their skill inventories—allocating resources toward high-ROI skill combinations, retiring redundant or ineffective ones.

## 2. Literature Review

Our study intersects three related streams: GenAI in marketing content generation, deconfounder method and causal representation learning.

### 2.1 AI in Marketing: Content Generation, Value Attribution and User Engagement

Generative AI powered by LLMs has been deployed widely in marketing applications, including automated copy generation (Liu et al., 2025; Reisenbichler et al., 2026), ad creative design (Chen et al., 2024; Kapoor & Kumar, 2025), traffic allocation (Ye et al., 2025), and live-streaming commerce (Wang et al., 2025). Empirical evidence suggests that LLMs reduce content production costs while increasing productivity, and human–AI collaboration can outperform either human or AI alone (Boussioux et al,2024; Hartmann et al., 2025). Yet, effective use is uneven: experts can craft more precise prompts aligned with brand strategy, whereas novices often rely on inefficient trial-and-error (Chen et al., 2024).

Recent advances have explored agentic and skill-based workflows for LLM-driven marketing content generation. These approaches typically decompose marketing expertise into modular “skills” or specialized agent roles (e.g., strategist, designer, copywriter) within multi-agent systems (Wang et al., 2025; Quan et al., 2025). This line of work showcases a move from holistic evaluation of LLM outputs toward component-level assessment of agent skills. Some research decomposes LLM-based workflows into stages such as planning, tool selection, tool calling, and response generation stages, proposing benchmarks such as CToolEval and AgentQuest to evaluate agents in modular settings (Guo et al., 2024; Gioacchini et al., 2024). Close to our objetives, CapaBench uses Shapley values to quantify the marginal contribution and interaction effects of modules such as planning, reasoning, action execution, and reflection in LLM agents (Yang et al., 2025). Related attribution work in RAG and model interpretability further traces generated answers to retrieved documents or model predictions to influential training examples (Qi et al., 2024; Wang et al., 2024). However, these studies focus on technical modules, retrieved evidence, or training data, leaving unclear the causal value of skills, individually or in combination, in shaping LLM-generated content and downstream engagement, when confounds are present but unobserved.

Alongside GenAI, a large literature uses machine learning to predict marketing outcome and personalize delivery. Typical approaches build user profiles from their historical behavior (Trusov et al., 2016; Märtin et al., 2023) and train models on clicks and conversions (Rafieian & Yoganarasimhan, 2021; Chen et al., 2024) to optimize content allocation (Besbes et al., 2016), often improving personalization accuracy (Yoganarasimhan, 2020). However, these methods do not aim to determine value attribution of content creation causally. For example, higher engagement for creatives with a co-branded IP label may reflect the label’s effect or simply the fact that such labels are more common in high-budget, high-quality campaigns. Complementary work examines user psychology (Deng et al., 2022; Sun et al., 2025) or content features (Xiao et al., 2024; Ellickson et al., 2023) as possible drivers to engagement, but these are typically downstream insights that provide limited guidance during upstream content creation.

Methodologically grounded causal studies of LLM-driven marketing remain limited. Existing work commonly (i) treats LLM outputs as static inputs for engagement prediction via regressions or black-box neural models (Chen & Chan, 2024; Kapoor & Kumar, 2025), leaving decoding stochasticity unaddressed; (ii) estimates the average effect of AI assistance using experiments (Ye et al., 2025; Reisenbichler et al., 2026), but does not identify the effects of specific prompt components or label combinations; or (iii) studies human–AI workflows using econometrics at the workflow level but not at the level of combinatorial label inputs (Wang et al., 2025; Lou & Xie, 2021).

In advertising practice, experts often operationalize marketing knowledge into structured labels that serve as prompts for LLMs. We therefore study labels not merely as observable content features, but as upstream interventions in the content-generation process, and aim to identify their indirect causal effects on engagement as well as the marginal and combinatorial value of individual labels.

### 2.2 Deconfounder Methods

Confounding is particularly severe in GenAI-driven marketing because it arises from both observable multimodal factors (e.g., text–image pairing) and latent drivers (e.g., campaign budget, platform dynamics). These factors jointly shape label choice and engagement but are rarely fully measured, leaving causal identification confounded.

Existing methods have only partially addressed this problem. Classical proxy-based approaches require strong and often unverifiable links between observed proxies and latent confounders (Frost 1979). Matching and representation-balancing approaches (e.g., propensity score matching, TARNet) can reduce imbalance in measured covariates, but they cannot recover confounding information absent from the data (Caliendo & Kopeinig 2008; Shalit et al., 2017). ML-based causal estimators such as Meta-Learners (Künzel et al., 2019) and Double Machine Learning (Chernozhukov et al., 2018) flexibly adjust for observed covariates, but their validity still rests on selection on observables, not on unobserved confounds. This is a critical limitation in our setting, where label choices and engagement are shaped by external semantic factors, campaign conditions, and market context that are not directly observed — and LLM-mediated generation compounds the issue by routing the label–engagement relationship through an additional unobserved process: how the LLM produces the ad copy.

The use of an LLM, however, also creates a new opportunity. Unlike traditional ML estimators (e.g., double machine learning) that are trained solely on the study data, pretrained LLMs have already encoded external semantic associations from their large training corpus. These associations are unobserved in the study data but shape both how labels are transformed into ad copy and how users respond to the generated content. In effect, a powerful LLM has *already absorbed* some, if not all, of the confounding signals during generation. Recent GenAI-Powered Inference (GPI) builds on this idea, recovering confounding information from LLM representations that encode semantic knowledge beyond observed variables (Imai & Nakamura 2025).

We adopt this logic but adapt it to a different causal structure: labels, generated content, and engagement are observed, while the external semantic factors implicitly reflected in the LLM's generation process are not. We therefore learn a latent representation from this process and exploit the multi-treatment label structure to separate shared confounding from the marginal and combinatorial value of individual labels. This contrasts with the Deconfounder framework of Wang and Blei (2019), which infers shared latent confounders primarily from treatment co-occurrence patterns; our setting demands recovering confounding signals embedded in both combinatorial label assignment and LLM-mediated semantic transformation.

**2.3 Causal Representation Learning**

Causal representation learning embeds causal structure into latent variables to enable intervention and "what-if" reasoning. In LLM applications, prior work has integrated Structural Causal Models (SCM) with generative models for counterfactual generation, including SCM-VAE (Komanduri et al., 2022) and DEAR (Shen et al., 2022), while diffusion-based approaches such as DiffPO (Ma et al., 2024) model potential outcome distributions to estimate treatment effects. Among these, C-VAE (Yang et al., 2021) is especially influential and relevant: unlike standard VAEs that encourage statistical independence among latent factors (Kingma & Welling, 2013), it imposes causal dependencies through a weakly supervised conditional prior $P(Z \mid L)$ and a learnable SCM layer that induces a DAG over latent dimensions. This design supports interventions that propagate through a structured latent space, making it attractive for workflows where labels encode strategic intents.

However, C-VAE is not sufficient for our problem for three reasons. First, C-VAE incorporates related signals ($L$) mainly as auxiliary supervision for latent factor learning. While we also employ labels $L$ to guide representation learning, engagement $Y$ serves as the ultimate causal target whose response to interventions on $L$ must be explicitly predicted. Second, it does not address confounders that jointly affect label selection and engagement. Third, it does not explicitly model the stochastic mediation through $X$, where identical labels can yield variant copies that complicate identification of label effects.

In sum, the literature reveals a divide: deconfounder methods account for confounds but often lack strong counterfactual prediction in complex generative pipelines, while causal representation learning enables intervention in latent space but does not guarantee adequate adjustment for confounders. Our problem requires both: (1) recovering confounders from partially observed data, and (2) modeling the causal mechanism from labels to engagement to support counterfactual prediction without relying on actual copy generation. This motivates combining deconfounding with causal representation learning.

## 3. Research Context and Theoretical Background

This section discusses the research context on LLM-enabled ad generation and provides theoretical backgrounds for estimating the causal effect of labeling skills on user engagement.

### 3.1 Research Context: Marketing Label Attribution

We frame our research context as a marketing attribution problem: how do marketer-selected labels causally influence user engagement in an LLM-assisted creative pipeline? Consider a campaign where marketers specify a binary label vector $L \in \{0,1\}^K$ that encodes the strategic attributes, such as "long-wear" + "skin-nourishing" versus "natural finish" + "oily/acne-prone friendly" characterizing the brand, product, and target persona for some cosmetics product. Each element $L_j = 1$ indicates that label $j$ (e.g., "oil control") is selected, and 0 otherwise. Given $L$, an LLM generates marketing copy $X \in \mathcal{X}$, but this mapping is inherently stochastic: even with the same labels, decoding variability can yield different wording, tone, and narrative framing. Moreover, the generated copy often includes elements not explicitly specified in $L$, such as usage scenarios, temporal framing, or persona cues that can materially affect how audiences respond, added by LLM to improve user engagement. Such improvements reflect the ability of LLM rather than the skills of the marketers, and if not isolated out, can confound the value attribution of labels. User engagement $Y \in \mathbb{R}^N$(e.g., likes, comments, collects) is then realized as the audience reaction to the full creative $X$, rather than to the label choices $L$ alone. This setting implies a mediated pathway, $L \to X \to Y$: labels shape engagement indirectly by steering the generated content, but they do not directly determine it. This creates a core identification challenge for marketing: the same label configuration can produce distinct copies with different engagement

**Example 1: Label-Driven Narrative Divergence in LLM-Generated Marketing Copy**

A marketer uses label vector $L$ to guide the generation of copy $X$. We compare two label configurations differing only in highlighted product benefits:

$\boldsymbol{\ell}_1$: long-wear + skin-nourishing

> Generated copy: This office-to-gallery makeup foundation is a winner! As a CBD office worker, I wore it 12 straight hours—coffee runs, pitches, gallery selfies—and my skin stayed flawless. An intern even asked if it was my natural skin!

$\boldsymbol{\ell}_2$: natural finish + oily/acne-prone friendly (same ingredients).

> Generated copy: Strangers asked for my foundation link on the high-speed train! 17 hours of wear—baggage chaos, client meetings, clubbing—zero patchiness, even under a mask!

Though both makeup foundation copies recommend the same product, they carry distinctive contextual details:

- $\boldsymbol{X}_{\ell_1}$ centers on a CBD office lifestyle, 12-hour wear, and a junior colleague's compliment;
- $\boldsymbol{X}_{\ell_2}$ focuses on a frequent business traveler persona, 17-hour wear, and a stranger's link request.

outcomes, making it difficult to causally attribute engagement differences to labels rather than to stochastic or incidental variations in generated text. **Example 1** illustrates this workflow.

We model this workflow using a SCM $M = \langle \{L, X, Y\}, \{f_X, f_Y\}, P(u) \rangle$, where $X := f_X(L, \varepsilon_X), Y := f_Y(X, \varepsilon_Y)$. Here $\varepsilon_X$ represents unobserved factors in the generation process (e.g., decoding randomness and latent narrative choices) and $\varepsilon_Y$ captures unobserved drivers of audience response. Figure 2(a) visualizes this causal structure.

In defining the causal estimand, we focus on the first-stage content-response effect of label configurations. This scope reflects the operational structure of common large-scale AI-assisted content production. In our setting, advertisers produce hundreds of ad copies per day; at this scale, channels, budgets, and KOL (Key Opinion Leader) selections are not redesigned separately for each label configuration. Instead, distribution resources (budgets, channel and KOL choices) are managed and adjusted at much coarser monthly or quarterly level. Firms typically re-allocate budget or choose a different distribution channel only when certain label combinations stand the test of time. Our causal attribution method provides the foundation to the firm for determining the value of labels. Under this setup, which reflects the practical decision sequence in AI-enabled marketing, value attribution informs later resource allocation, not the other way around.

This implies that, when we estimate how engagement would change if the label configuration were altered for a given creative context, it suffices to hold constant the deployment environment observed in the data. We do not need to anticipate how the choice of a different label would change the resource reallocation in an endogenous manner in our setting.

**3.2 Why label effects are hard to identify: ITE, Mediation, and Confounding**

In practice, marketing decisions operate at both overall portfolio and individual campaign levels[1], each requiring distinct causal estimands. Standard estimands such as the average treatment effect (ATE) effectively summarize population-level engagement shifts and remain highly valuable for identifying mean effects to guide broad resource allocation at the portfolio level in our setting. However, for personalized creative optimization, the central question shifts from "what works on average?" to "how would engagement change if we altered the label configuration for this specific ad campaign?" This requires an individual-level counterfactual query formalized as the Individual Treatment Effect (ITE). ITE isolates the context-specific causal pathway by accounting for stochastic LLM mediation, tracing how a particular label shift alters the generated narrative and subsequently drives engagement for a given audience.

To illustrate this distinction, consider a creative $X_{\ell 1}$ targeting the CBD-office-worker, generated under label $\ell_1$ in Example 1. The operational question is a within-campaign counterfactual: had the label used $\ell_2$ instead, what would engagement have become for this exact product, brand, and visual context? The ITE directly addresses this by evaluating how the label intervention reshapes the copy narrative for that specific campaign. By comparison, an ATE estimate averages across a portfolio of campaigns with different audiences and market conditions. The same label shift may amplify engagement for students while diminishing it for urban professionals. While ATE provides essential macro-level insights for strategy formulation, it inherently masks the heterogeneous mediation mechanisms that marketers must navigate when tailoring individual creatives.

[1] A campaign refers to a targeted advertising initiative, while a portfolio means a collection of campaigns. For example, swapping a label from "speed" to "comfort" to target a different group of customers in a single ad represents a specific campaign, while prioritizing "sustainability" vs. "performance" themes across all Q4 ads consists of a portfolio of campaigns.

Consequently, in our setting where decisions hinge on precise audience-context alignment, the key estimand is an individual-level mediated effect: the change in engagement induced by changing labels through the change in generated copy. Under our SCM $L \to X \to Y$, labels influence engagement only via the LLM-generated copy. Accordingly, for two-label configurations $\ell^0, \ell^1 \in \{0,1\}^K$, the natural direct effect (NDE) is structurally zero:

$$\mathrm{NDE}_{\ell^0,\ell^1}(\mathrm{y}) = \mathrm{P}\left(\mathrm{Y}_{\ell^1,\mathrm{X}_{\ell^0}} = \mathrm{y}\right) - \mathrm{P}\left(\mathrm{Y}_{\ell^0,\mathrm{X}_{\ell^0}} = \mathrm{y}\right) = 0 \tag{1}$$

because there is no direct arrow $L \to Y$ (because advertisers do not display the labels directly to users, users see ad copies instead). The natural indirect effect (NIE) captures the change in engagement driven by the label-induced change in copy:

$$\mathrm{NIE}_{\ell^0,\ell^1}(y) = \mathrm{P}\left(Y_{\ell^0,\mathrm{X}_{\ell^1}} = y\right) - \mathrm{P}\left(Y_{\ell^0,X\ell^0} = y\right). \tag{2}$$

Then the ITE of intervention $(\ell^0 \to \ell^1)$ is

$$\mathrm{ITE}_{\ell^0,\ell^1}(y) = \mathrm{NDE}_{\ell^0,\ell^1}(y) + \mathrm{NIE}_{\ell^0,\ell^1}(y) = \mathrm{NIE}_{\ell^0,\ell^1}(y). \tag{3}$$

Intuitively, NIE asks how engagement would change if we kept the "downstream engagement response mechanism" fixed at label setting $\ell^0$, but replaced the realized copy $\mathrm{X}_{\ell^0}$, with the counterfactual copy $X_{\ell^1}$ that would have been generated under an alternative label configuration. The difficulty is that $\mathrm{NIE}_{\ell^0,\ell^1}(y)$ is generally not identifiable under either Pearl's observational (Layer 1) or interventional (Layer 2) scenarios, when the mediation pathway is affected by unobserved factors (See Figure 1 in Appendix D). Even with randomized assignment of $L$, identifying NIE requires, for the same unit/campaign, access to both the factual outcome under $(\ell^0, \mathrm{X}_{\ell^0})$ and the counterfactual outcome under $(\ell^0, \mathrm{X}_{\ell^1})$. This directly reflects the Fundamental Problem of Causal Inference (Holland, 1986): a single campaign cannot simultaneously realize two different copies generated under different label configurations. In generative marketing, this problem is amplified because $X$ is not a deterministic function of $L$: stochastic decoding and emergent narrative choices inject variability (e.g., lifestyle cues, temporal framing) that are difficult to fully observe or control, and that may correlate with both label selection and engagement.

As a side note, non-identifiability does not imply non-realizability (Raghavan & Bareinboim, 2025). Identifiability asks whether the counterfactual quantity is computable from observed interventional distributions; realizability asks whether we can design a procedure that can approximate or sample from the relevant counterfactual distribution. In our context, while we cannot observe both $Y_{\ell^0, X_{\ell^1}}$ and $Y_{\ell^0, X_{\ell^0}}$ for the same campaign from historical logs, we may still be able to make counterfactual reasoning operational by constructing proxy interventions that emulate the label-induced narrative shift in a controlled space.

This motivates the following latent-space intervention framework. Rather than requiring the platform to (unrealistically) regenerate a copy under alternative labels and re-run the campaign, we learn a causal representation in which the label-induced narrative components can be perturbed in a structured way. Doing so makes the counterfactual query "What would engagement have been under an alternative label configuration?" to be operational (realizable) without relying on repeated trial-and-error deployments or on directly observing impossible counterfactual outcomes.

### 3.3 Latent Space Perspective: Construction and Rationale

Although counterfactual reasoning is theoretically realizable, neither the discrete label space nor the stochastic ad copy space supports valid, unconfounded interventions. Directly editing the realized copy $X$ violates ceteris paribus by altering uncontrolled stylistic or narrative features. Conversely, swapping factual labels and regenerating copy injects uncontrolled decoding variation, breaking the isolation of the $L \rightarrow Y$ pathway. To enable valid proxy interventions, we construct a causally structured latent representation $Z_L$ that separates label-driven variation from stochastic mediation and environmental noise.

The latent space $Z_L$ is not postulated abstractly; rather, it is learned directly from the generative pipeline through a joint optimization objective. Specifically, we extract $Z_L$ from the LLM's high-dimensional hidden-state representation $R = h_{\phi}(X)$ of the generated marketing copy $X$ (Figure 2(b)). Following the Interchange Intervention approach (Li et al. 2026) and the Distributed Alignment Search method (DAS; Geiger et al.,2024), we treat the expert label vector $L$ as a high-level causal variable and learn $Z_L$ as its distributed realization within the model's hidden-state space, rather than as a localized neuron, coordinate, or surface-level textual feature. This extraction is guided by two simultaneous objectives:

1. Label supervision: $Z_L$ is constrained to align with the observed expert label vector $L$, ensuring it captures strategic intent rather than arbitrary stylistic fluctuations;
2. Outcome supervision: $Z_L$ is optimized to predict downstream user engagement $Y$, anchoring the representation in actual market resonance.

Formally, we reparametrized the generative process through $Z_L = h_\theta(R)$ with a loss function $\mathcal{L}(\theta) = \mathcal{L}_{label}(Z_L, L) + \lambda\mathcal{L}_{outcome}(Z_L, Y)$, yielding the structural equations:

$$X = f(Z_L, \varepsilon_X), Y = g(Z_L, \varepsilon_Y), \text{ s.t. } L - L' = 0 \Rightarrow Z_L = Z_L' \Rightarrow \text{E}[\text{Y}|Z_L] = \text{E}[Y'|Z_L']. \quad (4)$$

How $Z_L$ addresses mediator-induced bias? The effectiveness of this design rests on two complementary insights. First, leveraging the LLM's internal representations provides a rich semantic proxy for latent contextual factors (Imai &Nakamura, 2025). LLMs trained on vast, diverse corpora encode, in their hidden states, not only explicit semantics but also narrative conventions, emotional tones, and stylistic priors. These internal states naturally absorb many of the contextual signals that co-vary with treatment generation and audience response. Second, the joint optimization with label supervision and outcome alignment purposefully filters out such stochastic decoding noise. Identical label sets can yield semantically aligned but lexically divergent copies due to LLM-induced randomness. The latent space compresses this variation into a stable, continuous representation that preserves the causal core of marketer intent (underlying skills of deliberately choosing certain labels) while discarding incidental narrative artifacts. Consequently, intervening on $Z_L$ bypasses the noisy $L \rightarrow X$ generation step for counterfactual prediction, isolating the true causal signal of label semantics from mediator-induced stochasticity.

Under this construction, $Z_L$ satisfies two formal requirements for valid counterfactual reasoning:

(1) *Intervention Fidelity (Label-to-Latent Equivalence)*: $Z_L$ preserves the causal role of the labels in the generation mechanism. Intervening on $Z_L$ should reproduce the downstream effect of intervening on $L$: $\mathbb{E}[Y \mid do(Z_L = z)] \approx \mathbb{E}[Y \mid do(L = \ell)]$. Conceptually, this metric is analogous to the *Interchange Intervention Accuracy* (IIA) metric proposed in Li et al. (2026) and Geiger et al. (2024) where intervening on a subset of input space is interchangeable to intervening on the whole space.

(2) *Disentanglement from Stochastic Mediation:* $Z_L$ isolate the stable, label-driven semantic signal from the stochastic decoding noise and emergent narrative artifacts in $X$. By compressing the $L \to X$ pathway into a continuous representation, $Z_L$ makes counterfactual prediction realizable without requiring actual copy regeneration.

However, satisfying these two requirements ensures only that our intervention is coherent and stable; it does not guarantee that the resulting estimates are unbiased in a causal sense. As detailed next, the system remains vulnerable to external confounders—such as campaign budget, platform algorithms, or cross-modal visual factors—that jointly influence both label selection and user engagement. Therefore, achieving robust *Causal Inference of Engagement* requires addressing these remaining sources of bias, which motivates the deconfounding framework discussed next.

### 3.4 Sources and Complexity of Confounding

Even with a well-constructed latent label space $Z_L$, the causal system of our research context, as specified in Figure 3, can remain confounded. Such confounding impedes both (*i*) identification of the causal effect of labels on engagement and (*ii*) learning causal relations *within* the latent label structure ($e.g., Z_{L_i} \to Z_{L_j}$).

The confounders can arise from two sources. The first is copy-induced confounders (embedded in $X$). The generated copy $X$ contains stylistic and narrative artifacts $\widetilde{X_L}$ that are not explicitly encoded in the label vector $L$ but are systematically co-produced with it. Examples include context choice (e.g., "CBD office worker" vs. "frequent traveler" in **Example 1**), emotional tone, temporal framing, and incidental character interactions. These features are not randomized or controlled during generation, and they can influence engagement directly. As a result, they create spurious associations between $L$ and $Y$, and can also induce apparent dependencies across label dimensions. For instance, the frequent co-occurrence of "618" and "gift-for-self" may reflect campaign conventions rather than a true causal link between the underlying label semantics. While such artifacts can be partially summarized via LLM embeddings ($X_L$), they remain a source of confounding unless explicitly accounted for.

The second source comes from unobserved environment ($E_u$). Beyond what is expressed in the copy, there may exist confounders jointly affect label choice and engagement. We group these into:

- Marketer-level factors: tacit expertise, preferences, and risk attitudes that shape how labels are interpreted and translated into narrative form.
- Market-level dynamics: competition intensity, campaign budget, trending topics, and seasonal demand shifts that influence both label selection and consumer response.
- Platform-level factors: algorithm configurations, audience segmentation logic, and cross-campaign interference that affect exposure and engagement but are typically unobserved in logs.

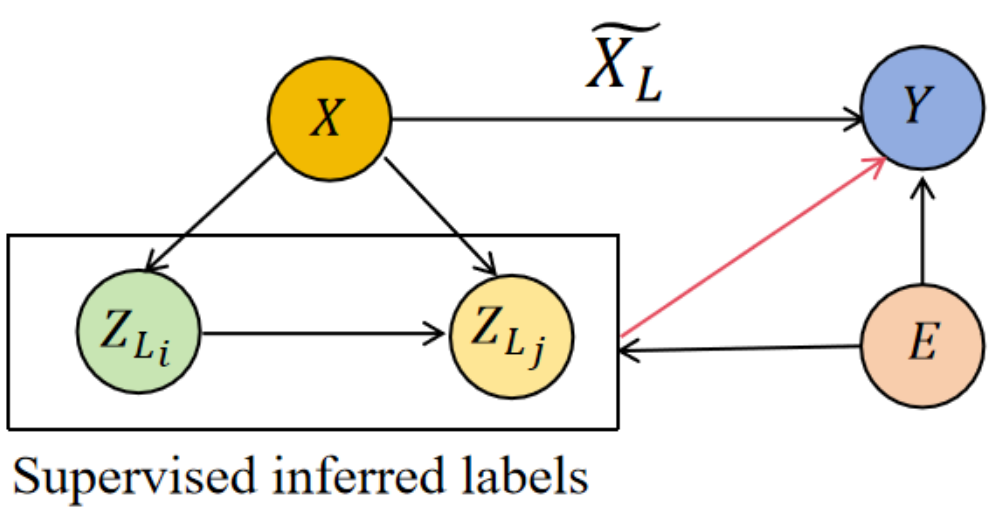

**Figure 3: Proposed Causal Structure**

**Note:** X (marketing copy) and E (environmental variables containing observed C and unobserved $E_u$) are the confounders of $Z_{L_i} \to Z_{L_j}$ and $Z_L \to Y$, where $Z_{L_i}$ (latent causal label variables), $Z_{L_j}$(latent result label variables), Y is the user engagement.

Some observed environmental variables can serve as proxies for parts of the latent environment, denoted $C$ (controls). Examples include publish time, text–image coherence metrics, influencer (KOL) attributes (e.g., followers, historical like/collect rates), and brand popularity. Conditioning on $(X_L, C)$ can reduce bias. However, it does not eliminate unobserved selection, meaning the conditional independence required for causal identification in the latent space generally fails:

$$Y_Z \perp Z_L \mid X_L, C \tag{5}$$

This remaining dependence motivates our use of deconfounder methods in the multiple-cause setting, together with an explicit discussion of omitted-variable bias bounds, since one cannot enumerate and measure all unobserved confounders in generative marketing environments.

## 4. Method: D-VAE

In this section, we propose D-VAE (Deconfounder Variational Autoencoder), a unified framework that addresses the identification challenges discussed in Section 3 for causal skill attribution, whose effects are mediated by LLM. D-VAE jointly models three interwoven structures. First, it characterizes the confounding structure of multi-label assignment, distinguishing shared confounders affecting multiple labels from single-label confounders that cannot be recovered from co-occurrence alone. Second, it uses the recovered confounder information to construct a confounder-adjusted latent representation of expert labeling logic. Third, it embeds this representation in a way that interventions on one label can propagate through learned dependencies among labels.

### 4.1 Structure of Treatment Confounder: Multi- vs. Single-Treatment Confounder

Our approach draws on a key insight from Wang & Blei (2019): when treatments are high-dimensional and combinatorial, their joint patterns reveal information about shared latent confounders that simultaneously influence multiple labels. Intuitively, if certain label combinations (e.g., "co-branded IP" + "beauty enthusiast") repeatedly co-occur across campaigns, this regularity likely stems from a common hidden driver—such as campaign budget or market niche. By modeling these co-occurrence patterns, we can construct a proxy for that shared confounder and account for it.

However, real-world labeling involves a more complex confounding structure: not all confounders affect multiple labels. Some may influence only one label while also boosting engagement, e.g., the temporary popularity of a specific cartoon IP. These single-cause confounders leave no trace in the covariance structure across labels and are therefore unrecoverable by any extant deconfounder methods. Wang & Blei's framework simply assumes such confounders are absent; while our work explicitly factors in their presence.

Our contribution is to prove the theoretical bounds of the impact of single-cause confounders and establish the conditions as to when they are ignorable, relative to the shared confounders. Crucially, we extend Oster's (2019) linear sensitivity logic by constructing a worst-case scenario over a W-2 ambiguity ball following Wasserstein DRO (distributionally robust optimization) approaches (Esfahani & Kuhn 2018). This diagnostic extends naturally to nonlinear models, including neural networks, under Lipschitz-type regularity.

Formally, when dealing with multiple binary treatment $L = (L_1, \dots, L_K)^\top \in \{0,1\}^K$, unobserved confounders $E_u$ may appear in two distinct channels: (1) common confounder $e_m \in R^M$, which simultaneously influences the selection of multiple ($\geq 2$) treatment labels; (2) single-treatment confounders $\{e_s^{(j)}\}_{j=1}^K$, where each $e_s^{(j)} \in R$ affects only the $j$-th treatment $L_j$. This structure is illustrated in Figure 4(a).

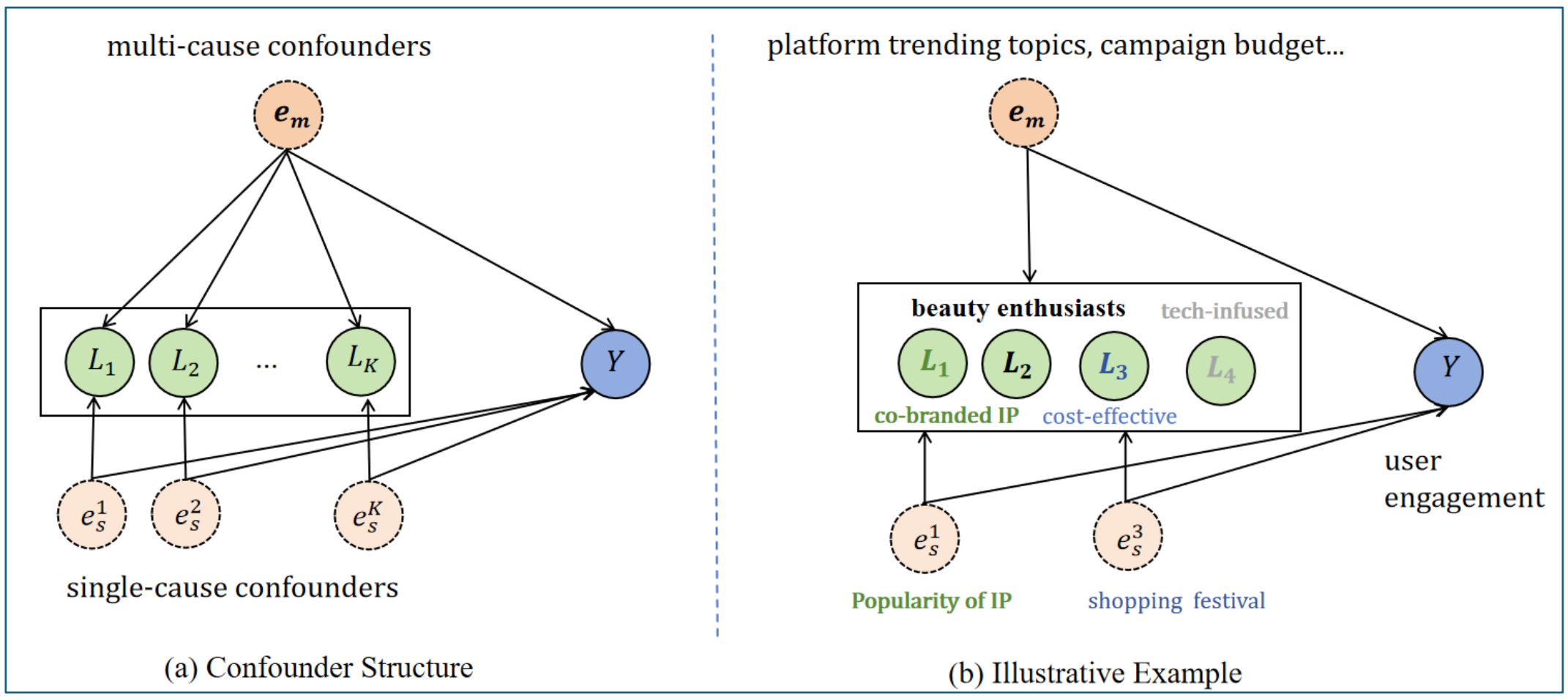

**Figure 4: The Confounders Problem of Causal Inference on Label to User Engagement**

Importantly, while we denote a distinct $e_s^{(j)}$ for each $L_j$ for notational clarity, this does not imply that every treatment necessarily has its own unique unobserved driver, some may share components or be unaffected. In practical multi-label selection settings, common confounders are likely to be more prevalent. This is because label assignment typically occurs in a joint decision process, campaign managers rarely select each label in complete isolation. Consider a marketing campaign, a platform-wide trending topic (e.g., a beauty festival) or campaign budget may act as a common confounder $e_m$, simultaneously increasing the likelihood of selecting multiple labels such as “co-branded IP”, “beauty enthusiasts”, “cost-effective”, and “tech-infused”. And the popularity of a certain IP may serve as a single-treatment confounder $e_s^{(1)}$ that boosts user engagement through the appeal of the IP, while also affecting the adoption of the co-branded IP label. Similarly, a shopping festival $e_s^{(3)}$ may elevate preference for the “cost-effective” label and independently heighten user responsiveness to discount-related content.

### 4.2 Deconfounder under multiple treatments

To operationalize this separation, we impose a factorized latent-confounder model that distinguishes shared cross-label dependency from label-specific residual confounding. The standard Gaussian priors is leveraged to fix the scale of otherwise unobserved latent factors and provide a regularized parameterization. Our deconfounder approach assumes the following generative structure:

$$e_m \sim N(0, I_M), e_s^{(j)} \sim N(0,1), \forall j = 1, \dots, \mathrm{K}, \tag{6}$$

with mutual independence:

$$e_m \perp e_s^{(j)}, e_s^{(j)} \perp e_s^{(k)} \text{ for j} \neq k.$$

The treatment assignment follows a factorized model conditional on the latent confounders:

$$L_j \mid e_m, e_s^{(j)} \sim Bernoulli\left(\sigma\left(\lambda_j^\top e_m + \theta_j e_s^{(j)}\right)\right), j = 1, \dots, \mathrm{K}, \tag{7}$$

or equivalently, $\mathrm{P}(L_1, \dots, L_K \mid e_m, e_s) = \prod_{j=1}^{K} P\left(L_j \mid e_m, e_s^{(j)}\right)$, where $\mathrm{e_s} = \left(\mathrm{e_s^{(1)}}, \dots, \mathrm{e_s^{(K)}}\right)^\top$, $\lambda_{\mathrm{j}} \in \mathrm{R^M}$, and $\theta_j \in R$.

Building on the factorized treatment-assignment model above, we specify the following outcome equation:

$$Y = \beta^\top L + \Gamma e_m + \Theta^\top e_s + \widetilde{X_L}^\top \alpha + \rho^\top C + \epsilon \tag{8}$$

where $e_m$ and $e_s$ capture pre-treatment confounding in label assignment: $e_m$ absorbs common sources that induce cross-label dependency, while $e_s$ captures treatment-specific assignment heterogeneity. In contrast, $\tilde{X}_L$ plays a different role. Let

$$X_L = \phi_{\mathrm{LLM}}(X), \tilde{X}_L = X_L - \mathbb{E}[X_L \mid L], \tag{9}$$

where $X_L \in \mathbb{R}^T$ denotes the LLM representation of the generated copy. The residual $\tilde{X}_L$ is the $L$-orthogonal component of the representation, isolating post-assignment variation that is not explained by the label vector $L$ but may still affect $Y$. Thus, while $e_m$ and $e_s$ deconfound the assignment of labels, $\tilde{X}_L$ partials out mediator-side nuisance variation in the generated copy. $C \in \mathbb{R}^Q$ denotes observed covariates (controls), $\epsilon$ denotes the noise term $\epsilon \sim \mathcal{N}(0, \tau^2)$. Then given $e_m, e_s, \widetilde{X_L}, C$, the ignorability condition holds:

$$Y(l) \perp L |\, e_m, e_s, \widetilde{X_L}, C \tag{10}$$

Under this condition, the estimator of $\hat{\beta}$ in the regression model would yield an unbiased estimate of the true causal effect $\beta$. Specifically, the projection of the unobserved confounders onto the treatment space is eliminated, ensuring that $\mathbb{E}[\hat{\beta}] = \beta$. As specified in Figure 3, we decompose the confounder source as environmental factors $E = (C, E_u)$. The unobserved environmental component $E_u$ is represented by the latent factors $(e_m, e_s)$, while the observed environmental component is captured by $C$, and the copy-level confounder is controlled by conditioning on the LLM-derived representation of the generated copy $X_L$ in the outcome model, which yields the following deconfounding property:

**Theorem 1 (Conditional Deconfounding of $Z_L$):** Suppose that $e_m, e_s$, $C$, and $X_L$ sufficiently capture the unobserved environmental, observed environmental, and copy-level confounders, respectively. Then, under the specified DGP (8),

$$Y(z) \perp Z_L \mid e_m, e_s, X_L, C \tag{11}$$

Consequently, after adjustment for $(e_m, e_s, X_L, C)$ the estimator $\hat{\beta}$ obtained through $Z_L$ is unbiased for $\beta$, the true causal effect of $L$ on $Y$: $\mathbb{E}\big[\hat{\beta}\big] = \beta$.

### 4.3 Bound of the Single Treatment Confounders

However, single-treatment confounders $\{e_s^{(j)}\}$ are unidentifiable from the observed treatment matrix alone. While we may observe some covariates $C \in R^Q$ that partially proxy for these latent factors, they are unlikely to fully capture all sources of single-cause idiosyncratic variations. This raises a critical question: if we estimate the common confounder $e_m$ while omitting the single-treatment confounders $e_s$ but partially controlled by $C$, when can the resulting omitted variable bias be ignorable?

This question reveals a key limitation acknowledged in Wang & Blei (2019): their deconfounder assumes all unobserved confounding is shared across treatments, rendering it oblivious to single-cause confounders that are common in real-world marketing. Because such confounders generate no cross-label correlation, they cannot be recovered by factor models, yet they can still bias causal estimates. To answer this, we establish a theoretical Probably Approximately Correct (PAC) bound on the omitted bias (Theorem 2) to assess whether $e_m$ dominates $e_s$. When $e_s$ is omitted, given $n$ samples, we get the estimator $\hat{\beta}$ under the following model:

$$Y_i = L_i^{\top}\hat{\beta} + \hat{e}_{m,i}^{\top}\hat{\Gamma} + \tilde{X}_{Li}^{T}\hat{\alpha} + C_i^{\top}\hat{\rho} + \hat{\epsilon}_i \tag{12}$$

where $\widehat{e_m}$ estimated via Logit-Probabilistic PCA (Logit-PPCA) of $L$. The bias from $\mathrm{e}_s$ is scaled by the fraction of outcome variance unexplained after conditioning on $\mathrm{e}_m$. If $\mathrm{e}_m$ dominates the label selection process, the residual variance attributable to $\mathrm{e}_s$ is small, naturally capping the worst-case bias. This yields our main theoretical result:

**Theorem 2 (Bound under Single-Treatment Confounding):** Under the above DGP (eq. 12), with $\widehat{e_m}$ estimated via Logit-PPCA, and assuming (i) $e_m$ dominates $e_s$ in explaining label co-variation, and (ii) $\lambda_{min}\left(E\left[Var\left(L \mid e_m, e_s, \widetilde{X_L}, C\right)\right]\right) \geq \lambda > 0$, then for any $\eta \in (0,1)$, there exists a constant $A_0$ (depending only on model primitives) such that:

$$P\left(\| \hat{\beta} - \beta \|_2 \leq A_0\left(1 + \sqrt{M/n + M/K}\right)\right) \geq 1 - \eta. \tag{13}$$

The key insight is that the omitted bias from $e_s$ is bounded in population, while the estimation error in $\widehat{e_s}$ decays at rate $\sqrt{M/n + M/K}$. Together, these ensure that the estimation bias interval remains narrow even when single-treatment confounders are unobserved.

We then extend this guarantee to more generic nonlinear outcome settings, such as neural networks, provided the structural function belongs to a Lipschitz-continuous class with bounded weights and activations: $Y = f\left(L, e_m, \widetilde{X_L}, C; \beta\right) + \epsilon$ and the estimator is the M-estimator (empirical risk minimizer)

$$\hat{\beta} = \arg\min_{\beta \in B} \frac{1}{n}\sum_{i=1}^{n} \ell(Y_i, f(L_i, \hat{e}_{m,i}, \tilde{X}_{Li}, C_i; \beta)), \tag{14}$$

where $\ell$ is a convex loss. The same PAC bound holds with an identical convergence rate. This result hinges on the multi-label structure, because labels are selected jointly and exhibit rich combinatorial variation, their observed correlation matrix reveals the dominant source of confounding. The full proof is provided in **Online Appendix A**.

In practice, real data may violate this dominance assumption. We therefore develop an empirical sensitivity diagnostic test to assess its robustness based on observable-control shifts. By tracking how the estimated effect changes as observed confounders are progressively adjusted in the bias decomposition $\operatorname{plim}\hat{\beta} = \beta + B_{\hat{e}_m} + B_{e_s}$, the diagnostic quantifies how large the remaining unobserved bias would need to be to materially change the estimate, without imposing subjective sensitivity parameters. Let $P_0$ denote a reference distribution embodying the equal selection benchmark $\delta = 1$, where the relative selection strength of $e_s$ matches that of the estimated shared proxy $\hat{e}_m$. We consider all plausible confounding distributions within a Wasserstein-2 ball of radius $\varepsilon \geq 0$ around $P_0$: $\mathcal{P}_\varepsilon = \{P: W_2(P, P_0) \leq \varepsilon\}$, $\varepsilon = |\delta_{obs} - 1|$. This nonparametric construction bounds how far the true distribution of unobservables can deviate from the dominance condition without imposing restrictive parametric forms on $e_s$. The Wasserstein-2 distance is particularly well-suited for nonlinear outcome models. By the Kantorovich–Rubinstein duality, $W_2$ controls the expectation of any 1-Lipschitz function of the confounders. Combined with the Lipschitz continuity of the structural function $f$(with constant $L_f < \infty$, as assumed in Section 4), any bias or marginal effect under $P \in \mathcal{P}_\varepsilon$ deviates from that under $P_0$ by at most $O(L_f \cdot \varepsilon)$.

We compare two models: **(1)** *Uncontrolled Regression:* $Y_i = L_i^\top \beta + +\tilde{X}_{Li}^T \alpha + C_i^\top \rho + \epsilon_i$, with estimator $\beta_{\text{uncontrolled}}$, $R^2 = R_{\text{uncontrolled}}$ (or pseudo-$R^2$ based on the empirical loss); **(2)** *Controlled Regression:* $Y_i = L_i^\top \beta + \hat{e}_{m,i}^\top \Gamma + \tilde{X}_{Li}^T \alpha + C_i^\top \rho + \epsilon_i$, with estimator $\tilde{\beta}$, $R^2 = \tilde{R}$.
Then invert Oster's formula (adapted to the multi-treatment and M-estimator setting) to obtain the observed selection ratio:

$$\hat{\delta}_{\text{obs}} = \frac{\tilde{R} - R_{\text{uncontrolled}}}{R_{\text{max}} - R_{\text{uncontrolled}}} \tag{15}$$

A small $\varepsilon$ indicates that $\hat{e}_m$ has already absorbed the dominant fraction of confounding variation, leaving minimal explanatory power for $e_s$ to exploit.

For any $P \in \mathcal{P}_\varepsilon$, the bias-adjusted causal effect (extended to M-estimators via Lipschitz-bounded gradients) takes the closed form:

$$\beta_{\text{adj}}(P) = \tilde{\beta} + \delta(P)\frac{\tilde{\beta} - \beta_{\text{uncontrolled}}}{\tilde{R} - R_{\text{uncontrolled}}}\left(R_{\max} - \tilde{R}\right) \tag{16}$$

where $\delta(P) = \text{Cov}(L, e_s \mid P)/\text{Cov}(L, \hat{e}_m)$ is the effective selection ratio under $P$. The worst-case identified set for the causal parameter is therefore:

$$[\beta_L, \beta_U] = \left[\inf_{P\in\mathcal{P}_\varepsilon}\beta_{\text{adj}}(P), \sup_{P\in\mathcal{P}_\varepsilon}\beta_{\text{adj}}(P)\right]. \tag{17}$$

Under the maintained identification condition, the width of $[\beta_L, \beta_U]$ grows linearly with $\varepsilon$ $(\text{i.e.}, O(L_f \cdot \varepsilon)$ but remains strictly bounded by $O\left(\sqrt{M/n + M/K} + L_f \cdot \varepsilon\right)$.

We validate the PAC bound by comparing the empirical identified set with the theoretical PAC radius $r_{\text{PAC}} = C\left(1 + \sqrt{M/n + M/K}\right)$ from Theorem 2. The bound is deemed robust and informative if: $\beta_U - \beta_L \leq 2r_{\text{PAC}}$, and $[\beta_L, \beta_U]$ overlaps with the PAC interval centered at $\tilde{\beta}$.

In our empirical setting (**Appendix C Algorithm 1),** we find that after controlling for both observed confounders $C$ and the estimated common factor $\widehat{e_m}$, the relative ratio 0.07% is rather small, and the resulting bias-adjusted region for β: $\left[\hat{\beta} - \text{bias}, \hat{\beta}\right]$, does not contain the zero vector, and the bound is robust across reasonable choices of δ. This provides strong evidence that our deconfounder strategy effectively mitigates the dominant source of omitted-variable bias in multi-label treatment settings. Formally, we provide the following corollary.

**Corollary (Proxy-Controlled Unconfoundedness)**: Given the observed marketing copy *X* (the corresponding LLM representation $X_L$ ), environmental variables *C* and estimated common confounder $\widehat{E_m}$, the potential outcome $Y_z$ and label space $Z_L$ are approximately satisfying

$$Y_z \perp Z_L \mid X_L, \text{C}, \widehat{E_m} \tag{18}$$

**4.4 Counterfactual Prediction via Causal VAE**

After deconfounding, we then make counterfactual inference through intervening latent space $Z_L$. We aim not to reconstruct or edit content, but to infer how changes in label would alter user engagement.

This differs fundamentally from C-VAE, which embeds a learnable SCM into the latent space to support do-interventions on weakly supervised label factors for which the outcome of interest is the reconstructed sample itself (Yang et al., 2021). Further, C-VAE lacks two essential components for our task: (1) it assumes all relevant confounders are either observed or ignorable, an unrealist assumption that we directly address through deconfounder and (2) it uses a single-path latent representation that forces causal factors to also encode environmental noise from $X$ and $E$, thereby contaminating the causal signal with spurious correlations. Without explicit modeling these confounders or having a mechanism to isolate the engagement-predictive component of label semantics, C-VAE's interventions may reflect spurious correlation rather than true causal effects.

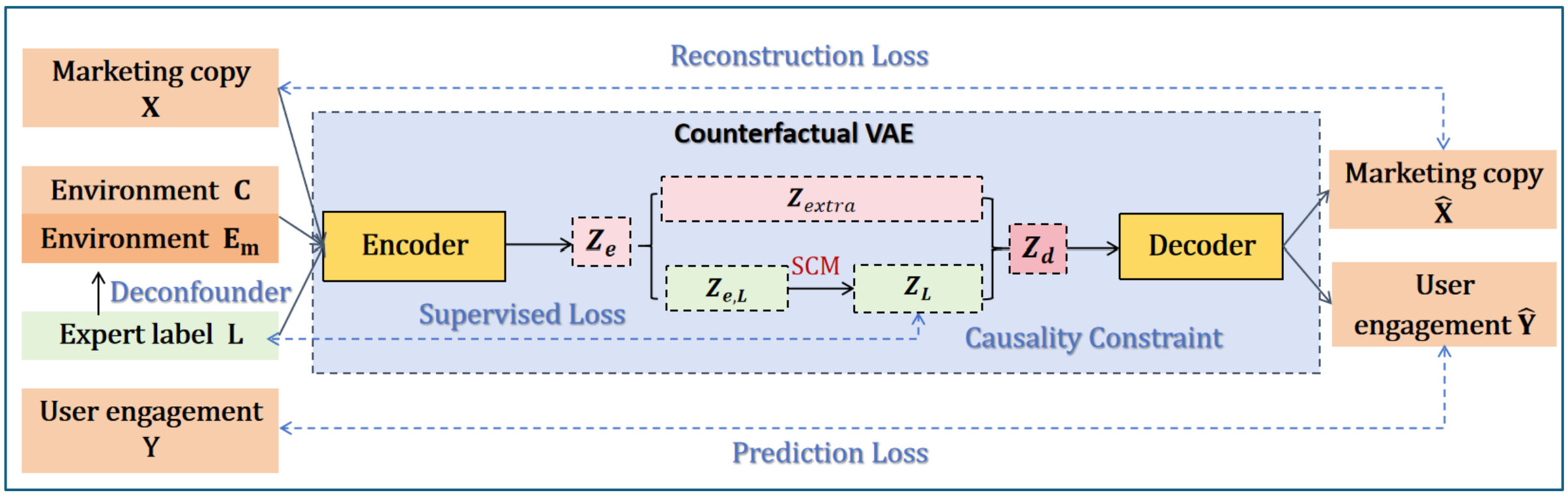

**Figure 5: Architecture of the D-VAE**

Note: Inputs $X$ (marketing copy), $L \in \{0,1\}^K$ (label vector), $C$ (observed covariates), and $\hat{E}_m$ (estimated common confounder) are jointly encoded into a structured latent space $Z_e = [Z_{e,L}, Z_{\text{extra}}]$. The causal subspace $Z_{e,L} \in \mathbb{R}^K$ is refined by a SCM layer (adjacency matrix $A$) to yield the intervention-ready representation $Z_L$. The auxiliary subspace $Z_{\text{extra}}$ captures residual variation from $X$, C, and $\hat{E}_m$. The full latent state $Z_d = [Z_L, Z_{\text{extra}}]$ is decoded to reconstruct $\hat{X}$ and predict user engagement $\hat{Y}$.

In contrast, our D-VAE addresses both gaps by integrating an estimated shared confounder (Section 4.1) and a dual-path latent architecture that decouples causal label semantics ($Z_L$) from auxiliary variation ($Z_{extra}$). This enables a principled multi-task formulation where $Z_L$ parallelly supports marketing copy reconstruction and engagement prediction. Grounded in behavior-aware VAE frameworks (Rao et al., 2023), sharing $Z_L$ across tasks captures robust global semantics, while task-specific decoding prevents overfitting. Specifically, reconstruction acts as a semantic regularizer to preserve textual fidelity, whereas prediction aligns the latent space with the downstream behavioral objective. Crucially, engagement ($Y$) is strictly excluded from the encoder. Given the causal pathway $L \rightarrow X \rightarrow Y$, conditioning on $Y$ induces post-treatment and mediator-collider bias, contaminating $Z_L$ with outcome-dependent shortcuts and violating the counterfactual premise of $do(L)$ interventions. By learning $Z_L$ solely from confounder-adjusted $X$ and optimizing it via parallel losses, our architecture ensures that engagement forecasts under label interventions remain causally valid, robust to stylistic noise, and free from outcome leakage, as illustrated in Figure 5.

The model takes as input the marketing copy $X$, label vector $L \in \{0,1\}^K$, observed environment covariates $C$, estimated unobserved common confounders $E_m$ from Section 4.1. Specifically, the marketing copy is represented by the last hidden state of LLM. These are encoded into a structured latent space $Z_e = [Z_{e,L}, Z_{\text{extra}}]$, where $Z_{e,L} \in R^P$ is dedicated to the representation of labels. Since the dimensions of $Z_{e,L}$ is independent, in order to model dependencies among labels, we introduce a Causal Layer (Figure 6a) that applies a learnable adjacency matrix $A \in R^{P \times P}$ (parameterizing directed information flow across label dimensions) and leak the information $\epsilon$ from first reparametrized $Z_{e,L}$:

$$z_j = f_j\left(\sum_{k=1}^{K} A_{jk} \cdot z_{e,L_{jk}}, \epsilon_j; \theta_j\right). \tag{19}$$

The DAG constraint $\mathcal{L}_{DAG} = \mathrm{tr}\left(e^{A \odot A^T}\right) - P$ enforces acyclicity, and sparsity is encouraged by $\mathcal{L}_{sparse} = \|A\|_1$.

When we intervene on a specific label $L_j$ to get counterfactual (flip the label and keep other labels unchanged), the intervention is applied to the corresponding latent space $Z_L$, which then interacts with other latent labels through the Latent Interaction Network $A$ to directly influence user engagement (Figure 6b). For instance, activating the *"dating-ready"* label directly perturbs the label-aligned latent space $\mathrm{Z}_L$, which then propagates through A. This propagation induces structured shifts in the latent space $\mathrm{Z_L}$: dimensions causally linked to social/romantic contexts are activated, whereas those encoding routine or purely functional usage are suppressed. Then the latent space capturing the interactions between labels $Z_d = [Z_L, Z_{\text{extra}}]$ is decoded to marketing copy $X$ and the user engagement *Y*. The updated joint representation $\mathrm{Z}_d = [\mathrm{Z}_L, \mathrm{Z}_{\text{extra}}]$ thus captures the scenario-shifted semantic distribution. When decoded, these latent changes translate into precise textual adaptations—such as shifting from baseline "daily commute" phrasing to "dating-ready, complexion enhancing" cues, full example can be seen in **Appendix B**.

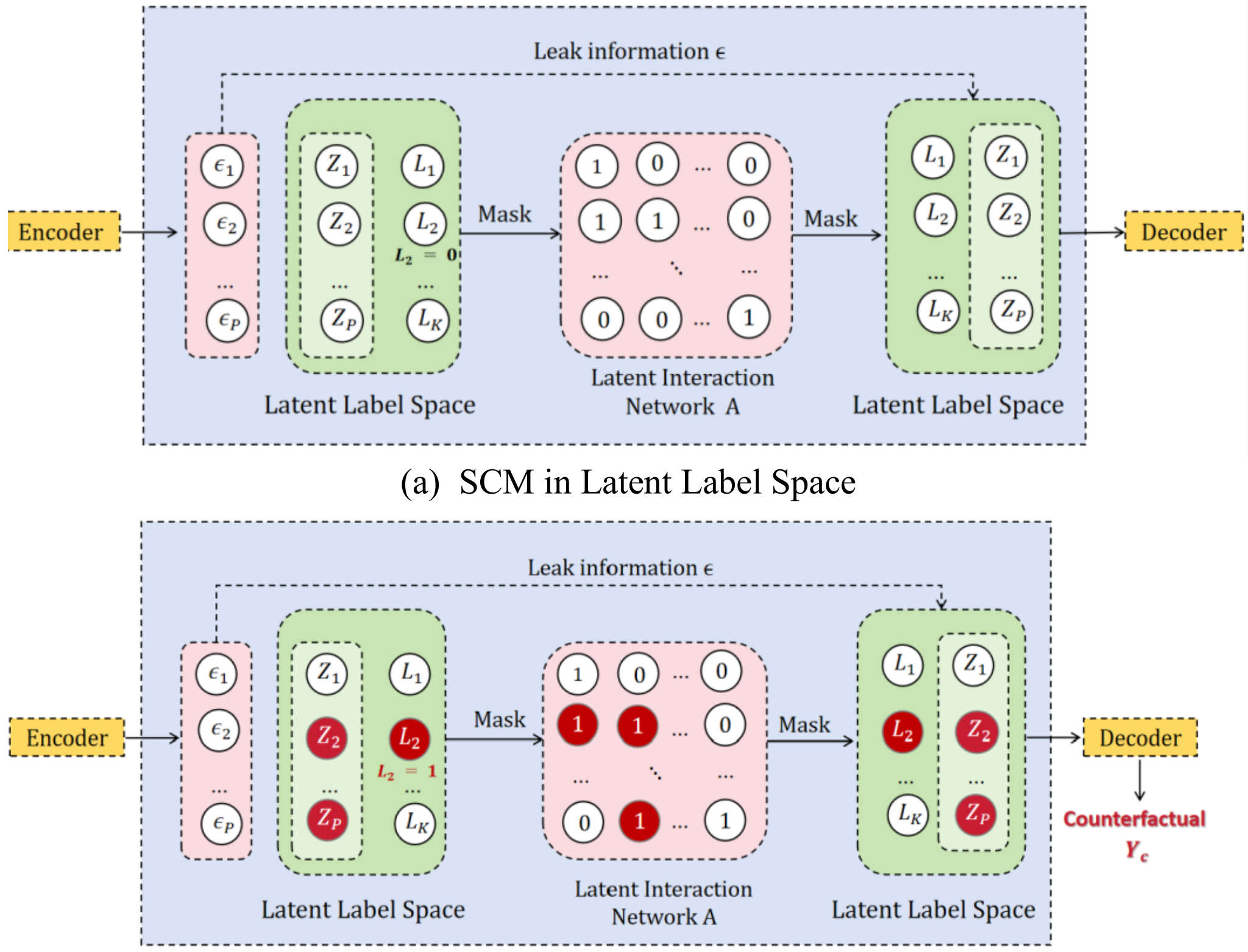

(b) Flip on Factual $L_j$ through label space $Z_L$ to get counterfactual $Y_c$

**Figure 6: Causal Layer of D-VAE and Counterfactual Estimation**

The inference of $Z_L$ as a proxy for intervention relies on two mechanisms. First, we enforce label supervision via

$$\mathcal{L}_{label} = \|g(Z_L) - L\|_2^2 \tag{20}$$

which ensures $Z_L$ faithfully reconstructs $L$, where $g$ is an MLP. Second, $Z_L$ is used in the decoder to reconstruct both $X$ and $Y$. Specifically, the text reconstruction loss

$$\mathcal{L}_X = \left\|X - \hat{X}\right\|_2^2 \tag{21}$$

forces $Z_L$ to learn how label selection $L$ shape the marketing copy $X$, while the outcome prediction loss

$$\mathcal{L}_Y = \left\|Y - \hat{Y}\right\|_2^2 \tag{22}$$

ensures $Z_L$ captures how labels drive engagement.

However, restricting all causal reasoning to $Z_L$ risks discarding information about other covariates $(\widetilde{X_L}, \widehat{E_m}, C)$ due to the label supervision, leading to prediction bias. As Melnychuk et al. (2023)

emphasize, low-dimensional causal representations reduce variance in low-sample CATE estimation but often violate ignorability assumption. To resolve this, we introduce an auxiliary subspace $Z_{extra} \in R^{d_{extra}}$ that captures residual variation from $\widetilde{X_L}$, $C$, and $\widehat{E_m}$. The full latent space $Z = [Z_L, Z_{extra}]$ is used in the decoder for reconstruction and prediction, ensuring confounders are controlled without contaminating the causal subsystem since $Z_L$ and $Z_{extra}$ are independent and the SCM structure doesn't change $Z_{extra}$. These guarantee that intervening on $Z_L$ simultaneously controls $X$ and other downstream factors to predict $Y$.

The inductive biases (Li et al., 2024) in this case are:

$$\mathcal{L}_{causal_\ell} = CE\big(\mathcal{C}(L_i, \Omega), L_j\big) - \gamma CE\big(\mathcal{C}(L_{i_0}, \Omega), L_j\big),$$

$$\mathcal{L}_{causal_y} = MSE(\mathcal{D}(L_i, \Omega), Y) - \delta MSE\big(\mathcal{D}(L_{i_0}, \Omega), Y\big). \tag{23}$$

where CE and MSE stand for Cross-Entropy loss and Mean Square Error, respectively. $\mathcal{C}$ is a classifier of labels, $\mathcal{D}$ is the decoder, $L_i$ is the original value, $L_{i_0}$ denotes that wiping out $L_i$ by setting it to zero, $\Omega$ are other inputs of classifier and predictor, γ, δ denotes a dynamic parameter. We introduce Counterfactual forward (iteratively feeding counterfactual generated by flipping on $L_i$ to $L_{i_0}$ through the model during training) to remove the inductive bias **(Online Appendix C Algorithm 3).**

The model is trained under a causally informed ELBO with other mentioned constraints,

$$\mathcal{L}_{ELBO} = E_{q_Z}[\log p(X|Z_d)] + E_{q_Z}[\log p(Y|Z_d)] - D_{KL}\left(q\big(Z_d\big|X, L, C, \widehat{E_m}\big) \,|\, p\big(Z_d\big|X, L, C, \widehat{E_m}\big)\right) \tag{24}$$

which consists of two parts: the reconstruction term, which decomposes into $\mathcal{L}_X$ (for copy $X$) and $\mathcal{L}_Y$ (for user engagement); the KL divergence term which regularizes the posterior toward a confounder-conditioned prior to preserving ignorability.

By Lagrange multiplier method, we have the new loss function:

$$\mathcal{L}_{D-VAE} = -\mathcal{L}_{ELBO} + \alpha\mathcal{L}_{label} + \beta\big(\mathcal{L}_{DAG} + \mathcal{L}_{sparse}\big) + \xi\left(\mathcal{L}_{causal_\ell} + \mathcal{L}_{causal_y}\right) \tag{25}$$

where $\alpha, \beta, \xi$ denote regularization hyperparameters. The detailed algorithm implementation is in Appendix C algorithm 2.

**Theorem 3 (Causal Consistency of D-VAE):** Under the *Proxy-Controlled Unconfoundedness* condition, let $\hat{Y}_i(\ell)$ denote the counterfactual outcome predicted for unit $i$ by intervening on $Z_L$ according to label configuration $\ell$, D-VAE is causally consistent provided that the latent-intervention approximation error and outcome-prediction error vanish as the sample size grows:

$$\hat{Y}_i(\ell) \xrightarrow{p} \mathbb{E}\left[Y_i(\ell) \mid C_i, \hat{E}_{m,i}, Z_{\text{extra},i}\right]. \tag{26}$$

Consequently, for any two label configurations $\ell_1$ and $\ell_0$, the estimated causal effect for unit $i$ satisfies

$$\hat{\tau}_i(\ell_1, \ell_0) = \hat{Y}_i(\ell_1) - \hat{Y}_i(\ell_0) \xrightarrow{p} \tau_i(\ell_1, \ell_0) \equiv \mathbb{E}\left[Y_i(\ell_1) - Y_i(\ell_0) \mid C_i, \hat{E}_{m,i}, Z_{extra,i}\right]. \tag{27}$$

Averaging $\hat{\tau}_i(\ell_1, \ell_0)$ across units yields a consistent estimate $\hat{\tau}(\ell_1, \ell_0)$ of the corresponding ATE $\tau(\ell_1, \ell_0) \equiv \mathbb{E}[\tau_i(\ell_1, \ell_0)]$.

In summary, the workflow of D-VAE (Figure 7) first recovers multiple treatment confounders via Logit-PPCA and computes a theoretical PAC bound, then validating whether the empirical W-bound width satisfies the threshold condition $W_{[\beta_L, \beta_U]} \leq 2r_{\text{PAC}}$; if violated, iterative deconfounding is applied until the bound is satisfied, or otherwise terminates. Once this condition holds, the framework activates the D-VAE architecture to learn a causally disentangled latent label space $Z_L$, enabling robust counterfactual engagement prediction and causal inference of label combination.

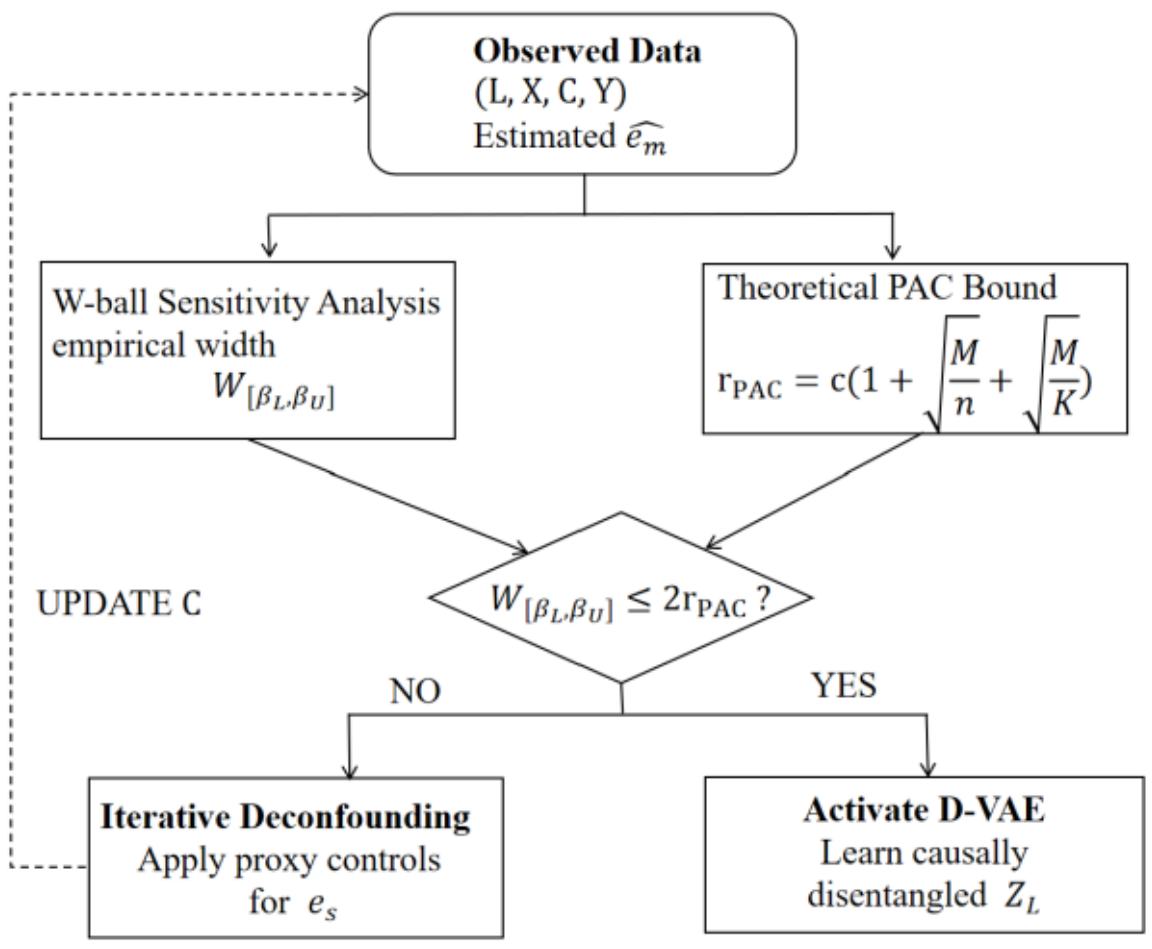

**Figure 7: Workflow of D-VAE**

Note: The iterative deconfounding denotes a refinement step that seeks to recover or proxy for $e_s$ when the PAC W-bound condition is not satisfied.

## 5. Data, Analysis and Results

### 5.1 Data Description and Variables

Our data are provided by a leading Chinese marketing agency with over 20 years of experience in the advertising industry. The agency was among the first in China to integrate generative AI into its content production pipeline and has developed a mature, expert-driven labeling system for marketing creatives. The dataset contains 29,177 branded contents (BGC) and professionally generated contents (PGC) used in advertising creatives across three complexion product categories: liquid foundation, pressed powder, and cushion compacts. All content was launched on the Little Red Book platform (also known as RED or Xiaohongshu). These three categories share similar formulation technologies & applications, and target customer segments—such as coverage, longevity, skin feel, and compatibility with sensitive or oily skin—making them well suited for studying the transferability of expert labeling logic. Moreover, these categories are among the most actively marketed and consumer-engaged beauty segments on RED. As a result, they generate high-volume, high-quality content that reflects real-world GenAI deployment in expert-guided marketing workflows. The sample includes posts created by the agency between June 1, 2024, and September 1, 2025 using an internally fine-tuned LLM specialized in beauty copywriting.

The majority of these marketing copies range from 50 to 400 Chinese characters in length. For each post, we examined three user engagement metrics: number of likes, comments, and collects (See Figure 1 for an example). As shown in Figure 2 in **Appendix D**, these metrics exhibit a typical power-law distribution; therefore, we applied a logarithmic transformation to all engagement variables in subsequent modeling to mitigate skewness.

The dataset is accompanied by a glossary of 72 expert-curated labels, derived from their marketing experience. These labels span multiple dimensions: target audience (e.g., age, occupation, usage scenario), product ingredients, claimed benefits, sensory experience, brand attributes, and content format. How to choose the right labels in a campaign represents the core marketing skill of these experts.

**Table 1. Descriptive Statistics of Variables**

| Type | Variable | Definition | count | mean | std | min | max |
|---|---|---|---|---|---|---|---|
| Engagement | Likes | The number of user likes | 29,177 | 38.68 | 126.48 | 0 | 2,908 |
| | Comments | The number of user comments | 29,177 | 168.56 | 338.22 | 0 | 3,000 |
| | Collects | The number of user collects | 29,177 | 62.28 | 141.35 | 0 | 2,792 |
| Label Description | Number of Assigned Labels[2] | Number of marketing labels assigned with the marketing copy | 29,177 | 11.24 | 7.08 | 2 | 46 |
| Controls | Copy_len | Length of the marketing copy in Chinese characters | 29,177 | 355.05 | 231.93 | 3 | 1,930 |
| | Copy_image_coherence | Semantic coherence similarity score between text and images | 29,177 | 0.23 | 0.12 | 0 | 0.51 |
| | Emojis_num | number of emojis used in the marketing copy | 29,177 | 3.61 | 5.99 | 0 | 89 |
| | Image_num | Number of images included in the post | 29,177 | 5.67 | 3.77 | 0 | 19 |
| | Brand_popularity | Average engagement score of the brand on Red | 29,177 | 3175.10 | 3,925.15 | 0 | 46,899 |
| | KOL_fans | Number of followers of KOL | 29,177 | 54,958.97 | 168,253.96 | 58 | 8,366,415 |
| | KOL_works | Number of posts previously published by the KOL | 29,177 | 469.39 | 819.91 | 1 | 12,700 |
| | KOL_likes | KOL's historical total likes | 29,177 | 318,067.83 | 886,948.43 | 12 | 99,717,750 |
| | KOL_collects | KOL's historical total collects received | 29,177 | 131,215.11 | 319,767.04 | 0 | 13,714,744 |
| | Kol_level | KOL tier as classified by the platform (tier 1 = top-tier, tier 2 = mid-tier, …) | 29,177 | 3.99 | 0.87 | 1 | 6 |
| | Is_holiday | Whether the copy was published on a public holiday (0 = no, 1 = yes) | 29,177 | 0.05 | 0.22 | 0 | 1 |
| | Day_of_week | The day of week the copy was published (0-6 represents Monday-Sunday) | 29,177 | 2.73 | 1.86 | 0 | 6 |
| | Publish_hour | The hour of day the copy was published (0-23) | 29,177 | 8.25 | 3.96 | 0 | 23 |

For textual representation of marketing copies ($emd(X)$), we used the last hidden state of Qwen3-32B, a leading open-source LLM, as the semantic embedding of each copy. Additionally, we counted the number of emojis in each post (Das et al., 2019), a feature particularly salient on RED, where emotional and visual cues also drive user engagement. For multimodal content, we computed text-image coherence using the Qwen-VL model and recorded the number of images included in each post (*Copy_image_coherence*). We also controlled for several external factors related to channel and platform in distribution stage that may influence engagement: whether the marketing copy was published on a public holiday *(Is_holiday*), on a weekend (*Day_of_week*), the specific hour of publication (*Publish_hour*), the overall brand popularity index on RED (*Brand_popularity*), and key KOL characteristics, including follower count (*KOL_fans*), total number of past posts (*KOL_works*), historical total engagement (*KOL_likes*, *KOL_collects*), and platform-verified KOL tier (*KOL_level*). Table 1 presents the descriptive statistics for those variables.

[2] The 72-label glossary generates a high-dimensional combinatorial assignment space, with 2,448 observed two-label combinations and 50,013 observed three-label combinations in the data.

### 5.2 Model Evaluation and Results

#### 5.2.1 Deconfounding Performance: Confounding Bias Mitigation

First, we evaluate whether our proposed deconfounding strategy effectively mitigates unobserved confounding effects. We implement the deconfounder framework to identify the common confounders $E_m$ of multi-treatment $L$. Then we assess deconfounder performance following Wang & Blei (2019) in terms of out-of-sample predictive score:

$$\text{Predictive Score} = \mathbb{P}\left(t\left(L_{\text{rep}}^{\text{held}}\right) < t\left(L^{\text{held}}\right)\right), t(\cdot) = \mathbb{E}_{E_m|L^{\text{obs}}}\left[\log p\left(L^{\text{held}} \mid E_m\right)\right], \tag{28}$$

which tests whether the factor model can plausibly reconstruct unseen(held-out) labels; a score of 0.72 $\gg$ 0.1 of our method confirms effective capture of shared confounding.

To empirically assess the PAC guarantee from Theorem 2, we examine the W-bound as specified in Section 4.3. Given $M = 12$, $n = 29{,}177$, and $K = 72$, with $\eta = 0.05$, we calibrate $c = 0.17$ based on the observed signal-to-noise ratio (mean causal effect $\approx 0.081$, residual variance $\approx 0.78$) and the $R^2$ gain $\approx 0.13$, yielding a theoretical PAC radius of $r_{\text{PAC}} = 0.243$. The data-driven Wasserstein radius $\varepsilon = \mid \hat{\delta}_{\text{obs}} - 1 \mid$ produces an empirical bias width of $W_{[\beta_L, \beta_U]} \approx 0.033$ and a mean relative bias of $0.07\%$. Applying the diagnostic criterion $W_{[\beta_L, \beta_U]} \leq 2r_{\text{PAC}}$, the bound is deemed highly informative, as the empirical width is substantially smaller than the 95% high-probability threshold. This confirms that the small-bias condition holds robustly as we discussed before *Proxy-Controlled Unconfoundedness Corollary*, validating the ignorability of single-treatment confounds for the subsequent D-VAE causal inference.

#### 5.2.2 D-VAE performance: ITE Estimation

Next, we show how D-VAE tackles with causality in terms of: (1) *Intervention Fidelity* and (2)

*Causal Inference of Engagement* as discussed in Section 3.2. *Intervention Fidelity* examines whether interventions in the latent label space $Z_L$ faithfully reproduce the downstream effects of directly manipulating the original label vector $L$. This requires that (i) semantic changes induced by a label intervention are coherently reflected in the regenerated copy, and (ii) causal dependencies among labels are respected during generation, preserving narrative consistency. We measure *Intervention Fidelity* through quantitative reconstruction loss ($MSE_X$) and *Causal Inference of Engagement* by out-of-sample counterfactual prediction accuracy ($MSE_Y$). Together, these criteria ensure that $Z_L$ functions as an actionable, causally disentangled proxy for marketer labels.

To understand whether interventions on $L$ are faithfully propagated through $Z_L$, we first conduct a diagnostic check as an illustration: we add the "dating-ready" label to the original label set and generate a new copy using the intervened $Z_L$. As illustrated in Figure 8(b), the resulting copy naturally incorporates dating-related messaging ("perfect for daily wear or a date—gives you that fresh, radiant look that boosts your charm") and delete the related scenario of "8 a.m. rush" while preserving the original functional claims. This coherent semantic shift demonstrates that $Z_L$ successfully translates discrete label interventions into contextually appropriate textual variations, thereby satisfying the requirement of Intervention Fidelity. In another 100 random cases, we conducted the same diagnostic test, the model achieves an IIA score of 96.7% (Li et al. 2026). Moreover, our model achieves a low marketing copy embedding reconstruction loss on the test set ($MSE_X$= 0.1103), indicating that the decoder conditioned on $Z_L$ and the auxiliary subspace $Z_{extra}$ produces marketing copies semantically aligned with the originals. As shown in Figure 8(a), the reconstructed copy retains core product claims such as suitability for combination-oily skin, balanced oil control and hydration, and all-day wearability. It omits brand-specific slang and emotional expressions including "another all-day face-sticker monster" and "devotee of Little Mix Cat". This is not a failure of reconstruction but an inherent design feature of $Z_L$. Under the posterior mean across multiple stochastic samples, $Z_L$ prioritizes generalizable, high-frequency semantic signals and filters out rare stylistic and non-structural noise. This property is consistent with standard VAE-based compressive representations (Kingma & Welling, 2013). The abstractive characteristic positions $Z_L$ as a causal proxy for strategic label choice, rather than a verbatim stylistic replicator.

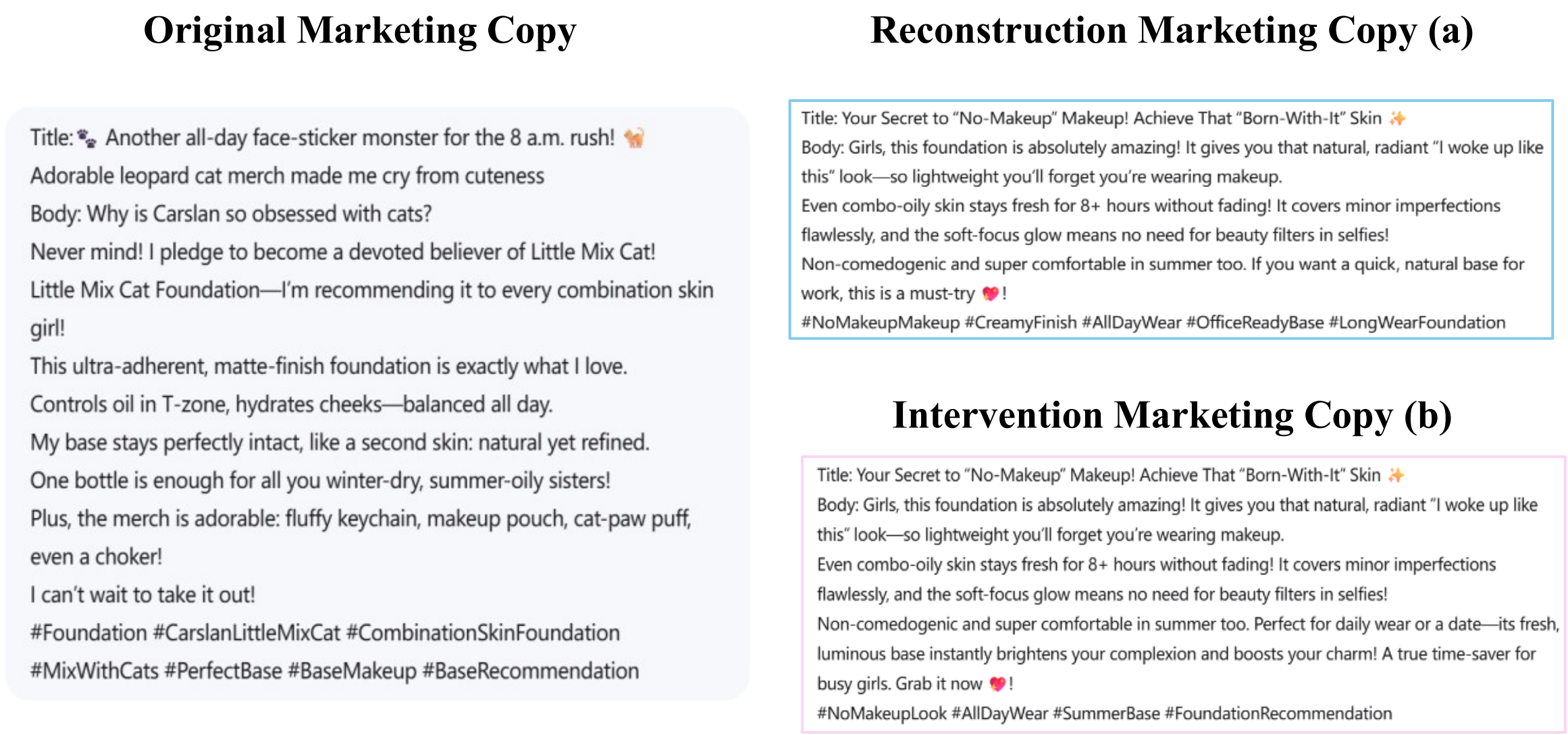

**Figure 8: Illustration of Reconstruction and Intervention**

Turning to *Causal Inference of Engagement*, our model achieves a low user engagement prediction loss on the test set ($MSE_Y = 0.0330$). To estimate ITE, we sample 50 times in $Z_L$ when intervening on certain labels to predict counterfactuals, then take the mean difference between these predictions and the factual value. In the diagnostic example (Fig.8), when the "dating-ready" label is introduced, the model predicts the following mean individual causal effect on user engagement: $\Delta Y = [3.6\%, 1.8\%, 3.4\%]$ for likes, comments and collects, respectively with all corresponding p-values$<$ 0.05. This pattern is behaviorally explainable: dating-themed content may increase both immediate social reactions, such as likes and comments, and utility-related actions, such as collecting, because it combines emotional appeal with occasion-specific relevance. Thus, the directional changes in $\hat{Y}$ reflect genuine user psychology rather than algorithmic artifacts. For space consideration, we only report the estimated individual treatment effect (ITE) for 10 labels (or label combinations) that are most impactful for our specific case study in Table 2.

**Table2. ITE in the "Carslan Little Mix Cat Foundation" Campaign**

| Label Intervention | likes | likes effect interval | comments | comments effect interval | collects | collects effect interval |
|---|---|---|---|---|---|---|
| dating-ready | 3.6% | [2.8%, 4.4%] | 1.8% | [1.4%, 2.2%] | 3.4% | [2.7%, 4.0%] |
| promotional campaigns | 6.7% | [5.8%, 7.6%] | 3.4% | [2.9%, 3.8%] | 6.1% | [5.3%, 6.9%] |
| sensitive skin | 6.6% | [5.7%, 7.5%] | 3.4% | [2.9%, 3.8%] | 6.1% | [5.2%, 7.0%] |
| coverage | 2.9% | [1.9%, 3.9%] | 1.4% | [0.8%, 1.8%] | 2.6% | [1.6%, 3.5%] |
| cost-effective | 1.3% | [0.4%, 2.1%] | 0.5% | [0.1%, 1.0%] | 1.3% | [0.5%, 2.1%] |
| business-occasion | -0.7% | [-1.5%, -0.1%] | -0.5% | [-1.0%, -0.1%] | -0.7% | [-1.5%, 0.0%] |
| co-branded campaign | -2.1% | [-3.2%, -1.0%] | -1.0% | [-1.6%, -0.5%] | -1.9% | [-2.9%, -0.9%] |
| co-branded campaign +new design | 9.5% | [8.6%, 10.3%] | 4.6% | [4.2%, 5.0%] | 8.6% | [7.8%, 9.3%] |
| cost-effective +dating-ready | 4.6% | [3.9%, 5.4%] | 2.3% | [1.9%, 2.7%] | 4.3% | [3.6%, 5.0%] |
| oil control +breathability +coverage | 4.0% | [3.4%, 4.6%] | 2.0% | [1.7%, 2.3%] | 3.8% | [3.2%, 4.4%] |

Note: All the p-values of ITE are <0.05. The intervals are 95% Confidence Intervals.

**5.2.3 D-VAE Performance: ATE Estimation**

To provide more generalizable insights into the causal contribution of individual marketing labels, we examine the estimated ATE for each label across 24,800 campaigns in terms of the three engagement dimensions. The ATE is calculated as a weighted average based on the ITEs in each campaign. Given the high heterogeneity of ITEs, we use a normal distribution as the weighting scheme. The results reveal distinct patterns of label efficacy (Table 3). Labels conveying broad product benefits or leveraging external social hooks consistently drive immediate social engagement. Specifically, "skin beautifying," "charity/CSR campaigns," and "promotional campaigns" show the strongest positive effects across the two positive engagement dimensions, likes and comments, with estimated total effects of +26.7%, +26.3%, and +23.9%, respectively, and all effects statistically significant ($p < 0.001$). These labels, however, exhibit a consistent trade-off: while they amplify short-term customer reactions, they are associated with negative or null effects on long-term utility-driven actions in terms of collects, suggesting their appeal is primarily emotional or transactional. Conversely, labels targeting specific demographics, such as "male audience" (–16.7%), "domestic brand" (–16.1%), show strong negative effects on collects (all $p < 0.001$), indicating that such positioning may be perceived as less worthy of future reference by users. This pattern underscores the importance of aligning label strategy with campaign objectives: promotional or cause-related labels excel at driving virality, whereas demographic-specific cues may sacrifice long-term engagement for short-term targeting precision.

**Table 3. ATE of Most Influential Labels Across Campaigns**

| Labels Intervention | likes | likes effect interval | comments | comments effect interval | collects | collects effect interval |
|---|---|---|---|---|---|---|
| skin beautifying | 18.4% | [18.0%, 18.8%] | 8.3% | [8.0%, 8.6%] | -4.0% | [-4.4%, -3.7%] |
| charity/CSR campaigns | 18.2% | [17.9%, 18.6%] | 8.1% | [7.8%, 8.5%] | -4.4% | [-4.8%, -4.0%] |
| promotional campaigns | 16.6% | [16.2%, 17.0%] | 7.3% | [7.0%, 7.6%] | -4.0% | [-4.4%, -3.7%] |
| male audience | 3.0% | [2.6%, 3.3%] | 0.3% | [0.1%, 0.7%] | -16.7% | [-17.0%, -16.3%] |
| domestic brand | 4.2% | [3.9%, 4.6%] | 1.2% | [0.8%, 1.5%] | -16.1% | [-16.5%, -15.8%] |
| natural gradient + coverage | 6.0% | [5.4%, 6.6%] | 3.8% | [3.3%, 4.4%] | 0.3% | [0.1%, 0.5%] |
| oil control + coverage | 1.4% | [0.8%, 1.9%] | 2.8% | [2.4%, 3.3%] | 1.0% | [0.5%, 1.5%] |
| cost-effective+ coverage | 1.1% | [0.5%, 1.6%] | 3.3% | [2.8%, 3.8%] | 0.8% | [0.2%, 1.4%] |
| oil control + oily skin + coverage | 4.6% | [4.2%, 5.0%] | 5.6% | [5.1%, 6.0%] | 1.3% | [0.8%, 1.8%] |
| dry skin + oily skin + coverage | 2.8% | [2.3%, 3.3%] | 4.5% | [4.0%, 4.9%] | 1.3% | [0.7%, 1.8%] |
| oil control + even skin tone + coverage | 1.2% | [0.7%, 1.7%] | 3.6% | [3.1%, 4.0%] | 1.0% | [0.5%, 1.5%] |

Note: "Most Influential Labels" are defined based on the absolute magnitude of the estimated coefficients. All the p-values of ATE are <0.001. The intervals are 95% Confidence Intervals.

Beyond marginal ATE of single labels, our analysis of label combinations reveals critical non-linear interactions that significantly amplify engagement effects. Crucially, while individual labels are constrained by an inherent trade-off that consistently yields negative or null effects on long-term collects, strategically designed combinations break this limitation, transforming isolated drawbacks into universally positive gains across all engagement dimensions. Contrary to the intuition that frequently co-occurring labels are optimal, we find that the highest-value combinations are frequently underutilized in practice, suggesting substantial room for efficiency gains in expert label selection strategies. Among the top 150 frequent two-label combinations, the highest-value pairs are strikingly underused. The most promising combination is [natural gradient +coverage] (frequency rank 139), which delivers positive effects across all engagement metrics: 6.0% likes, 3.8% comments, and a modest but statistically significant 0.3% lift in collects. Two other coverage-centric pairs also demonstrate Pareto-improving patterns: [oil control + coverage] (rank 38) yields 1.4% likes and 2.8% comments with a 1.0% collects lift, while [cost-effective + coverage] (rank 53) achieves 1.1% likes, 3.3% comments, and 0.8% collects. Notably, all three combinations share "coverage" as a core anchor, indicating that functional benefit framing may serve as a robust engagement multiplier when paired with complementary attributes.

This pattern is more intriguing in three-label combinations. The most impactful triplets are not those most frequently used, but rather those that precisely fuse core efficacy with a targeted persona—a combination largely overlooked in current practice. The highest-performing combination is [oil control + oily skin + coverage], which achieves the strongest engagement lift of 11.5%, driven by robust gains in likes (+4.6%) and comments (+5.6%). Despite its superior performance, this synergistic trio ranks only 133rd in historical usage frequency. [dry skin + oily skin + coverage] (rank 61) yields a notable 8.6% lift, directly addressing the nuanced needs of combination skin users who experience both dryness and oiliness. Even more revealing is [oil control + even skin tone + coverage] (rank 77), which achieves a total effect of +5.8%, highlighting the value of emphasizing product feel and wearability alongside functional benefits. These results underscore that the most effective labeling strategies may not be the most common ones, but those that address specific user needs with precision.

### 5.3 Ablation Analysis

Here, we demonstrate the contribution of each of the main components in D-VAE through an ablation analysis. We first compare the full model against C-VAE as our benchmark since it also supports SCM-based counterfactual interventions. We do not consider purely predictive yet non-causal machine learning baselines, as they are not designed to support valid counterfactual reasoning. Compared to the chosen benchmark, D-VAE achieves a reconstruction loss of 0.1103, while C-VAE yields significantly higher reconstruction loss (0.1214, difference =9.2%, p= 0.0002). In terms of prediction losses, D-VAE achieves 0.0330, while C-VAE yields a higher value of 0.0339 (difference = 2.7%, p = 0.0292).

**Table 4. Ablation study**

| | Intervention Fidelity | Causal Inference of Engagement | $MSE_X$ | $MSE_Y$ | $MSE_Y$ PER DIMENSION |
|---|---|---|---|---|---|
| D-VAE (Full) | 96.7% | Yes | 0.1103 | 0.0330 | [0.0281,0.0365, 0.0345] |
| C-VAE (w/o deconfounder Module) | 83.5% | No | 0.1214 | 0.0339 | [0.0296,0.0380, 0.0344] |
| w/o $Z_{extra}$ Module | 51.7% | No | 0.1077 | 0.0381 | [0.0323, 0.0432, 0.0387] |
| w/o SCM Module | 20.3% | No | 0.0920 | 0.0302 | [0.0265,0.0331, 0.0309] |
| w/o Inductive Bias Module | 48.4% | No | 0.1133 | 0.0353 | [0.0295,0.0404, 0.0359] |

Note: Each row removes one module from the full model at a time. Intervention Fidelity is quantified via IIA Score across 100 samples, which measures (i) success rate on the intervened label and (ii) preservation of non-intervened attributes subject to compatibility constraints with the intervened label. $MSE_X$ and $MSE_Y$ are reported as auxiliary diagnostics.

We then investigate the incremental contribution of other architectural components, with results summarized in Table 4. The full model achieves high intervention fidelity (96.7%), demonstrating precise semantic control during label interventions. Ablating the auxiliary subspace $Z_{extra}$ reduces fidelity to 51.7%. Intriguingly, while this variant yields a marginally lower $MSE_X$ (−2.4%), the fidelity drops suggest that there are uncontrolled variations brought by LLM and environments leaking into the causal subspace $Z_L$, which distorts intervention boundaries. Removing the SCM module causes a catastrophic collapse in fidelity (20.3%). Although it achieves the lowest $MSE_X$ (−16.6% relative to full), the generated text becomes logically incoherent: for example, activating the “dating-ready” label inserts romantic cues while retaining contradictory “early-morning student” context (e.g., “8 a.m. rush”) in the Carslan Campaign (Figure8). Without the SCM component, label dimensions are modeled independently, preventing semantically coherent propagation and breaking narrative consistency. Finally, ablating the Inductive Bias module reduces fidelity to 48.4%. The absence of counterfactual forward debiasing leaves residual narrative traces from the original copy (e.g., “budget-conscious” phrasing persists under “premium” interventions).

Regarding *Causal Inference of Engagement*, valid out-of-sample counterfactual prediction requires both faithful interventions and a disentangled latent causal structure. Only the full model supports reliable engagement inference. Removing $Z_{extra}$ increases $MSE_Y$ by 15.5% and the Inductive Bias ablation raises $MSE_Y$ by 7.0%, reflecting biased engagement estimates due to unremoved confounding traces. Notably, the SCM-ablated variant achieves the lowest $MSE_Y$ but exhibits a markedly elevated KL divergence (1.97 vs. ≈0 for others), indicating a latent posterior that severely diverges from the confounder-conditioned prior. This suggests that the model likely memorizes spurious surface co-occurrences rather than learning the underlying causal mechanism, rendering its low prediction error invalid for counterfactual scenarios. The full model's balanced $MSE_Y$ (0.0330) and stable per-dimension errors ([0.0281, 0.0365, 0.0345]) demonstrate robust, structurally sound engagement inference. Here, $MSE_Y$ and posterior diagnostics act as auxiliary validators: they yield a low value when corroborating causal validity, but a low value alone cannot certify causal validity when structural constraints are violated, making it a necessary but not sufficient indicator.

Collectively, these results confirm that $Z_L$ functions as an interpretable and causally disentangled embedding of $L$. It not only reconstructs and intervenes on copy with semantic fidelity but also yields engagement predictions grounded in true label-driven mechanisms. This dual effectiveness enables direct optimization of label selection for desired outcomes without requiring observing counterfactuals. In the next section, we evaluate whether these counterfactual insights can empower non-expert marketers to make data-informed copywriting decisions.

**5.4 Direct Evidence from Controlled Experiment**

To evaluate how the counterfactual predictions generated by D-VAE can effectively support real-world marketing decisions, we conduct a small-scale controlled experiment to assess whether D-VAE recommended label configurations yield higher user engagement compared to expert-driven optimization, C-VAE based intervention, and the unmodified original copy as the baseline control. We randomly select 30 foundation-related marketing copies from our dataset; each originally annotated with 15–20 expert labels. For each copy, we allow the addition or removal of up to 5 labels (approximately 25%–33% of the initial set), a range that provides sufficient flexibility to explore multi-level label combinations while preserving thematic coherence. The total number of labels after modification is capped at 25 to avoid logical inconsistencies (e.g., simultaneously selecting mutually exclusive labels such as "premium pricing" and "high value-for-money").

A panel of five senior marketing experts with at least 3 years of industry experience manually optimizes the label set for each post. In parallel, D-VAE and C-VAE automatically select label configurations that maximize predicted counterfactual engagement under identical constraints (detailed in Appendix C Algorithm 4). To isolate the effect of label selection from generative model variability, all four versions: original (control), expert-modified, D-VAE-modified, and C-VAE-modified are generated using the same LLM (Qwen-3) conditioned on their respective label sets. To validate that label interventions produced semantically coherent shifts in ad copy, we conducted an automated manipulation check using a different LLM (GPT-5.4). For each intervened ad, the model scored (1–10) how well the modified copy reflected the intended label changes—specifically, emphasizing added labels and de-emphasizing removed ones. Across all 90 (3×30) manipulated creatives, the average alignment score was 8.5 (SD = 0.78), with 97.8% of cases scoring ≥7. This confirms that our interventions successfully induced the targeted semantic shifts in generated contents.

We randomly recruited 120 participants from the beauty-focused RED users. The sample reflects the distribution of RED users with a majority identifying as female and exhibiting diverse makeup needs (e.g., "Learn Makeup," "Date Makeup," "Anti-aging Skincare"). Detailed demographic distributions are provided in Appendix E. We then randomly assigned them to one of the four experimental conditions to decide whether to like, comment, or collect it. We use the aggregated engagement score (sum of the three actions) per post per condition as our unit of analysis.

As shown in Figure 9, both the expert and D-VAE conditions achieve higher engagement than the control and C-VAE conditions. To formally test this pattern, we perform a two-way ANOVA without replication, treating version type (4 levels) and original copy ID (30 groups) as fixed factors. Results (Table 5) reveal a highly significant main effect of version type ( $p < 0.001$), confirming that the choice of label intervention strategy has a significant impact on user engagement. The effect of copy ID is also statistically significant ( $p < 0.001$ ), reflecting expected heterogeneity in baseline appeal across different ad creatives. Importantly, the significance of version type persists even after accounting for this creative-level variation, supporting the conclusion that label intervention consistently influences engagement outcomes. Post-hoc pairwise comparisons (Table 6) with Bonferroni correction ($\alpha =$ 0.0125) further show that D-VAE performs equivalently well as human experts do ($p = 0.9332$), and both significantly outperform the control ($p = 0.0053$ for D-VAE; $p = 0.0097$ for Expert). In contrast, C-VAE shows no significant improvement over the unmodified baseline ($p = 0.3081$).

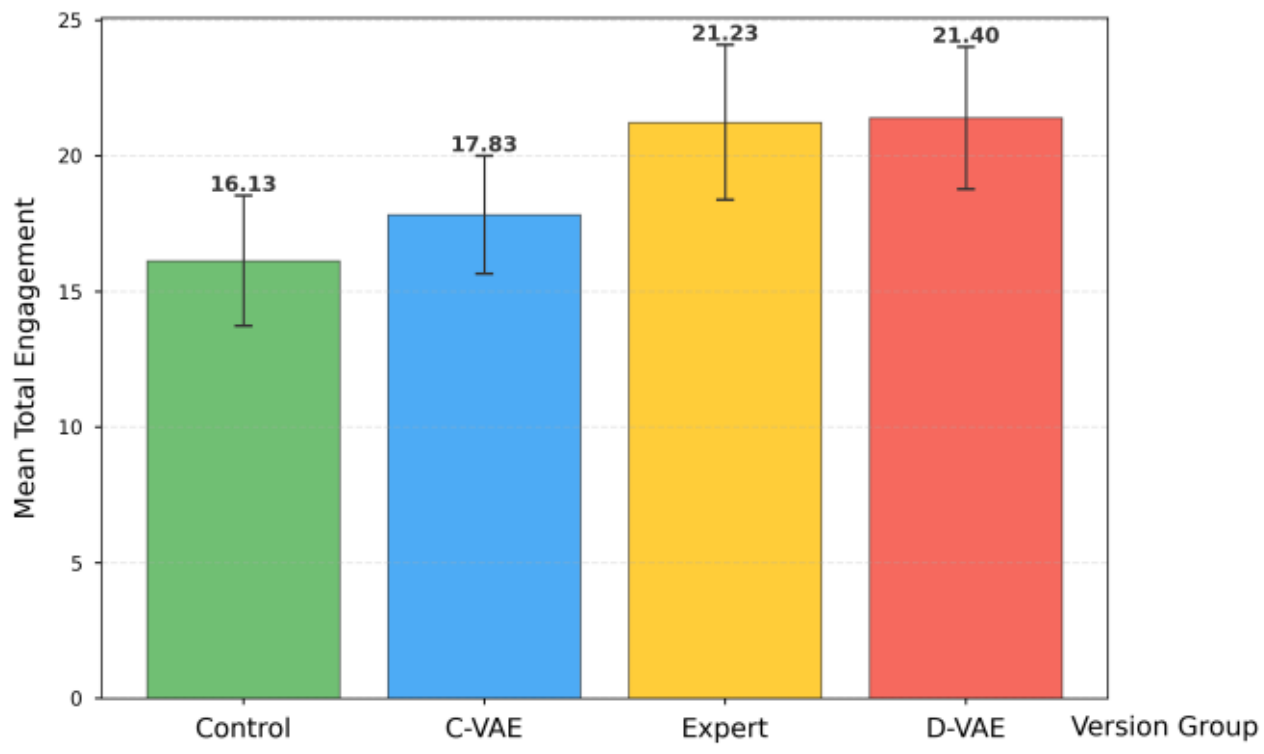

**Figure 9. The Mean Total Engagement by Version Group (with 95% CI)**

These findings demonstrate that the counterfactual predictions produced by D-VAE translate directly into measurable real-world impact: labels recommended by the model enable non-expert users to generate marketing creatives that achieve engagement statistically indistinguishable from those crafted by seasoned professionals. This validates that expert marketing skills can be successfully digitized and replicated.

**Table 5. Results of Two-way ANOVA Without Replication**

| Source | Sum of Squares | df | F-value | Pr(>F) |
|---|---|---|---|---|
| C(version) | 607.1 | 3 | 6.448103 | 0.000541 |
| C(copy_id) | 3057.8 | 29 | 3.359728 | 0.000007 |
| Residual | 2730.4 | 87 | - | - |

**Table 6. Results of Independent Samples T-tests (Bonferroni-corrected)**

| Comparison Group | t-value | df | Two-tailed p-value | Bonferroni-corrected α | Significance |
|---|---|---|---|---|---|
| Expert vs D-VAE | -0.0842 | 58 | 0.9332 | 0.0125 | no |
| Expert vs Control | 2.6765 | 58 | 0.0097 | 0.0125 | yes |
| D-VAE vs Control | 2.8994 | 58 | 0.0053 | 0.0125 | yes |
| C-VAE vs Control | 1.0283 | 58 | 0.3081 | 0.0125 | no |

The inferior performance of the C-VAE baseline stems from its inability to adjust for unobserved confounders. A concrete example is the "Carslan Little Mix Cat Foundation" campaign (Fig. 8). The original high-performing labels emphasized suitability for combination skin, oil control, skin-nourishing benefits, and co-branded design. Our D-VAE correctly identifies efficacy-related labels (e.g., "sensitive-skin friendly," "natural ingredients," and "buildable coverage") as the true drivers of engagement. In contrast, C-VAE spuriously recommends "business occasion" and "high-income audience", labels that co-occur with premium brands' high-budget professional campaigns on RED, creating a confounding correlation overlooked by C-VAE (Fig. 4(b)). For a mass-market product like Carslan, such mismatched positioning produces inauthentic messaging and lower engagement. By incorporating a deconfounder that captures latent shared factors (e.g., brand tier and KOL expertise), D-VAE successfully disentangles causal signals from platform-induced noise, yielding both accurate and operationally interpretable recommendations.

## 6. Simulation Study

Having established D-VAE's empirical credibility in a large-scale, real-world marketing application and further demonstrated its practical value through a controlled field experiment, we turn to two remaining questions. First, how accurately can D-VAE recover causal effects when the ground truth is known? Second, how does the framework generalize beyond the setting of marketing-label attribution?

These questions are difficult to answer with observational data because true causal effects and latent confounders are inherently unobserved. Real-world validation must therefore rely on indirect evidence, such as counterfactual prediction, intervention consistency, and application-specific diagnostics. A simulation study overcomes this limitation by allowing us to specify both the ground-truth causal effects as well as the level and structure of unobserved confounding. It therefore provides a direct and systematic test of causal-effect recovery.

We design the simulation around a general class of GenAI-mediated settings which we instantiate in a labor-market setting: job candidates select keywords to guide an LLM in generating professional biographies intended to attract recruiters' interest. By varying the form and intensity of shared and attribute-specific confounding, we evaluate D-VAE's causal-recovery accuracy, robustness, and boundary conditions.

We begin by formalizing the general class of problems to which D-VAE applies.

**Table 7. Generic Data Structure and Notation**

| Notation | Status | Description |
|---|---|---|
| $L_i$ | Observed | A $K$-dimensional vector of actionable attributes or skill primitives for unit $i$ |
| $X_i$ | Observed | A high-dimensional unstructured artifact generated by an LLM |
| $Y_i$ | Observed | The downstream outcome |
| $C_i$ | Observed/Optional | Pre-treatment covariates that may be observed for adjustment |
| $e_{m,i}$ | Latent | A shared confounder affecting multiple dimensions of $L_i$ and $Y_i$ |
| $e_{s,i,j}$ | Latent | A single-treatment confounder affecting a focal attribute $L_{ij}$ and $Y_i$ |
| $R_i = h_\phi(X_i)$ | Learned | A representation of semantic and stylistic information contained in the artifact |
| $Y_i(\ell)$ | Counterfactual | The outcome that would be observed under attribute configuration $L_i = \ell$ |

## 6.1 Generic Problem Formulation for GenAI-Mediated Skill Valuation

Many GenAI-augmented workflows share a common causal-attribution problem. A decision selects multiple actionable skill primitives, and these inputs steer LLM to generate a high-dimensional artifact that drives a performance outcome. The empirical objective is not merely to predict performance, but to determine which individual attributes (skills) and attribute configurations causally improve it. The marketing-label application examined in this study represents one practical instance of this broader problem.

Consider $N$ observational units indexed by $i = 1, \ldots, N$ (see **Table 7** for key notations and descriptions). For each unit, the analysts observe

$$\mathcal{D}_N = \{(L_i, X_i, Y_i, C_i)\}_{i=1}^N, \qquad L_i = (L_{i1}, \dots, L_{iK})^\top \in \mathcal{L},$$

where $L_i$ is a $K$-dimensional vector of attributes, $X_i$ is the artifact generated by the LLM, $Y_i$ is the downstream outcome, and $C_i$ contains observed pre-treatment covariates. The focal causal pathway is $L_i \to X_i \to Y_i$: the selected attributes shape the generated artifact, which in turn influences performance.

Attribute assignment is generally not randomized. We distinguish a shared confounder $e_{m,i}$, which can affect several attributes jointly, from an attribute-specific confounder $e_{s,ij}$, which affects a focal attribute and the outcome without necessarily inducing dependence across attributes. A generic structural system is

$$L_{ij} = g_j\left(e_{m,i}, e_{s,ij}, \varepsilon_{ij}^L\right), \qquad j = 1, \dots, K.$$

$$X_i = g_X\left(L_i, C_i, e_{m,i}, \varepsilon_i^X\right), \qquad Y_i = g_Y\left(L_i, X_i, C_i, e_{m,i}, e_{s,i}, \varepsilon_i^Y\right). \tag{29}$$

where $e_{s,i} = \left(e_{s,i1}, \dots, e_{s,iK}\right)^\top$. The joint assignment pattern of $L_i$ may contain information about $e_{m,i}$, whereas an attribute-specific confounder need not leave a comparable cross-attribute signature. In addition, $X_i$ is a stochastic, high-dimensional mediator whose semantic and stylistic variation may affect $Y_i$ beyond what is encoded directly in $L_i$. The learned representation $R_i = h_\phi(X_i)$ is therefore used to summarize mediator-side information relevant to counterfactual prediction.

Let $Y_i(\ell)$ denote the potential outcome under the joint intervention $L_i = \ell$. For attribute $j$, the average causal effect of changing its value from $a'$ to $a$, while averaging over a reference distribution $Q_{-j}$ of the remaining attributes, is

$$\tau_j(a, a') = \mathbb{E}_{L_{i,-j} \sim Q_{-j}}\left[Y_i\left(a, L_{i,-j}\right) - Y_i\left(a', L_{i,-j}\right)\right].$$

For a binary attribute, the conventional contrast is $\tau_j = \tau_j(1,0)$. More generally, the causal effect of replacing one complete attribute configuration $\ell'$ with another configuration $\ell$ is

$$\tau(\ell, \ell') = \mathbb{E}[Y_i(\ell) - Y_i(\ell')].$$

Both estimands are total effects: they allow interventions on $L_i$ to propagate through the generated artifact $X_i$ before affecting $Y_i$. The estimation objective is to recover these effects despite three identification challenges: latent confounding shared across multiple attributes, confounding specific to a single attribute, and high-dimensional stochastic mediation through the LLM-generated artifact.

D-VAE addresses these challenges by combining a multi-treatment deconfounder with causal representation learning. The deconfounder uses joint attribute-assignment patterns to recover shared latent structure, while the representation model captures intervention-relevant variation in $X_i$. Because single-treatment confounders cannot generally be recovered from cross-attribute dependence alone, D-VAE's theoretical analysis characterizes the remaining omitted-variable bias through PAC-based bounds rather than assuming it away. In empirical applications, where true causal effects are unobserved, causal fidelity is evaluated indirectly through intervention consistency and predictive diagnostics, including *IIA*, $MSE_X$ , and $MSE_Y$ . In the simulation, the data-generating process supplies the ground-truth effects, allowing direct evaluation through bias, RMSE, and confidence-interval coverage.

## 6.2 Simulation Design: LLM-Generated Political Candidate Biographies

To evaluate D-VAE against known causal effects and outside the marketing application, we adapt the candidate-biography simulation of Imai and Nakamura (2025), which builds on the political candidate profile experiment of Fong and Grimmer (2016). In that setting, survey respondents evaluate biographies of political candidates, and the focal textual feature is a candidate's military background. Imai and Nakamura use Llama 3–8B to generate 4,000 biographies of hypothetical politicians, code military background using the keywords such as "military," "veteran," and "army," and represent confounding features using a politics-and-education topic and TextBlob sentiment. We use the same 4,000-biography corpus as a realistic textual foundation, but extend the design to a multi-attribute treatment space and a fully specified data-generating process with known causal effects. The biographies retain semantic, stylistic, and narrative variation beyond the coded attributes, thereby preserving the high-dimensional mediator-side complexity that motivates D-VAE.

For biography $i$, let $L_i = (L_{i1}, L_{i2}, L_{i3}, L_{i4}, L_{i5})^\top \in \{0,1\}$ denote five binary attributes. We set $L_{ij} = 1$ when the biography contains at least one term from the dictionary for attribute $j$, and $L_{ij} = 0$ otherwise. The attributes capture elite education ($L_{i1}$; e.g., Ivy League, honors degree, law degree), political experience ($L_{i2}$; e.g., congress, senate, mayor), military experience ($L_{i3}$; military, veteran, army), business and economic experience ($L_{i4}$; e.g., business, budget, finance), and public-service orientation ($L_{i5}$; e.g., healthcare, education reform, community). The military dictionary reproduces the coding used by Imai and Nakamura (2025); the remaining four attributes are study-specific extensions informed by prior research on candidate competence, experience, and public orientation. We inspect marginal frequencies and pairwise co-occurrence before estimation to rule out near-degenerate labels and severe overlap problems.

We next construct shared and attribute-specific confounders that are known to the simulation but withheld from the estimators. The shared multi-treatment confounder $e_{m,i} = \left(e_{m1,i}, e_{m2,i}, e_{m3,i}\right)^\top$ captures social capital, seniority, and civic orientation. These components are proxied by biography-level cues such as prominent family or high-status careers, biography length or number of offices held, and community-service or public-service language. The single-treatment confounder $e_{s,i}$ targets the military attribute $L_{i3}$ and captures national-security salience. We first compute a security score $S_i^{\text{sec}}$ from terms such as “national security,” “public safety,” and “law enforcement,” which do not overlap with the military-label dictionary. We then orthogonalize this score with respect to the remaining labels and the shared confounder:

$$e_{s,i} = S_i^{\text{sec}} - \widehat{\mathbb{E}}\left[S_i^{\text{sec}} \mid L_{i,-3}, e_{m,i}\right].$$

This residualization isolates security-related semantic variation associated specifically with $L_{i3}$

rather than with general label co-occurrence. Finally, define $M_i = \text{sentiment}(X_i) \in [-1,1]$ as the TextBlob polarity score of biography $X_i$. Imai and Nakamura (2025) use TextBlob sentiment as a confounding feature in their simulation. We retain the same text summary but assign it a distinct role in our data-generating process: $M_i$ captures outcome-relevant mediator-side variation embedded in the generated biography and is not supplied to the estimators as a structured control.

The synthetic candidate-evaluation score is generated by the following DGP:

$$Y_i = \beta^\top L_i + L_i^\top B e_{m,i} + \Gamma^\top e_{m,i} + \alpha_m M_i + \Theta e_{s,i} + \epsilon_i \tag{29}$$

The disturbance satisfies $\epsilon_i \sim N(0, 0.5^2)$. The vector $\beta = (0.3, 0.2, 0.35, 0.15, 0.25)$ contains the baseline attribute effects, $\Gamma = (0.3, -0.20, 0.25)$ governs the effect of the shared confounding, $B$ allows each attribute effect to vary with the shared confounder, and $\alpha_m = 0.20$ controls the contribution of biography sentiment. The interaction matrix is

$$B = \begin{pmatrix} 0.12 & 0.04 & 0.02 \\ 0.03 & 0.10 & 0.08 \\ 0.02 & 0.06 & 0.05 \\ 0.08 & 0.08 & 0.04 \\ 0.05 & 0.03 & 0.09 \end{pmatrix}.$$

The parameters are calibrated to reflect the core structure in Equation (29): the direct effects of $L_i$ and the shared confounder $e_m$ are the main determinants of $Y_i$, because they represent the target causal signal and the central identification challenge. Treatment-effect heterogeneity and mediator-side variation play smaller but meaningful roles, preserving the $L_i \to X_i \to Y_i$ mechanism without allowing these components to dominate, while the disturbance term prevents the outcome from becoming nearly deterministic. Under this calibration, differences in performance across estimators primarily reflect their ability to recover the effects of $L_i$ while accounting for shared confounding, mediator-side variation, and treatment heterogeneity, rather than sensitivity to the exact numerical values of the coefficients.

Because the structural model is known, causal fidelity can be evaluated against the exact sample-average effect. Holding $L_{i,-j}, e_{m,i}, e_{s,i}, M_i$, and $\varepsilon_i$ fixed, define the unit-level effect of attribute $j$ as

$$\Delta_{ij} = Y_i(1, L_{i,-j}) - Y_i(0, L_{i,-j}).$$

The corresponding ground-truth sample-average effect is

$$\tau_j^{\star} = \frac{1}{n}\sum_{i=1}^{n} \Delta_{ij} = \beta_j + \frac{1}{n}\sum_{i=1}^{n} B_{j\cdot}\, e_{m,i}.$$

Thus, the simulation provides an exact benchmark against which the estimated attribute effects can be evaluated, while retaining heterogeneous effects through the interaction between $L_i$ and $e_{m,i}$.

We evaluate four confounding scenarios (Table 8), holding all other parameters fixed and varying only the outcome sensitivity $\Theta$ to the military-specific confounder. Let $\| \Gamma \|_2$ denote the Euclidean magnitude of the shared-confounder coefficient vector.

**Table 8: Four Scenarios of Different Confounding Structure**

| Scenario | Single-confounder strength | Interpretation |
|---|---|---|
| Shared only | $\Theta = 0$ | Only confounding shared across multiple attributes affects the outcome. |
| Weak single | $0 < \Theta < \| \Gamma \|_2$ | Introduces a modest confounder specific to the military attribute. |
| Bounded single | $\Theta = \| \Gamma \|_2$ | Sets single-treatment confounding equal to the magnitude of shared confounding. |
| Strong single | $\Theta = 2 \| \Gamma \|_2$ | Stress-tests D-VAE under dominant single-treatment confounding. |

This sequence progressively weakens the multi-treatment deconfounder condition. It therefore tests whether D-VAE can recover attribute-level causal effects when confounding is shared across treatments, when a focal attribute has additional idiosyncratic confounding, and when that single-treatment confounding becomes the dominant source of bias.

### 6.3 Estimators and Evaluation Metrics

For each biography, we extract an internal representation $R_i$ from the hidden states of Qwen-3. D-VAE and Gen-AI Powered Inference (GPI) then use these representations directly. For the conventional estimators, we additionally construct BERT embeddings as text-level controls (Devlin et al., 2019). We compare D-VAE with three established benchmarks: ordinary least squares (OLS), which provides a transparent linear estimate of average attribute effects; Double Machine Learning (DML), which uses flexible nuisance models and orthogonization to reduce bias from observed covariates (Chernozhukov et al., 2018); and Causal Forest (CF), which accommodates heterogenous treatment effects through tree-based estimation and sample splitting. For each conventional estimators, we evaluate both a baseline specification and a BERT-augmented specification. GPI is the closest GenAI-based benchmark because it similarly leverages LLM-derived text representations for causal inference (Imai and Nakamura, 2025).

All estimators are evaluated using identical simulated-samples and use the same 70%/30% train–test split within each replication. The biography corpus, attribute labels, confounders, and ground-truth effects are held fixed across methods, while outcome noise is regenerated for each Monte Carlo seed. Each estimator is then refitted and used to recover the five attribute-specific average treatment effects. We conduct 50 Monte Carlo replications for each confounding scenario.

Performance is evaluated using mean absolute bias, RMSE, and the empirical coverage of bootstrap 95% confidence intervals. Mean absolute bias captures systematic deviations from the ground-truth effects, whereas RMSE captures both bias and variation across simulation replications and places greater weight on large errors. Lower values of both metrics indicate more accurate causal effect recovery. Coverage measures the proportion of confidence intervals that contains the corresponding true effects, with values near 95% indicating well-calibrated uncertainty. Together, these metrics assess both point-estimation accuracy and inferential reliability.

### 6.4 Simulation Results

Figure 10 reports the results. Across all four confounding scenarios, D-VAE achieves the lowest mean absolute bias, RMSE, and among the best for coverage confidence intervals, demonstrating consistently stronger recovery of the ground-truth attribute effects. Conventional estimators that rely only on structured covariates—OLS, DML, and CF—perform substantially worse in this text-mediated setting. Adding BERT embeddings improves their performances, reducing average absolute bias by 36.3%, 33.3%, and 26.5% respectively. Nevertheless, all BERT-augmented estimators remain less accurate than D-VAE. Although BERT-augmented CF attains relatively high confidence-interval coverage in some scenarios, its substantially larger bias and RMSE indicate that coverage alone does not translate into accurate effect recovery.

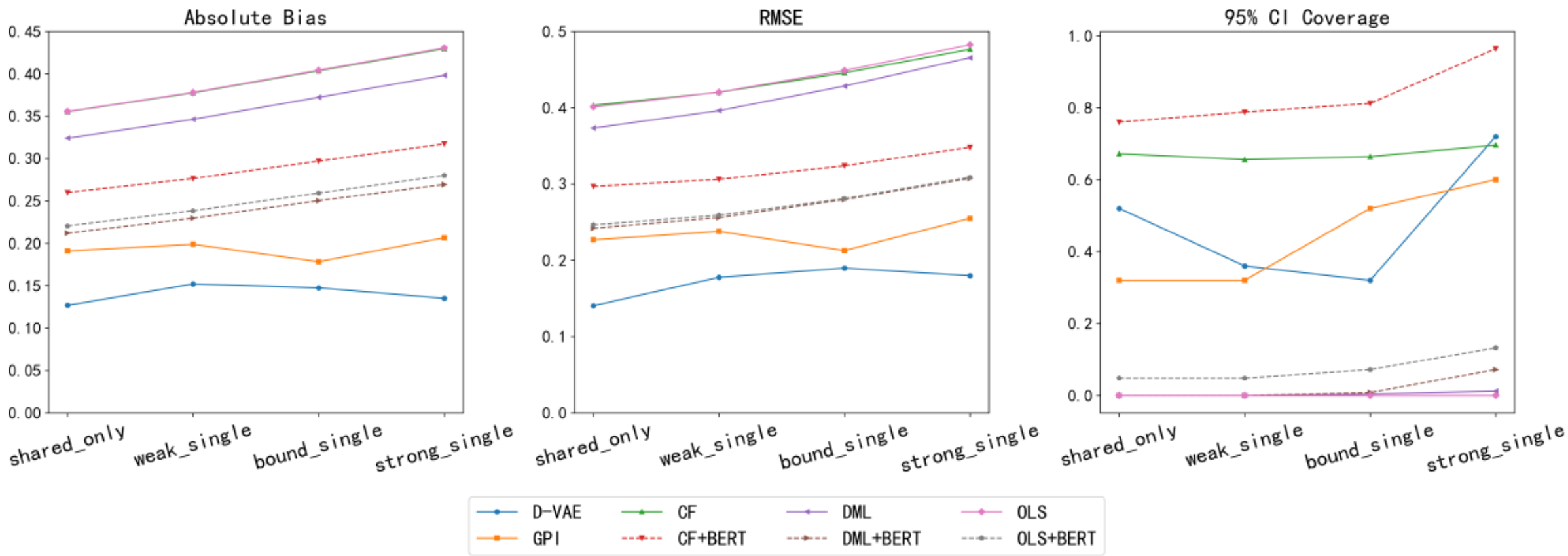

**Figure 10. Model Performance of Average Label Effect Inference Across Scenarios**

D-VAE also outperforms GPI, the closest GenAI-based benchmark. Relative to GPI, D-VAE reduces absolute bias by 33.6%, 23.6%, 17.3%, and 34.6% in the shared-only, weak-single, bounded-single, and strong-single scenarios, respectively. The corresponding RMSE reductions are 38.1%, 25.4%, 10.8%, and 29.5%. D-VAE's bootstrap 95% confidence-interval coverage is 52%, 36%, 32%, and 72% across the four scenarios, compared with 32%, 32%, 52%, and 60% for GPI. Thus, D-VAE provides more accurate point estimates in every scenario and achieves comparable or better interval coverage in three of the four scenarios. However, coverage remains below the nominal 95% level for both methods, indicating that uncertainty calibration remains challenging even when point-estimation accuracy improves.

The attribute-level results clarify the source of D-VAE's advantage (Figure 11). D-VAE performs particularly well for elite education ($L_1$), business and economic experience ($L_4$), and public-service orientation ($L_5$). These attributes benefit most from modeling the joint assignment structure, suggesting that their confounding is largely shared across attributes. GPI is occasionally competitive for political experience ($L_2$), whose effect appears more readily recoverable from the LLM representation alone. Military experience ($L_3$) is the most difficult attribute for all estimators because the simulation deliberately assigns its focal confounder only to that treatment. Such single-treatment confounding may not generate cross-attribute dependence and therefore cannot be fully recovered from treatment co-occurrence. D-VAE nevertheless remains comparatively robust when this confounding is bounded.

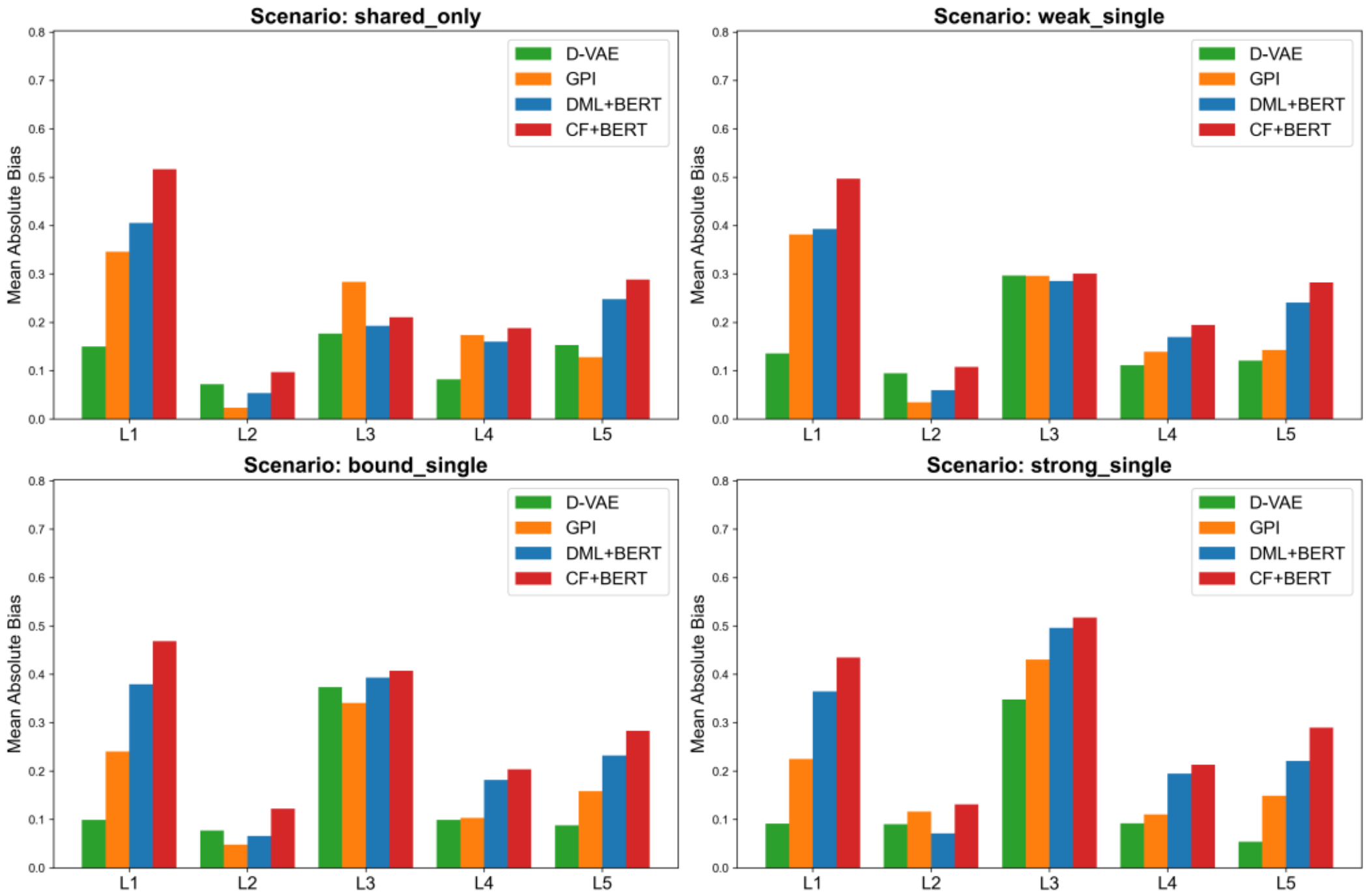

**Figure 11. Model Performance of Individual Label Effect Inference**

Overall, the simulation provides a direct ground-truth assessment of causal fidelity and demonstrates that D-VAE generalizes beyond marketing-label attribution. Its advantage is strongest when shared confounding remains informative, but narrows as single-treatment confounding becomes dominant. These findings both support D-VAE's applicability to high-dimensional, LLM-mediated, multi-attribute settings and identify an important boundary condition: the framework is most effective when the joint treatment structure contains sufficient information about the dominant latent confounders.

## 7. Discussion

We introduce a deconfounding framework that attributes engagement outcomes to labels, shifting the analytical focus from evaluating AI-generated products to upstream skill assessment. In modern AIGC workflows, practitioners no longer control final artifacts directly, instead they intervene via structured inputs such as labels as explicit skill primitives. However, a concomitant challenge arises: identical skill configurations can yield stochastic outputs, and skill value attribution is systematically confounded by unobserved confounds such as firm budget constraints, LLM capability and platform matching algorithms. Causal inference that determines the causal impact of skill on final performance is thus essential: it needs to isolate the true actionable drivers of downstream value from spurious correlations and mediator noise.

Methodologically, D-VAE contributes by adapting deconfounding and causal representation learning to a new causal object: upstream skill configurations in stochastic LLM-mediated production. Standard deconfounder methods use treatment co-occurrence to recover shared latent confounders, but they do not address settings where treatment effects are mediated through generated content or where label-specific confounders may still bias attribution. D-VAE accommodates this logic by combining multi-label confounder recovery with an explicit bound on omitted bias from single-treatment confounders in nonlinear situations. VAE variants, by contrast, learn structured latent factors mainly for disentangled generation or sample-level interventions. D-VAE repurposes the latent space for causal skill valuation: it represents expert label-selection logic, separates label-driven semantic variation from auxiliary environmental and mediator-side noise, and predicts downstream engagement under label interventions. The resulting framework attributes performance to actionable label configurations rather than to incidental wording, platform selection, or historical co-occurrence. This makes D-VAE a unified framework for valuing AI-augmented skills, rather than a direct application of existing causal or generative models.

Empirically, D-VAE identifies valuable yet underutilized skill configurations, showing how latent expertise can be translated into economically meaningful interventions. Our simulation study extends this evidence to a broader class of multi-attribute, LLM-mediated attribution problems and reveals how the framework's performance varies as confounding shifts from shared to attribute-specific sources. A controlled experiment provides complementary evidence of practical impact: D-VAE-recommended label configurations increased total engagement by 32.7% relative to the original campaigns and performed comparably to configurations developed by senior experts. Taken together, these findings show that D-VAE can transform historical expert decisions into scalable, auditable guidance for selecting and combining AI-augmented skills. This capability enables organizations to move beyond ad hoc prompt engineering, costly trial-and-error, and exhaustive live experimentation toward systematic causal evaluation and optimization of skill portfolios.

Our work carries three key implications. First, it establishes a skill-to-product value attribution paradigm, resolving the identification gap in stochastic LLM mediation and providing a generalizable structure for decomposing aggregate outcomes into constituent skill contributions. Second, by formalizing expert labeling logic as a causally disentangled latent representation, we demonstrate that tacit know-hows can be systematically extracted and operationalized—democratizing high-quality AI-augmented creation. Third, our deconfounded counterfactual framework provides a methodological foundation for pricing and optimizing AI-augmented skill portfolios. Managers can now simulate what-if skill interventions, identify synergistic combinations, and allocate creative resources based on causal ROI rather than historical frequency, turning generative AI into an insight-driven co-pilot.

We acknowledge three limitations. First, our implementation focuses on a single product category (complexion cosmetics); marketing label systems emphasizing formulation and sensory cues may not directly transfer to domains with distinct attribute structures. Future work could develop cross-domain skill representations that capture shared structural mappings between strategic inputs and consumer response. Second, we assume a static label ontology, whereas expert knowledge and market trends evolve continuously. Extending this framework to dynamic causal models would allow firms to track how the marginal value of specific skills shifts with competitive activity, platform algorithms, or seasonal demand. Finally, our framework estimates the first-stage content-response value of label configurations under routine deployment conditions. This is the decision layer at which firms first need causal evidence before reallocating budget, channels, or KOL resources, they must know which label combinations create engagement value within comparable creative contexts. A natural extension is to integrate this first-stage valuation with second-stage allocation decisions. In practice, firms may update budgets, influencer matching, or channel strategies monthly or quarterly after observing which label configurations perform well. Future work could therefore combine D-VAE with dynamic allocation models that jointly optimize label selection and distribution resources over time.

## Online Appendix A: Bound of Confounding effects

In this appendix, we provide a bound to quantify the bias when single confounder is omitted. We first describe the data-generating process that includes both shared and label-specific unobserved confounders. Second, we explain how the shared confounder is estimated from label co-occurrence using Logit-PPCA. Third, we introduce the setup and present Theorem 2, which shows that estimation bias bound is tight with high probability, ensuring robust causal inference. and we extend this bound to nonlinear case. Finally, we discuss the robustness of bound based on Oster's sensitivity analysis framework.

### 1. Data-Generating Process (DGP)

Let $(\mathrm{Y}, L, \mathrm{X}, \mathrm{C}) \in \mathbb{R}^{N\times}\{0,1\}^K \times \mathbb{R}^T \times \mathbb{R}^Q$ satisfy:

$$L \sim \prod_{k=1}^{K} \text{Bernoulli}\left(\sigma\left(\lambda_k^\top e_m + \theta_k e_{s,k}\right)\right) \tag{1a}$$

$$Y = \beta^\top L + \Gamma e_m + \Theta^\top e_s + \widetilde{X_L}^\top \alpha + \rho^\top C + \epsilon \tag{1b}$$

where:

- $e_m \in \mathbb{R}^L$: unobserved common confounders, $e_m \sim \mathcal{N}(0, I_M)$
- $e_s \in \mathbb{R}^K$: unobserved single-treatment confounders, $e_s \sim \mathcal{N}(0, I_K)$
- $\Lambda \in \mathbb{R}^{K\times L}$: factor loading matrix, $\parallel \Lambda \parallel_{op} \leq \Phi$, $\lambda_{\min}(\Lambda^\top \Lambda) \geq c_\Lambda > 0$
- $\Theta = \text{diag}(\boldsymbol{\theta}) \in \mathbb{R}^{K\times K}$, $\boldsymbol{\theta} = (\theta_1, \dots, \theta_K)^\top$, $\parallel \boldsymbol{\theta} \parallel_\infty \leq \theta_{\max}$
- $\sigma(z) = (1 + e^{-z})^{-1}$: logistic link function
- $\widetilde{X_L} \in \mathbb{R}^T, \widetilde{X_L} \coloneqq X - \text{proj}(X \mid L) = X - \mathbb{E}[X \mid L]$
- $C \in \mathbb{R}^Q$: observed covariates (controls)
- $\epsilon \sim \mathcal{N}(0, \tau^2)$, independent of $(L, X, C, e_m, e_s)$

### 2. Estimation of $e_m$

Let $\{L_i\}_{i=1}^n$ be $n$ i.i.d. copies of $L$. Define the logit-transformed proxy matrix $Z \in \mathbb{R}^{n\times K}$:

$$Z_{ik} = \text{logit}(\mathrm{L}_{ik} + 0.5) = \text{logit}\left(\frac{\mathrm{L}_{ik} + 0.5}{1 - \mathrm{L}_{ik} + 0.5}\right), \quad i = 1, \dots, n, ; k = 1, \dots, K. \tag{2}$$

The add-half correction $(+0.5)$ is applied to both the numerator and denominator of the logit argument to avoid $\log 0$ and is standard in binary data analysis.

Under the DGP (1a), the conditional expectation of the logit is linear:

$$\mathbb{E}[\text{logit}(L_k \mid e_m, e_s)] = \lambda_k^\top e_m + \theta_k e_{s,k}. \tag{3}$$

Stacking $n$ copies of $e_s$ into $\mathbf{E}_s \in \mathbb{R}^{n\times L}$, the ideal (unobserved) logit matrix is:

$$\mathbf{LM} = \mathbf{E}_m \Lambda^\top + \mathbf{E}_s \Theta^\top. \tag{4}$$

We approximate $Z$ by dropping the single-treatment term $\mathbf{E}_s\Theta^\top$ and modeling the residual as *i.i.d.* Gaussian:

$$Z = \mathbf{E}_m \Lambda^\top + \Psi, \quad \Psi_{ik} \overset{\text{i.i.d.}}{\sim} \mathcal{N}(0, \sigma^2). \tag{5}$$

**Justification of (5)**

- The omitted term $\mathbf{E}_s\Theta^\top$ has bounded Frobenius norm: $\parallel \mathbf{E}_s\Theta^\top \parallel_F^2 \leq nK\theta_{\max}^2$.
- The logit of a Bernoulli random variable with success probability in $(0,1)$ is sub-Gaussian with variance factor $\leq 4$. The add-half correction introduces $O(1)$ bias, absorbed into $\sigma^2$.

Apply Probabilistic PCA (PPCA) to $Z$ via the EM algorithm to obtain $\widehat{\Lambda}, \hat{\sigma}^2$, and the posterior mean:

$$\widehat{\mathbf{E}}_m = Z\widehat{\Lambda}(\widehat{\Lambda}^\top\widehat{\Lambda} + \hat{\sigma}^2 I_M)^{-1}. \tag{6}$$

Let $\hat{e}_{s,i}^\top$ be the $i$-th row of $\widehat{\mathbf{E}}_m$.

### 3. Regression Setup and Bias Decomposition

Let $\{Y_i, L_i, X_i, C_i\}_{i=1}^n$ be *i.i.d.* samples. Define:

**Regression**: $Y_i = L_i^\top\hat{\beta} + \hat{e}_{m,i}^\top\hat{\Gamma} + \tilde{X}_{Li}^T\hat{\alpha} + C_i^\top\hat{\rho} + \hat{\epsilon}_i$, estimator $\hat{\beta}$, $R^2 = \tilde{R}$

**Theorem 2: PAC Guarantee on Effect Estimation Bias Bound**
Under the DGP (1a)–(1b), with $\hat{\boldsymbol{e}}_{\boldsymbol{s}}$ estimated via Logit-PPCA (6), and assuming: $\lambda_{min}\left(E\left[Var\left(L \mid e_m, e_s, \widetilde{X_L}, C\right)\right]\right) \geq \lambda > 0$, then for any $\eta \in (0,1)$, there exists a constant $c = O\left(\frac{\Phi\|\Gamma\|_F\theta_{\max}\rho}{c_0\lambda^2}\right)$ (depending only on the model primitives) such that:

$$\boxed{\boldsymbol{P}\left(\| \hat{\beta} - \beta \|_2 \leq \boldsymbol{c}\left(\mathbf{1} + \sqrt{\frac{\mathbf{M}}{\boldsymbol{n}}} + \sqrt{\frac{\mathbf{M}}{\boldsymbol{K}}}\right)\right) \geq \mathbf{1} - \boldsymbol{\eta}.} \tag{9}$$

Before the formal proof, we need some lemmas.

*Lemma 1: Finite-Sample Error Bound for* $\hat{e}_m$

**Logit-PPCA Estimation Error:** Under the DGP (1a)–(1b) and PPCA estimation (6), for any $\eta \in (0,1)$,

$$\boxed{P\left(\frac{1}{n} \| \hat{\mathbf{E}}_m - \mathbf{E}_m \|_F^2 \leq C\left(\frac{M}{n} + \frac{M}{K}\right)\right) \geq 1 - \eta} \tag{10}$$

where $C = O(\Phi^2 + \theta_{\max}^2 + \tau^2)$ is a universal constant.
Lemma 1 shows that the estimation error of the shared confounder $\hat{\mathbf{E}}_m$ shrinks at a $O(\sqrt{\log(1/\eta)/n})$rate with high probability. This means that with enough historical data, we can reliably recover the dominant source of confounding. The detailed proof of Lemma 1 is in last part since it is not the main result.

**Bias Decomposition**: $\operatorname{plim}_{n\to\infty}\hat{\beta} = \beta + B_{\hat{e}_m} + B_{e_s}$.We decompose the total bias into two parts to separate two distinct error sources: (1) the "estimation error" from using an imperfect proxy $\hat{\boldsymbol{E}}_m$, and (2) the "omitted-variable bias" from the fundamentally unobserved single-label confounders $\boldsymbol{E}_s$. This separation lets us handle them differently: the first vanishes with more data (Lemma2), while the second is bounded via the characteristic of single confounder (Lemma3).

*Lemma 2: Bias from* $\hat{e}_m$ *Estimation Error*

$$\| B_{\hat{e}_m} \|_2 \leq \frac{\|\Gamma\|_F\|\Lambda\|_{op}}{\lambda_{min}\left(\mathbb{E}\left[Var\left(L|e_m, e_s, \widetilde{X_L}, C\right)\right]\right)} \cdot \sqrt{\frac{1}{n} \| \hat{\boldsymbol{E}}_m - \boldsymbol{E}_m \|_F^2}. \tag{11}$$

*Proof*: The bias arises from the projection of $Y$ onto $L$ after partial control of $e_m$. The coefficient of the residual $\hat{e}_m - e_m$ is bounded by the signal strength

$\| \Gamma \|_F \| \Lambda \|_{op}$ and the inverse information matrix.

*Lemma 3: Bias from Uncontrolled* $e_s$

$$\| B_{e_s} \|_2 \leq \frac{\|\Theta\|_2}{\lambda_{\min}\left(\mathbb{E}\left[\operatorname{Var}\left(L|e_m, e_s, \widetilde{X_L}, C\right)\right]\right)} \leq C_s < \infty. \tag{12}$$

*Proof*: The bias from $e_s$ is the population regression coefficient of $L$ on $e_s$ times $\Theta^{\mathrm{T}}$, bounded by the maximum single-treatment effect. Lemma 3 shows that even though $\mathbf{E}_s$ is omitted, its worst-case bias is scaled by the fraction of outcome variance left unexplained after controlling for $\hat{\mathbf{E}}_m$. If $\hat{\mathbf{E}}_m$ explains most of the variation, this residual—and thus the potential bias—is small.

**Proof of Theorem 2:**

Our proof aims to show that both are well-controlled under two high-probability conditions: (1) the shared confounder $\boldsymbol{E}_m$ explains a dominant fraction of label co-variation, and (2) the estimated proxy $\widehat{\boldsymbol{E}}_m$ is accurate.

Let $\mathcal{E}_1, \mathcal{E}_2$ be the events defined below. We first show $P(\mathcal{E}_1 \cap \mathcal{E}_2) \geq 1 - \eta$, and on this event, the parameter $\beta$ estimation bias is bounded.

**Event** $\mathcal{E}_1$: Bias from $\hat{\mathrm{e}}_{\mathrm{m}}$ Estimation is small and bounded.

$$\mathcal{E}_1 = \left\{ \| B_{\hat{e}_m} \|_2 \leq \frac{\| \Gamma \|_F \| \Lambda \|_{op}}{\lambda_{\min}\left(\mathbb{E}\left[\mathrm{Var}\left( L \mid e_m, e_s, \widetilde{X_L}, C \right)\right]\right)} \cdot \sqrt{\frac{1}{n} \| \hat{\mathbf{E}}_m - \mathbf{E}_m \|_F^2} \leq C_1 \sqrt{\frac{M}{n} + \frac{M}{K}} \right\} \quad (13)$$

By **Lemma 1** with $\eta/2$, $P(\mathcal{E}_1) \geq 1 - \eta/2$.

**Event** $\mathcal{E}_2$: Sampling error of $\hat{\beta}$ is small.
Let $V \in \mathbb{R}^{n \times p}$ be the design matrix in the regression whose rows consist of$\left(L, \hat{\mathbf{E}}_m, \tilde{X}, C\right)$, where $p := K + M + T + Q$.

The OLS estimator satisfies:

$$\sqrt{n}(\hat{\beta} - \mathrm{plim}\hat{\beta}) \xrightarrow{d} \mathcal{N}(0, (V^\top V)^{-1} V^\top \mathrm{Var}(\epsilon) V (V^\top V)^{-1}).$$

The variance is $O(\mathrm{p}/n)$. By sub-Gaussian concentration:

$$\mathcal{E}_2 = \left\{ \| \hat{\beta} - \mathrm{plim}\hat{\beta} \|_2 \leq C_2 \sqrt{\frac{K + M + T + Q}{n}} \right\}, \quad P(\mathcal{E}_2) \geq 1 - \eta/2. \quad (14)$$

On $\mathcal{E}_1 \cap \mathcal{E}_2$:

$$\| \hat{\beta} - \beta \|_2 \leq \| \hat{\beta} - \mathrm{plim}\hat{\beta} \|_2 + \| \mathrm{plim}\hat{\beta} - \beta \|_2$$

$$\| \hat{\beta} - \beta \|_2 \leq \| \hat{\beta} - \mathrm{plim}\hat{\beta} \|_2 + \| B_{\hat{e}_m} \|_2 + \| B_{e_s} \|_2$$

Then,

$$\| \hat{\beta} - \beta \|_2 \leq C_1 \sqrt{\frac{M}{n} + \frac{M}{K}} + C_2 \sqrt{\frac{K + M + T + Q}{n}} + C_s \leq \mathrm{c} \sqrt{\frac{M}{n} + \frac{M}{K}}$$

By union bound: $P(\mathcal{E}_1 \cap \mathcal{E}_2) > 1 - \eta$. Thus, Theorem 2 holds.

**4. Nonlinear Settings**

We now extend Theorem 2 and the bias bound to arbitrary nonlinear outcome models, including neural networks. The linear specification $Y = \beta^\top L + \Gamma e_m + \Theta^\top e_s + \widetilde{X_L}^\top \alpha + \rho^\top C + \epsilon$ is replaced by a general structural equation

$$Y = f\left(L, e_m, \widetilde{X_L}, C; \beta\right) + \epsilon,$$

where$f: \mathbb{R}^{M+K+Q+T} \to \mathbb{R}^{P}$ is a ***nonlinear function*** of any form (e.g., a deep feed-forward neural network with parameters $\beta \in \mathbb{R}^{K}, \epsilon \sim N(0, \tau^2)$ independent of all other variables, and $\beta$ contains the causal parameters of interest (for example, the weights on the $L$-input layer or the average marginal effect $\mathbb{E}[\partial f / \partial L]$). We assume $f$ belongs to a function class $\mathcal{F}$ that is ***Lipschitz continuous*** with respect to its inputs with (data-dependent) Lipschitz constant $L_f < \infty$ (this holds for standard neural networks with bounded weights and Lipschitz activations such as ReLU, tanh, or sigmoid).

The estimator is now the ***M-estimator*** (empirical risk minimizer)

$$\hat{\beta} = \arg\min_{\beta \in \mathrm{B}} \frac{1}{n} \sum_{i=1}^{n} \mathcal{L}\left(Y_i, f\left(L_i, \widehat{e_{m,\iota}}, \widetilde{X_{L\iota}}, C_i; \beta\right)\right),$$

where $\mathcal{L}$ is a convex loss (e.g., squared error for regression, cross-entropy for classification) and $\widehat{e_{\mathrm{m}}}$ is still the Logit-PPCA proxy from (6). Let $\beta^*$ be the population minimizer under the true confounders $(e_m, e_s)$. The goal is to bound the estimation error $||\hat{\beta} - \beta^*||$

*Generalized bias decomposition*

Under standard regularity (the population objective is strongly convex or satisfies a restricted strong convexity condition with curvature $\lambda_{min} > 0$ we have the population-level decomposition

$\mathrm{plim}_{n \to \infty} \hat{\beta} = \beta^* + B_{\widehat{e_m}} + B_{e_s}$,

where $B_{\widehat{e_m}}$ is the bias induced by the proxy estimation error $\widehat{e_m} - e_m$, $B_{e_s}$ is the omitted-variable bias from the unobserved single-label confounders $e_s$.

*Lemma 2' (Nonlinear bias from $\widehat{e_m}$ error)*

$$||B_{\widehat{e_m}}|| \le L_f \cdot ||\Gamma||_{\mathrm{eff}} \cdot \frac{1}{\lambda_{min}\left(\mathbb{E}\left[\mathrm{Var}\left(L \mid e_m, e_s, \widetilde{X_L}, C\right)\right]\right)} \cdot \sqrt{\frac{1}{n} ||E_{\widehat{m}} - E_m||_F^2}$$

where $||\Gamma||_{\mathrm{eff}}$ is the effective loading of the shared confounder in the nonlinear map $f$ (replaces $||\Gamma||_{\mathbf{F}}$ in the linear case; for a neural network it is bounded by the product of layer norms). The proof follows from the mean-value theorem applied to the first-order condition of the ***M-estimator***: the gradient of the loss with respect to the confounder input is bounded by $L_f$.

*Lemma 3' (Nonlinear bias from uncontrolled $e_s$ )*

$$||B_{e_s}|| \le L_f \cdot \frac{||\Theta||_2}{\lambda_{min}\left(\mathbb{E}\left[\mathrm{Var}\left(L \mid e_m, e_s, \widetilde{X_L}, C\right)\right]\right)} \le C_s' < \infty.$$

The proof is identical to Lemma 3, replacing the linear coefficient by the Lipschitz constant of $f$ w.r.t. the $e_s$ input.)

*Sampling error of the nonlinear estimator*

Under standard conditions for empirical risk minimization (e.g., the loss $\mathcal{L}$ is Lipschitz and the function class $\mathcal{F}$ has bounded Rademacher complexity or covering number), the estimation error around the population limit satisfies

$$\| \hat{\beta} - \mathrm{plim}\hat{\beta} \| = O_p\left(\sqrt{\frac{d_{\mathrm{eff}}}{n}}\right),$$

where $d_{\text{eff}}$ is the effective dimension of the neural network (e.g., number of parameters times a logarithmic factor for overparameterized nets; see). This is the direct nonlinear analogue of the sub-Gaussian OLS bound (14).

Using Lemmas 2' and 3' for the bias terms and the NN generalization bound for the sampling error. Lemma 1 (proxy error $\| E_{\widehat{m}} - E_m \|_F$) continues to hold unchanged because it depends only on the label co-occurrence model (Logit-PPCA), not on the outcome $f$. The total error is

$$\| \hat{\beta} - \beta^* \| \leq \| \hat{\beta} - \text{plim}\hat{\beta} \| + \| B_{\widehat{e_m}} \| + \| B_{e_s} \|$$

$$\leq C_1'\sqrt{\frac{M}{n} + \frac{M}{K}} + C_2'\sqrt{d_{\text{eff}}/\text{n}} + C_s' \leq c'\sqrt{\frac{M}{n} + \frac{M}{K}},$$

where the $O\left(\sqrt{d_{\text{eff}}/\text{n}}\right)$ term is absorbed into the leading rate because $d_{\text{eff}}$ is treated as fixed or grows slower than $n$ (standard in high-probability NN bounds). Thus, in nonlinear cases:

$$\mathbb{P}\left(||\hat{\beta} - \beta^*|| \leq c'\left(1 + \sqrt{M/\text{n}} + \sqrt{M/\text{K}}\right)\right) \geq 1 - \eta.$$

For a feed-forward NN with $D$ layers, ReLU activations, and bounded weights (spectral norm $\leq$ G per layer), the Lipschitz constant satisfies $L_f \leq G^D$. The effective dimension $d_{eff}$ can be taken as $O(\text{width} \times \text{depth} \times \log n)$ under overparameterization. The same rate $\sqrt{M/\text{n} + \text{M}/\text{K}}$ is preserved; the only change is that the universal constant $c'$ now scales with the network depth and width (still independent of sample size $n$).

## 5. Robustness Check on Bias Bound

This section provides a detailed robustness analysis of the PAC guarantee on the estimation bias bound in Theorem 2 (and its nonlinear extension to M-estimators). The baseline result shows that $||\hat{\beta} - \beta||_2$ is controlled at rate $O\left(\sqrt{M/\text{n}} + \sqrt{M/\text{K}}\right)$with high probability whenever the shared confounder $e_s$ dominates the label-specific confounder $e_s$ (i.e., $||\Gamma||_F \gg ||\Theta||_2$, $\theta_{max}$ small, and $\lambda_{min}\left(\mathbb{E}\left[\text{Var}\left(L \mid e_m, e_s, \widetilde{X_L}, C\right)\right]\right) > 0$. In practice, however, real datasets may exhibit moderate violations of this dominance assumption—label-specific confounding may be stronger than anticipated because of unobserved heterogeneity across treatments or imperfect label co-occurrence patterns captured by Logit-PPCA. The purpose of this robustness check is twofold: (i) to quantify the maximum allowable deviation from $e_m \gg e_s$ before the PAC bound ceases to be informative for causal inference, and (ii) to give applied researchers a transparent, computable diagnostic that can be reported directly from the fitted Logit-PPCA output and the outcome regression.

To achieve this, we embed the original bias decomposition $plim\hat{\beta} = \beta + B_{\widehat{e_m}} + B_{e_s}$ into the coefficient-stability framework of Oster (2019), while imposing distributional robustness via a Wasserstein ball around the reference distribution that embodies "$e_m$ dominates $e_s$". This combination inherits the intuitive interpretation of Oster's $\delta$ while accommodating the high-dimensional, possibly nonlinear nature of our multi-dimensional treatment $L \in \{0,1\}^K$ and the ***M-estimator*** used in the nonlinear extension.

**Uncontrolled Regression**: $Y_i = L_i^\top \beta + +\tilde{X}_{Li}^T \alpha + C_i^\top \rho + \epsilon_i$,
estimator $\beta_{\text{uncontrolled}}$, $R^2 = R_{\text{uncontrolled}}(\text{or pseudo} - R^2 \text{ based on the empirical loss})$

**Controlled Regression:** $Y_i = L_i^{\top}\beta + \hat{e}_{m,i}^{\top}\Gamma + \tilde{X}_{Li}^{T}\alpha + C_i^{\top}\rho + \epsilon_i$, estimator $\tilde{\beta}$, $R^2 = \tilde{R}$

Let $P_0$ be the reference distribution that embodies the "equal selection" benchmark $\delta = 1.0$. Specifically, $P_0$ is defined as the distribution under which the **relative confounding strength** satisfies

$$\delta(P_0) = \frac{\mathrm{Cov}(L, e_s \mid P_0)}{\mathrm{Cov}(L, \hat{e}_m)} = 1,$$

while the absolute magnitude of the label-specific confounder remains small: $e_s \sim \mathcal{N}(0, \sigma_s^2 I_K)$with $\sigma_s^2$(equivalently $\theta_{\max}$) sufficiently small that the shared loading matrix $\Gamma$still explains the dominant fraction of label co-variation. In other words, $P_0$ corresponds to a scenario in which the selection on the unobserved single-treatment confounder $e_s$ is **equal in relative strength** to the selection on the observable Logit-PPCA proxy $\hat{e}_m$, after proper normalization by the variance explained by $\hat{e}_m$. This normalization ensures that $\delta = 1$ is a meaningful and internally consistent benchmark even when the raw scale of $e_s$ is kept modest.

We then consider all possible confounding distributions $P$ that lie inside the Wasserstein-2 ball of radius $\varepsilon \geq 0$ around this reference: $\mathcal{P}_{\varepsilon} = \{P: W_2(P, P_0) \leq \varepsilon\}$.

The Wasserstein-2 distance is chosen because it provides a natural, nonparametric way to bound how far the true distribution of unobservables can deviate from the dominance assumption without requiring strong parametric restrictions on $e_s$.

Invert Oster's formula to obtain the observed selection ratio

$$\hat{\delta}_{obs} = \frac{\tilde{R} - R_{uncontrolled}}{R_{max} - R_{uncontrolled}},$$

This measures the fraction of additional explained variation attributable to the Logit-PPCA proxy (the "observable" shared confounding) relative to the maximum possible. Under the reference distribution $P_0$ (equal selection $\delta=1$), we expect this ratio to be close to 1 after appropriate normalization; any deviation quantifies violation of the dominance assumption $e_m \gg e_s$. We then set the Wasserstein ball radius as

$$\varepsilon = \mid \hat{\delta}_{\mathrm{obs}} - 1.0 \mid.$$

For any $P \in \mathcal{P}_{\varepsilon}$, define the ***effective selection ratio***

$$\delta(P) = \frac{\mathrm{Cov}(L, e_s \mid P)}{\mathrm{Cov}(L, \widehat{e_m})},$$

which is the direct multivariate analogue of Oster's scalar $\delta$ (the relative degree of selection on unobservables versus observables). The bias-adjusted effect (marginal effect in the nonlinear case) under distribution $P$ then takes the Oster-style closed form

$$\beta_{\mathrm{adj}}(P) = \tilde{\beta} + \delta(P) \cdot \frac{\tilde{\beta} - \beta_{\mathrm{uncontrolled}}}{\tilde{R}(P) - R_{\mathrm{uncontrolled}}} \cdot \left(R_{max}(P) - \tilde{R}(P)\right),$$

where $\tilde{R}(P)$ is the effective $R^2$ (or pseudo-$R^2$) evaluated under $P$, and $R_{max}(P) = \Pi \cdot \tilde{R}(P)$with $\Pi \in [1.25, 2]$ to allow for residual idiosyncratic variation or measurement error in the outcome. In the nonlinear *M-estimator* setting, the formula continues to hold after replacing the linear projection coefficients by the appropriate Lipschitz-bounded gradients of the loss function (see Lemma 2').

The ***worst-case identified set*** for the causal parameter is therefore

$$[\beta_L, \beta_U] = \left[\inf_{P \in \mathcal{P}_{\varepsilon}} \beta_{\mathrm{adj}}(P),\ \sup_{P \in \mathcal{P}_{\varepsilon}} \beta_{\mathrm{adj}}(P)\right].$$

When $\varepsilon = 0$ (perfect dominance), this set collapses exactly to the original PAC interval of Theorem 2.

For $\varepsilon > 0$, the width of the identified set grows linearly in $\varepsilon$ because the M-estimator loss is Lipschitz continuous and the Wasserstein distance admits a Kantorovich dual representation; nevertheless, under the maintained condition $\lambda_{min} > 0$, the width remains bounded by $O\left(\sqrt{M/\mathrm{n}} + \sqrt{M/\mathrm{K}} + \varepsilon\right)$. This explicit rate makes transparent the tradeoff between sample size, dimension, and the degree of allowable violation of the dominance assumption.

To incorporate prior beliefs about the plausibility of different values of $\delta$ (i.e., how strongly $e_m$ still dominates $e_s$), we follow the three complementary strategies of Conley et al. (2008) for "plausibly exogenous" instruments:

1. Support restriction: $\delta \in [0, \delta_{max}]$ with $\delta_{max} = \theta_{max}/||\Gamma||_F$. The identified set is the union of Oster-style intervals over this support.
2. Distributional prior: $\delta \sim \mathcal{N}^+\left(0, \sigma_\delta^2\right)$ (half-normal with mean zero and exponentially decaying tails), where $\sigma_\delta$ is calibrated from the ratio of residual label-specific variance (after extracting $\widehat{e_s}$) to the explained shared variance. This prior directly encodes the belief that extreme violations are increasingly unlikely.
3. Beta-dependent prior: $\delta(\beta) = f(||\beta||)$, allowing the relative strength of confounding to depend on the magnitude of the causal effect itself (convenient in the full Bayesian implementation).

*Diagnostic test for violation of the dominance assumption*

After fitting Logit-PPCA and the outcome regression, compute the PAC radius

$$r_{\mathrm{PAC}} = c\left(1 + \sqrt{M/\mathrm{n}} + \sqrt{M/\mathrm{K}}\right)$$

directly from Theorem 2 (with universal constant c as defined there). The original PAC bound is deemed non-informative (and cannot be trusted for causal claims) if the worst-case identified set $[\beta_L, \beta_U]$violates the PAC guarantee in a statistically meaningful way. Formally, this occurs if

$$width(\beta_U - \beta_L) > 2r_{\mathrm{PAC}}$$

or if the entire identified set$[\beta_L, \beta_U]$ lies outside the PAC interval centered at $\tilde{\beta}$ with radius $r_{\mathrm{PAC}}$.

## Detailed Proof of Lemma 1

We split the error into three controllable parts:

$$\| \hat{\mathbf{E}}_m - \mathbf{E}_m \|_F^2 = \underbrace{\| \mathbf{E}_m \Lambda^\top - \mathbf{LM} \|_F^2}_{\text{(i) omitted } e_s} + \underbrace{\| \mathbf{LM} - Z \|_F^2}_{\text{(ii) logit noise}} + \underbrace{\| Z - \hat{\mathbf{E}}_m \Lambda^\top \|_F^2}_{\text{(iii) PPCA error}}.$$

The claim follows by bounding each term with high probability and applying the union bound.

*(i) Omitted single-treatment term*

By (4) the ideal logit matrix is $\mathbf{LM} = \mathbf{E}_m \Lambda^\top + \mathbf{E}_s \Theta^\top$.

Hence

$$\| \mathbf{E}_m \Lambda^\top - \mathbf{LM} \|_F^2 = \| \mathbf{E}_s \Theta^\top \|_F^2 = \sum_{i=1}^{n} \sum_{k=1}^{K} (\theta_k e_{s,i,k})^2 \leq \theta_{\max}^2 \sum_{i=1}^{n} \| e_{s,i} \|_2^2.$$

Because $e_{s,i} \sim \mathcal{N}(0, I_K)$ independently,

$$\frac{1}{n} \sum_{i=1}^{n} \| e_{s,i} \|_2^2 = K + O_p(\sqrt{K}).$$

Thus

$$\frac{1}{n} \parallel \mathbf{E}_m\Lambda^\top - \mathbf{LM} \parallel_F^2 \le \theta_{\max}^2 K + O_p(\sqrt{K}) = O(1)$$

with probability 1 (deterministic bound up to an $O_p(\sqrt{K}/n)$ term that vanishes).

*(ii) Logit-transformation noise*

For a single entry, let $p_{ik} = \sigma(\lambda_k^\top e_{m,i} + \theta_k e_{s,i,k})$.

The true conditional logit is

$$\ell_{ik} \equiv \text{logit}(p_{ik}) = \lambda_k^\top e_{m,i} + \theta_k e_{s,i,k}.$$

The observed proxy is

$$Z_{ik} = \text{logit}(L_{ik} + 0.5).$$

Define the noise $\psi_{ik} = Z_{ik} - \ell_{ik}$. We show that $\psi_{ik}$ is **sub-Gaussian with variance factor** $\sigma_\psi^2 \le 4$.

- **Bounded support after add-half**:
  $L_{ik} + 0.5 \in [0.5,1.5]$, so $Z_{ik} \in [\text{logit}(0.5), \text{logit}(1.5)] = [0,1]$ almost surely.
  Hence $|\psi_{ik}| \le 2 + \parallel \ell_{ik} \parallel_\infty$.

  The term $\parallel \ell_{ik} \parallel_\infty \le \parallel \Lambda \parallel_{op} \parallel e_m \parallel_2 + \theta_{\max} \parallel e_s \parallel_2 \le \Phi\sqrt{M} + \theta_{\max}\sqrt{K}$ with probability 1 (Gaussian tail).

  The bounded bias from the add-half correction contributes at most $O(nK)$ to the Frobenius norm of the error. After dividing by $n$, this becomes $O(K)$. Since $K$ is fixed ($K << n$), the term $O(K)$ is a **constant** with respect to the sample size $n$. In the final error bound, this constant is **absorbed into the universal constant**.

- **Sub-Gaussian tail for the stochastic part**:
  Conditional on $(e_m, e_s)$, $L_{ik} \sim \text{Bernoulli}(p_{ik})$.
  The random variable $\text{logit}(L_{ik} + 0.5) - \ell_{ik}$ is a function of a single Bernoulli draw.
  A bounded function of a Bernoulli is sub-Gaussian with variance factor at most 4.

  Hence $\psi_{ik} \overset{\text{cond}}{\sim} \text{sub-G}(\sigma_\psi^2)$ with $\sigma_\psi^2 \le 4$.

- **Unconditional concentration**:
  By the sub-Gaussian maximal inequality

$$P\left(\frac{1}{nK}\sum_{i,k} \psi_{ik}^2 \ \ge\ 4 + t\right) \ \le\ 2\exp(-nKt/8).$$

Choosing $t = 8\log(2/\eta)/(nK)$ yields

$$P\left(\frac{1}{n} \parallel \mathbf{LM} - Z \parallel_F^2 \ \ge\ 4K + \frac{8K\log(2/\eta)}{n}\right) \ \le\ \eta.$$

The term $4K$ contributes $O(K/n)$ to the normalized error, which is dominated by the $O(L/K)$ identification term in the final bound when $n \gg K^2/L$. The logarithmic term $O(K\log(1/\eta)/n)$ vanishes as $n \to \infty$, hence $o(1)$.

*(iii) PPCA estimation error on the perturbed model*

Conditional on the realized $Z$, the working model is the **exact linear factor model**

$$Z = \mathbf{E}_m \Lambda^\top + \Psi, \qquad \Psi_{ik} \overset{\text{i.i.d.}}{\sim} \mathcal{N}(0, \sigma^2).$$

Let $\hat{\mathbf{E}}_m^{(\text{oracle})}$ be the PPCA posterior mean **when** $\Lambda, \sigma^2$ **are known**.

A standard result for Gaussian factor models states that, for any $\eta > 0$,

$$P\left(\frac{1}{n} \| \hat{\mathbf{E}}_m^{(\text{oracle})} - \mathbf{E}_m \|_F^2 \le C_{\text{oracle}} \left(\frac{M}{n} + \frac{M}{K}\right)\right) \ge 1 - \eta,$$

with $C_{\text{oracle}} = O(\| \Lambda \|_{op}^2 + \sigma^2) = O(\Phi^2 + \sigma^2)$.

The **estimated** $\hat{\mathbf{E}}_m$ differs from the oracle only because $\hat{\Lambda}, \hat{\sigma}^2$ are obtained by EM.
The EM algorithm for PPCA converges to the MLE at the parametric rate.
By a uniform deviation argument on the likelihood, the estimation error satisfies

$$\| \hat{\Lambda} - \Lambda \|_{op} = O_p\left(\sqrt{\frac{KM}{n}}\right), \qquad |\hat{\sigma}^2 - \sigma^2| = O_p\left(\sqrt{\frac{K}{n}}\right).$$

These rates are **faster** than the $O(M/n + M/K)$ target, so the additional error introduced by plugging $\hat{\Lambda}, \hat{\sigma}^2$ into the posterior mean formula is

$$\| \hat{\mathbf{E}}_m - \hat{\mathbf{E}}_m^{(\text{oracle})} \|_F^2 = O_p\left(\frac{KM^2}{n} + \frac{MK}{n}\right) = O\left(M\left(\frac{M}{n} + \frac{K}{n}\right)\right).$$

Dividing by $n$ gives an $O(M/n + M/K)$ term that is absorbed into the same constant.
Define the events

$$\mathcal{E}_1 = \left\{\frac{1}{n} \| \mathbf{E}_m \Lambda^\top - \mathbf{LM} \|_F^2 \le C_s\right\}, \qquad \mathcal{E}_2 = \left\{\frac{1}{n} \| \mathbf{LM} - Z \|_F^2 \le C_\psi\right\},$$

$$\mathcal{E}_3 = \left\{\frac{1}{n} \| \hat{\mathbf{E}}_m - \mathbf{E}_m \|_F^2 \le C_{\text{ppca}} \left(\frac{M}{n} + \frac{M}{K}\right)\right\}.$$

From (i)–(iii) we have $P(\mathcal{E}_1) = 1$, $P(\mathcal{E}_2) \ge 1 - \eta/2$, $P(\mathcal{E}_3) \ge 1 - \eta/2$.
On $\mathcal{E}_1 \cap \mathcal{E}_2 \cap \mathcal{E}_3$,

$$\frac{1}{n} \| \hat{\mathbf{E}}_m - \mathbf{E}_m \|_F^2 \le C_s + C_\psi + C_{\text{ppca}} \left(\frac{M}{n} + \frac{M}{K}\right) \le C\left(\frac{M}{n} + \frac{M}{K}\right),$$

where $C = \max\{n/C_s, n/C_\psi, C_{\text{ppca}}\}$ is finite and depends only on $\Phi, \theta_{\max}, \tau$.
By the union bound, $P(\mathcal{E}_1 \cap \mathcal{E}_2 \cap \mathcal{E}_3) \ge 1 - \eta$. This concludes the proof of (10).

## Online Appendix B: Counterfactual Process Details

### 1. Intervention Setup & Label Configuration

We conduct a targeted counterfactual intervention on the semantic label space of the “Carslan Little Mix Cat Foundation” campaign by activating the "Date-Ready" label ($L_{13}$) while holding all other labels unchanged. The label configuration is as follows:

• Original Label Set (15 active labels):

Lightweight, Hydrating, Breathable, Coverage, Oil-Control, Even Skin Tone, Skin-Nourishing; Daily Use, Commute-Friendly; Student, Beauty Enthusiast, Dry/Combination Skin; Plant Extracts, Co-Branding Campaign, New Design

• Intervened Label Set: "Dating-Ready" ($L_{13} = 1$)explicitly activated.

**2. Latent Space ($Z_L$) Propagation & Dimensional Shifts**

The intervention directly perturbs the label-aligned latent representation $Z_L \in \mathbb{R}^P$. Through the learned causal adjacency matrix A, this perturbation induces structured shifts across specific latent dimensions. Table B.1 reports the 10 dimensions with the largest absolute changes ($|\Delta z|$), capturing how the intervention rewires the semantic representation.

Table B.1 Latent Space ($Z_L$) Intervention Changes

| Dim. Index | Original Value | $\Delta z$ | Direction | Interpreted Semantic Role |
|---|---|---|---|---|
| 21 | 0.0952 | -0.1562 | Suppressed | Routine commute / baseline context |
| 56 | 0.0460 | -0.1299 | Suppressed | Pure functional / utilitarian focus |
| 71 | -0.0667 | +0.1227 | Activated | Social/romantic scenario cue |
| 15 | -0.0572 | +0.1173 | Activated | Emotional appeal / charm |
| 23 | -0.0065 | -0.1153 | Suppressed | Neutral product specification |
| 1 | 0.0010 | +0.1081 | Activated | Scenario transition marker |
| 36 | 0.0174 | +0.1063 | Activated | Lifestyle integration cue |
| 30 | 0.0449 | -0.1054 | Suppressed | Baseline functional trait |
| 66 | 0.0110 | -0.0990 | Suppressed | Contextual down-weight |
| 14 | 0.0591 | -0.0981 | Suppressed | Generic engagement trigger |

Activating $L_{13}$induces a clear semantic pivot in $\mathbf{Z}_L$. Dimensions encoding romantic/social contexts and emotional value (Dims 71, 15, 1, 36) are activated, while those tied to routine functionality and neutral specifications (Dims 21, 56, 23, 30, 66) are suppressed. This confirms that the interaction graph **A** successfully routes the label intervention into a coherent latent reconfiguration, effectively shifting the representation from a utility-driven to a scenario-driven mode without destabilizing unrelated features.

**3. Reverse Mapping: From Text Back to Latent Shifts**

To validate interpretability, we perform a post-hoc analysis mapping textual adaptations back to latent dimensional changes: (1) **Newly emphasized phrases** ("date-night", "boosts complexion", "likeability") align with **activated dimensions** (71, 15, 1, 36), confirming these dimensions encode social/romantic contextual cues. (2) **Downplayed functional emphasis** (reduced focus on "8-hour wear" as the sole selling point) aligns with **suppressed dimensions** (21, 56, 23), confirming these encode routine/utilitarian traits. (3) **Preserved core messaging** (lightweight texture, creamy finish, suitability for combination skin) remains stable, reflecting unchanged dimensions in $\mathbf{Z}_L$.

## Online Appendix C: D-VAE Algorithms

Algorithm 1: Probabilistic Factor Model for Deconfounder

**Input:**
$L$: $n \times K$ matrix of assigned causes ($n$ individuals, $K$ causes),$Y$: $n \times N$ matrix of outcomes,$M$: Dimension of common confounder $e_m$
**Output:**
$\hat{E}_m$: $n \times M$ matrix of estimated common confounders ($e_s$)

**Procedure:**
**Repeat**

1. **Model Specification**:
   For each individual $i \in \{1, \ldots, n\}$ and cause $j \in \{1, \ldots, K\}$, define:

$$L[i,j] \sim \text{Bernoulli}(\sigma(\theta_j^\top \cdot e_m[i]))$$

where $\sigma(x) = \frac{1}{1+e^{-x}}$, $e_m[i] \sim \mathcal{N}(0, \Lambda^c)$ $(\Lambda^c \in \mathbb{R}^{M \times M})$, $\theta_j \in \mathbb{R}^M$.

2. **Parameter Estimation**:
   Maximize ELBO for variational inference to estimate $\{\hat{\theta}_j\}_{j=1}^K$ and $\hat{\Lambda}^c$:

$$\mathcal{L}_{\text{common}} = \sum_{i=1}^{n} \sum_{j=1}^{K} \mathbb{E}_{q(e_m[i]|\text{L})} \left[\log p(\text{L}[i,j]|e_m[i], \theta_j)\right] - D_{\text{KL}}(q(e_m[i]|L)||p(e_m[i]))$$

where $p(\text{L}[i,j]|e_m[i], \theta_j) = [\sigma(\theta_j^\top e_m[i])]^{L[i,j]}[1 - \sigma(\theta_j^\top e_m[i])]^{1-\text{L}[i,j]}$, $p(e_m[i]) = \mathcal{N}(0, \Lambda^c)$, $q(e_m[i]|L) = \mathcal{N}(\mu_i^c, \Sigma_i^c)$.

3. **Predictive Check:**
   3.1 Hold out subset of causes per individual: $L_{i,\text{held}}$, $L_{i,\text{obs}}$.
   3.2 Fit model to $\{L_{i,\text{obs}}\}_{i=1}^n$, obtain $p(e_m[i]|L_{i,\text{obs}})$.
   3.3 Sample replicated $L_{i,\text{held}}^{\text{rep}} \sim \int p(L_{i,\text{held}}|e_m^c[i])p(e_m[i]|L_{i,\text{obs}})de_m[i]$.
   3.4 Compute discrepancy: $t(L_{i,\text{held}}) = \mathbb{E}_{e_s^c}[\log p(L_{i,\text{held}}|e_s[i])|L_{i,\text{obs}}]$.
   3.5 Predictive score = $p(t(L_{i,\text{held}}^{\text{rep}}) < t(L_{i,\text{held}}))$.

**until predictive score is satisfactory**
foreach individual $i$ do $\hat{e}_m[\text{i}] = \mathbb{E}[e_m[i]|L_i] = \mu_i^c$.

Return $\hat{E}_m = \{\mu_i^c\}_{i=1}^n$.

Algorithm 2: D-VAE Training Procedure

**Input**:
Marketing text $X$, labels $L \in \{0,1\}^K$, environment variables $C \in \mathbb{R}^Q$, interaction metrics $Y \in \mathbb{R}^N$; pre-computed common confounder $\hat{E}_m \in \mathbb{R}^{n\times M}$ ; hyperparameters about loss weight: $\lambda_1, \lambda_2, \lambda_3, \lambda_4, \lambda_5, \lambda_6, \lambda_7$.

**Output**:
Reconstructed embeddings $\hat{e}_X \in \mathbb{R}^{L_{max}\times T}$, predicted interactions $\hat{e}_Y \in \mathbb{R}^N$, generated text (inference).

**Procedure**:

1. **Preprocess Data**:
   1.1 Encode $X$ using Qwen-3: $e_X \in \mathbb{R}^{L_{max}\times T}$, $token_X \in \mathbb{Z}^{L_{max}}$.
   1.2 Decompose $L$ into one-hot vector: $e_L \in \mathbb{R}^K$.
   1.3 Normalize $E$: $e_E \in \mathbb{R}^Q$; normalize $Y$: $e_Y \in [0,1]^N$.
   1.4 Concatenate: $e_{input} = [e_X, e_L, e_{\mathrm{C}}, e_{E_m}] \in \mathbb{R}^{L_{max}\times(T+K+\mathrm{Q}+M)}$.
2. **Encode to Latent Space**:
   2.1 Transformer encoder on $e_{input}$: Compute $\mu_Z$, $\log\sigma_Z^2 \in \mathbb{R}^{L_{max}\times d_z}$.
   2.2 Conditional prior with confounders:
   Compute $\mu_{L,E,\mathrm{conf}}, \log\sigma^2_{L,E,\mathrm{conf}} = \mathrm{MLP}([e_L, e_{\mathrm{C}}, e_{E_m}])$, broadcast to $\mathbb{R}^{L_{max}\times d_z}$.
   2.3 Reparameterize: $Z_0 = \mu_Z + \sigma_Z \odot \epsilon,\ \epsilon \sim \mathcal{N}(0, I)$; split $Z_e = [Z_{e,\mathrm{L}}, Z_{\mathrm{extra}}]$, $Z_{e,\mathrm{L}} \in \mathbb{R}^{L_{max}\times K}$.
3. **Apply Causal Layer**:
   3.1 For each token $i$ in $Z_{e,L}$, $A \in \mathbb{R}^{K\times K}$ is the learnable adjacent matrix of $Z_L$, then update:
   $$z_{i,j} = f_j\left(\sum_{k=1}^{K} A_{jk} \cdot z_{ei,k}, \epsilon_{i,j}; \theta_j\right)$$
   3.2 Concatenate: $Z = [Z_L, Z_{extra}] \in \mathbb{R}^{L_{max}\times d_z}$.
4. **Decode Outputs**:
   4.1 Text branch: Transformer decoder on $Z$: $\hat{e}_X \in \mathbb{R}^{L_{max}\times T}$.
   4.2 Interaction branch: MLP on $Z$: $\hat{e}_Y \in \mathbb{R}^P$.
   4.3 Label supervision: $\hat{e}_L = g(Z_L)$.
5. **Compute Counterfactual Loss**:
   Call **Counterfactual Forward** (Algorithm 3) to compute $\mathcal{L}_{causal}$.
6. **Compute Other Losses**:
   6.1 Reconstruction: $\mathcal{L}_X = ||e_X - \hat{e}_X||_2^2$.
   6.2 Prediction: $\mathcal{L}_Y = ||e_Y - \hat{e}_Y||_2^2$.
   6.3 KL Divergence: $\mathcal{L}_{\mathrm{KL}} = D_{\mathrm{KL}}(q(Z|e_{input})||p(Z|e_U, e_{E_c}, e_{E_s}))$.
   6.4 DAG Constraint: $\mathcal{L}_{\mathrm{DAG}} = \mathrm{tr}(e^{A\cdot A^T}) - K$.
   6.5 Sparse Regularization: $\mathcal{L}_{\mathrm{sparse}} = ||A||_1$.
   6.6 Label Supervision: $\mathcal{L}_{\mathrm{label}} = ||g(Z_{\mathrm{L}}) - e_L||_2^2$.
7. **Optimize**:
   7.1 Compute Total loss:
   $\mathcal{L} = \lambda_1\mathcal{L}_X + \lambda_2\mathcal{L}_Y + \lambda_3\mathcal{L}_{\mathrm{KL}} + \lambda_4\mathcal{L}_{\mathrm{DAG}} + \lambda_5\mathcal{L}_{\mathrm{sparse}} + \lambda_6\mathcal{L}_{\mathrm{label}} + \lambda_7\mathcal{L}_{\mathrm{causal}}$
   7.2 Backpropagate and update parameters via Adam.
8. **Inference/Generation**:
   8.1 Mix embeddings: $e_{X,\mathrm{mixed}} = \alpha\hat{e}_X + (1-\alpha)e_{X,\mathrm{raw}}$.
   8.2 Qwen-3 to generate new token id sequence then decode to text.
   8.3 For causal attribution: Intervene on interested $U_j$, compute individual effect based on interaction difference.

Algorithm 3: Counterfactual Forward (Intervention and Debiasing)

**Input**:
Labels $e_L \in \mathbb{R}^K$, environment $e_{\mathrm{C}} \in \mathbb{R}^Q$, common confounders $e_{E_m} \in \mathbb{R}^M$, initial latents $Z_e \in \mathbb{R}^{L_{max} \times d_z}$, causal matrix $A \in \mathbb{R}^{K \times K}$. Hyperparameters about causal loss $\gamma, \delta$.
**Output**:
Counterfactual loss $\mathcal{L}_{\text{causal}}$; intervened predictions $\hat{e}_{Y,\text{int}_j}$.

**Procedure**:
1. **Factual Forward**:
   1.1 Compute $Z_{\mathrm{L}}$ via causal layer on $Z_{e,\mathrm{L}}$.
   1.2 Concatenate: $Z_d = [Z_{\mathrm{L}}, Z_{\text{extra}}]$.
   1.3 Decode $Z$: $\hat{e}_Y \in \mathbb{R}^P$.
2. **Interventions**:
   Intervene on labels set $j$ in $[1, K]$,
   2.1 Construct $e_{U_j}$: Set sub-dimensions to baseline $\mathrm{do}(U_j = 0)$
   2.2 Conditional prior: $\mu_{L_j,E,\text{conf}}, \log\sigma^2_{L_j,E,\text{conf}} = \mathrm{MLP}([e_{L_j}, e_{\mathrm{C}}, e_{E_m}])$
   2.3 Sample $Z_{\mathrm{L},\text{int}j} \sim \mathcal{N}(\mu_{L_j,E,\text{conf}}, \sigma^2_{L_j,E,\text{conf}})$ for 10 times.
3. **Causal Propagation**:
   3.1 Apply causal layer: $Z_{\text{L, intj}} = f\left(A \cdot Z_{L,int_j}, \epsilon\right)$.
   3.2 Concatenate: $Z_{int_j} = [Z_{L,int_j}, Z_{extra}]$.
4. **Decode Intervened Outputs**:
   Decode $Z_{int_j}$: $\hat{e}_{Y,int_j} \in \mathbb{R}^P$ for 10 times and take mean.
5. **Compute Counterfactual Losses**:
   5.1 $Z \rightarrow Y$:
   $\mathcal{L}_{\text{causal},Y} = \mathrm{MSE}(\hat{e}_Y, e_Y) - \gamma \mathrm{MSE}(\hat{e}_{Y,int_j}, e_Y)$
   5.2 $Z \rightarrow Z$:
   $\mathcal{L}_{\text{causal},A} = \mathrm{CE}(g(Z_{\mathrm{L}}), e_L) - \delta \mathrm{CE}(g(Z_{\mathrm{L},int_j}), e_L)$.
   5.3 Total: $\mathcal{L}_{\text{causal}} = \mathcal{L}_{\text{causal,Y}} + \mathcal{L}_{\text{causal,A}}$.
6. **Return**:
   Return $\mathcal{L}_{\text{causal}}$, $\hat{e}_{Y,int_j}$.

Algorithm 4: Optimal Label Intervention Search for Maximum Interaction

**Input**:
$L$: $\{0,1\}^{72}$ matrix of binary labels
data: Model input data, including Marketing text $X$, environment variables $C \in \mathbb{R}^{Q}$, common confounders $e_{E_m} \in \mathbb{R}^{M}$, trained D-VAE model.
max_flips: Maximum number of labels to intervene (default: 5).
inference.causal_intervention: Function to compute counterfactual interaction predictions $\hat{e}_{Y,int} \in \mathbb{R}^{N}$.

**Output**:
$\hat{L}$: $1 \times 72$ matrix of optimal flipped labels.
$S$: Set of flipped label indices ($|S| \leq$ max_flips).
best_score: Maximum predicted interaction sum.

**Procedure**:

1. **Initialization**:
   1.1 Initialize empty flip set: $S = \emptyset$.
   1.2 Set current labels: $L_{\text{current}} = L$, baseline interaction: $\hat{e}_{Y,\text{base}} = e_Y$ ,Set initial score: best_score $= \sum_{p=1}^{P} \hat{e}_{Y,\text{base}}[p]$.
   1.3 Initialize optimal outputs: best_S $= \emptyset$, $\widehat{U} = U$.
2. **Greedy Search for Optimal Flips**:
   For $t = 1$ to max_flips:
   2.1 Initialize temporary maximum score: max_score = best_score.
   2.2 Initialize best candidate tag: $j^* =$ None.
   2.3 For each tag $j \in \{1,2,\dots,72\} \backslash S$:
   a. Construct intervention set: $S' = S \cup \{j\}$.
   b. Compute counterfactual interaction:
   $\hat{e}_{Y,\text{int}}$ = D-VAE.inference.causal_intervention($X, C, e_{E_m}$,intervention $= S'$)
   c. Calculate score: score$_j = \sum_{p=1}^{P} \hat{e}_{Y,\text{int}}[p]$.
   d. If score$_j >$ max_score: Update max_score = score$_j$, Set $j^* = j$.
   2.4 If $j^* =$ None: Break (no further improvement possible).
   2.5 Update flip set: $S = S \cup \{j^*\}$.
   2.6 Update current labels:
   $L_{\text{current}}[:, j^*] = 1 - L_{\text{current}}[:, j^*]$
   2.7 Update optimal outputs: best_S $= S$, $\hat{L} = L_{\text{current}}$, best_score = max_score.
3. **Return**:
   Return $\hat{L}$, best_S, best_score.

## Online Appendix D: Supplementary Charts

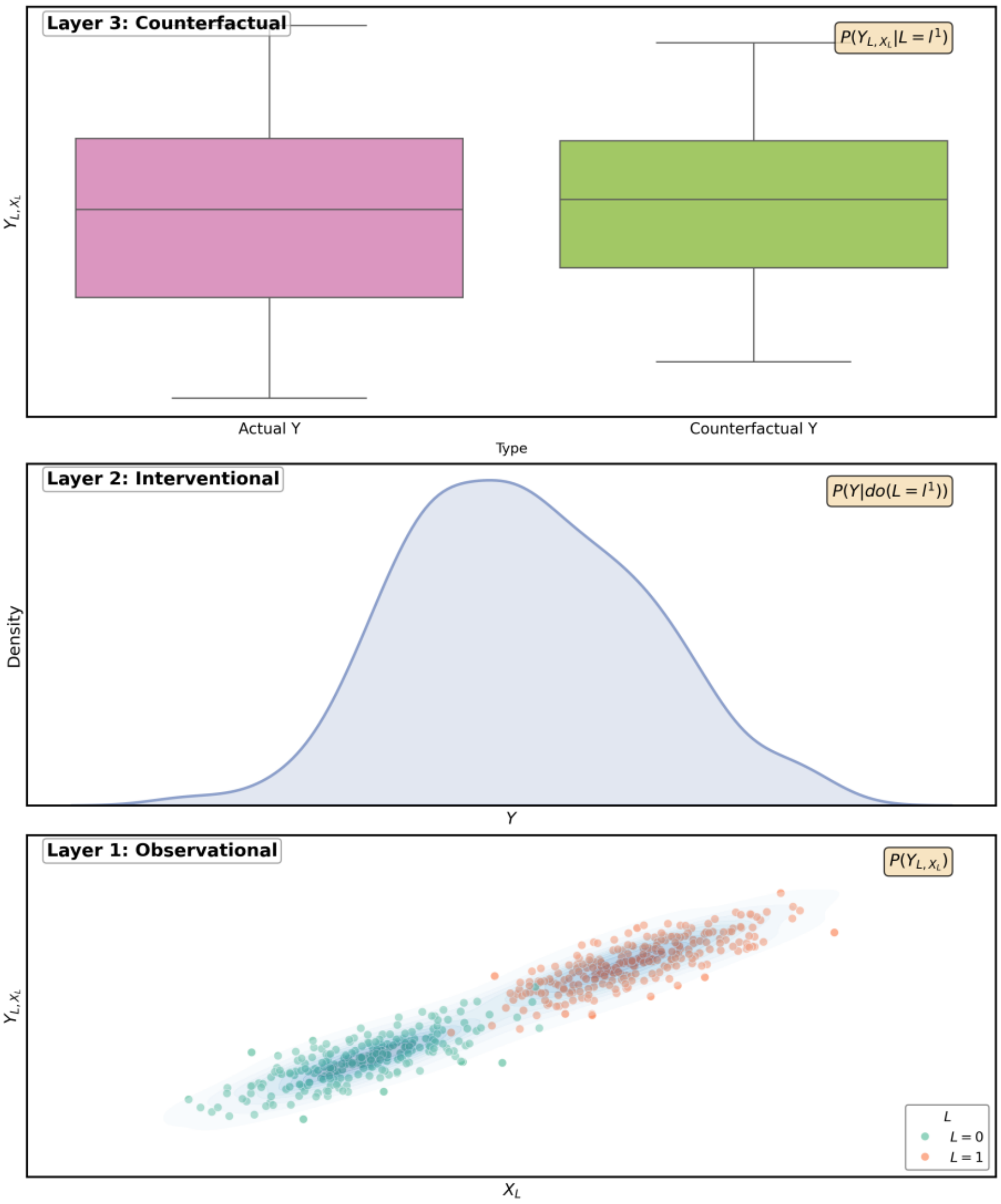

**Figure 1: Pearl's Causal Ladder: Data Distributions Across Three Levels of Causality**

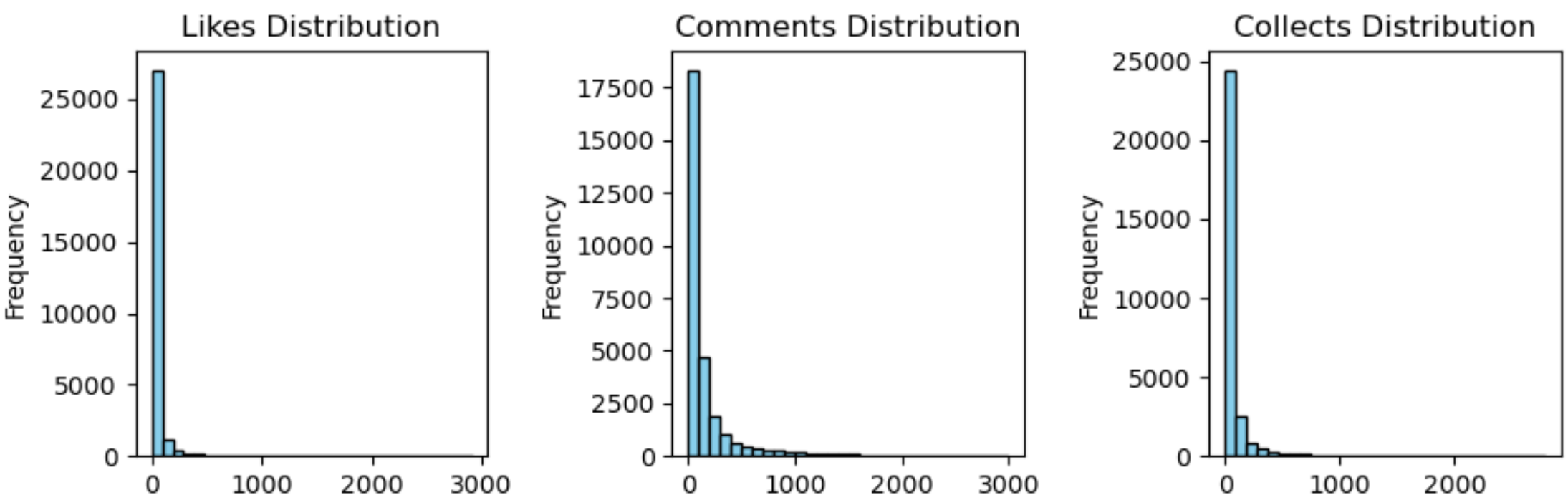

**Figure 2: User Engagement Distribution in Our Dataset**

**Online Appendix E: Additional Information on the Field Experiment**

1. Manipulation Check

1.1 Prompt Template for LLM-based Manipulation Check

> You are an expert marketing analyst. I will provide you with:
>
> A target label change: one or more labels added and/or removed;
>
> The original ad copy (before intervention);
>
> The modified ad copy (after intervention).
>
> Your task is to evaluate whether the modified copy faithfully reflects the intended label change by:
>
> Emphasizing content consistent with the added label(s), and
>
> Removing or de-emphasizing content associated with the removed label(s).
>
> For each label in the intervention, assign a score from 1 to 10, where:
>
> 1–3: The copy contradicts or ignores the label;4–6: The copy weakly or ambiguously aligns with the label;7–10: The copy clearly and coherently embodies the label's intent.

1.2 Manipulation Check Score

2. Demographic Distributions of Respondents

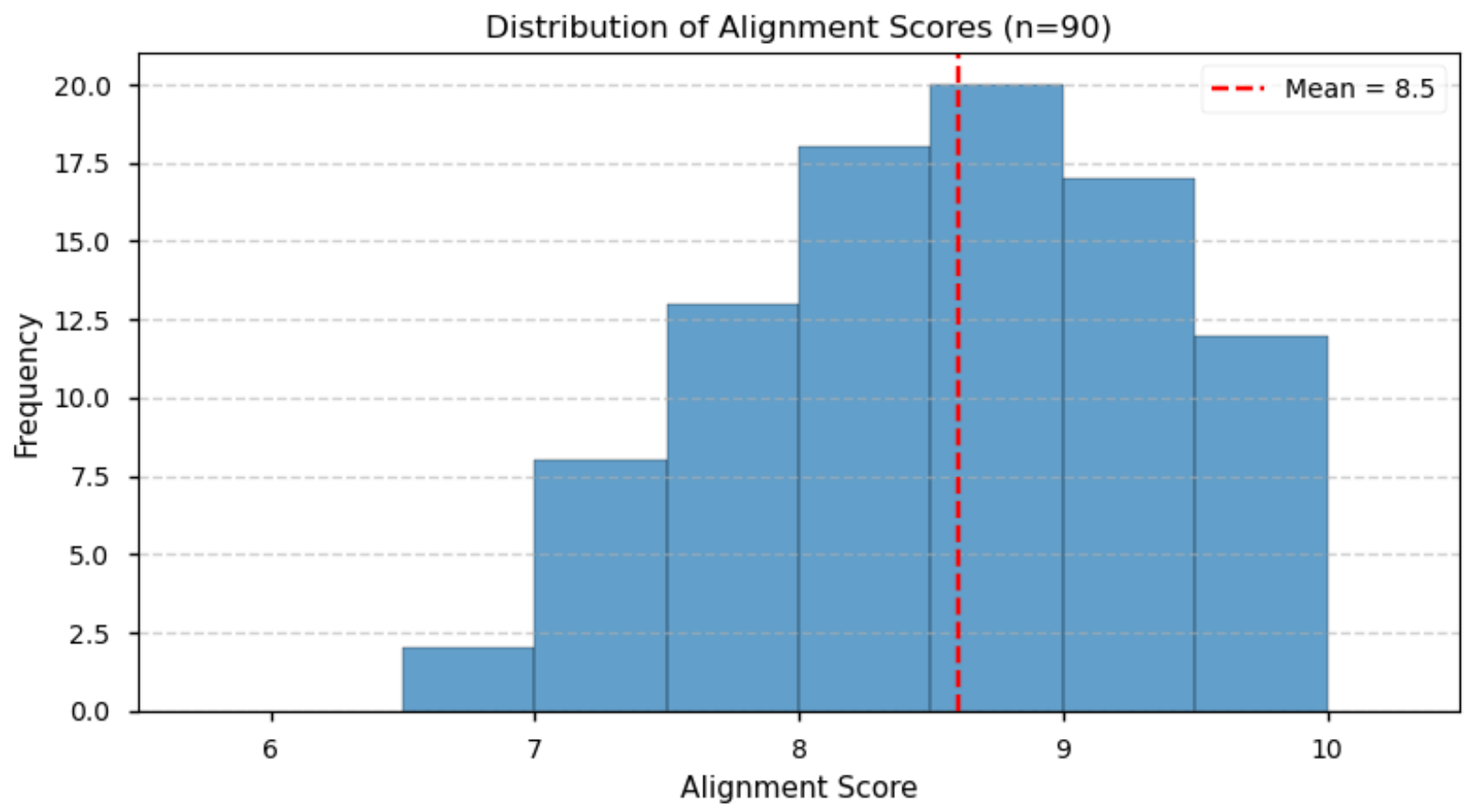

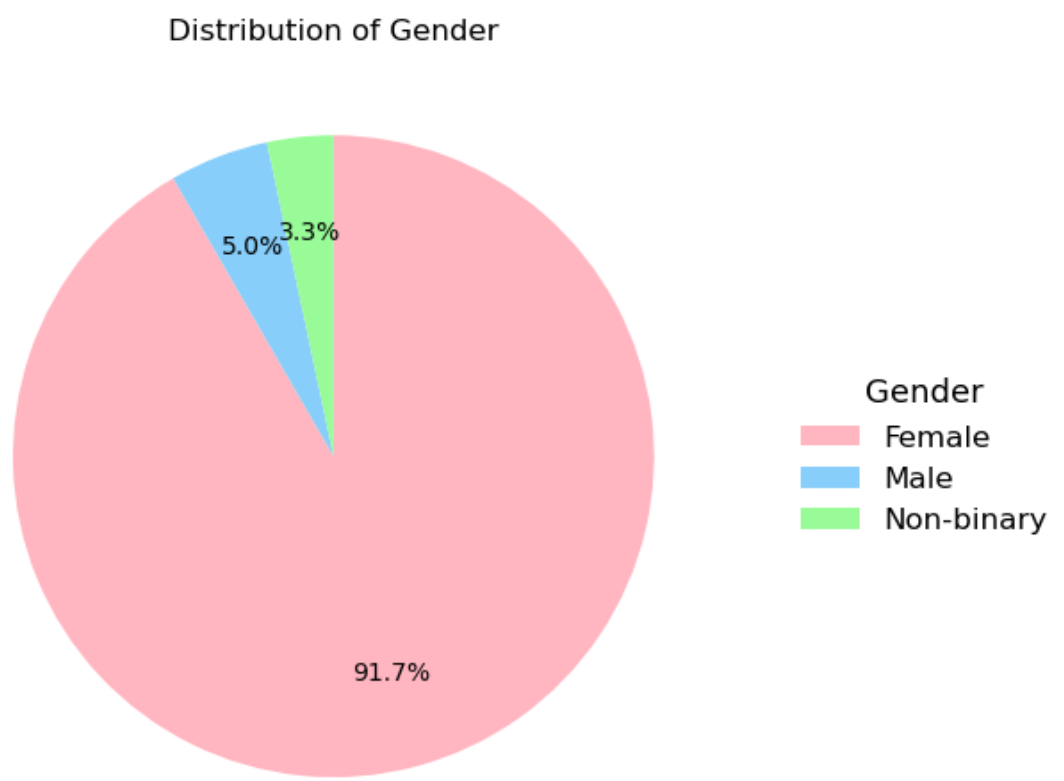

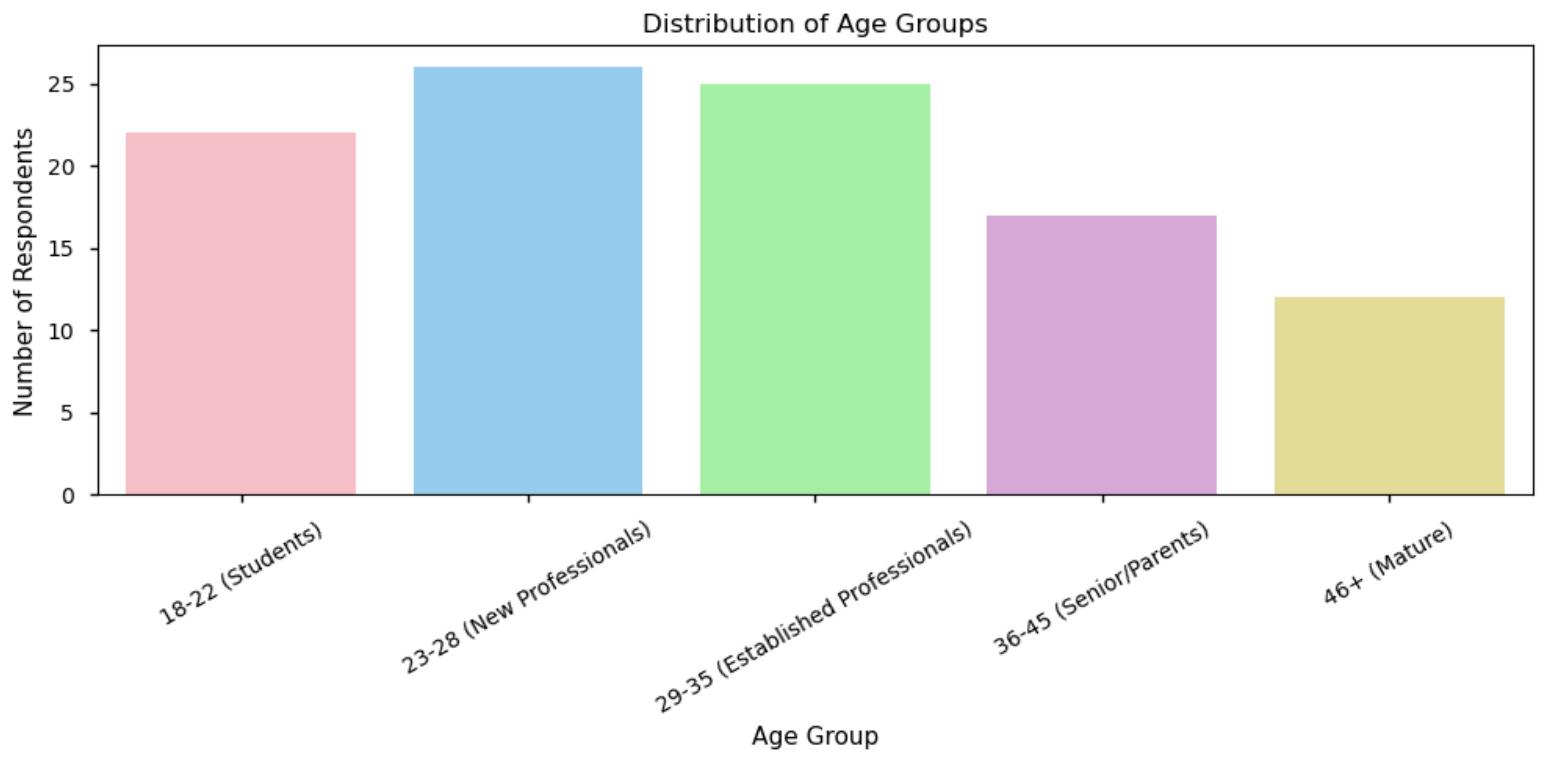
Distribution of Age Groups
Number of Respondents
0
5
10
15
20
25
18-22 (Students)
23-28 (New Professionals)
29-35 (Established Professionals)
36-45 (Senior/Parents)
46+ (Mature)
Age Group

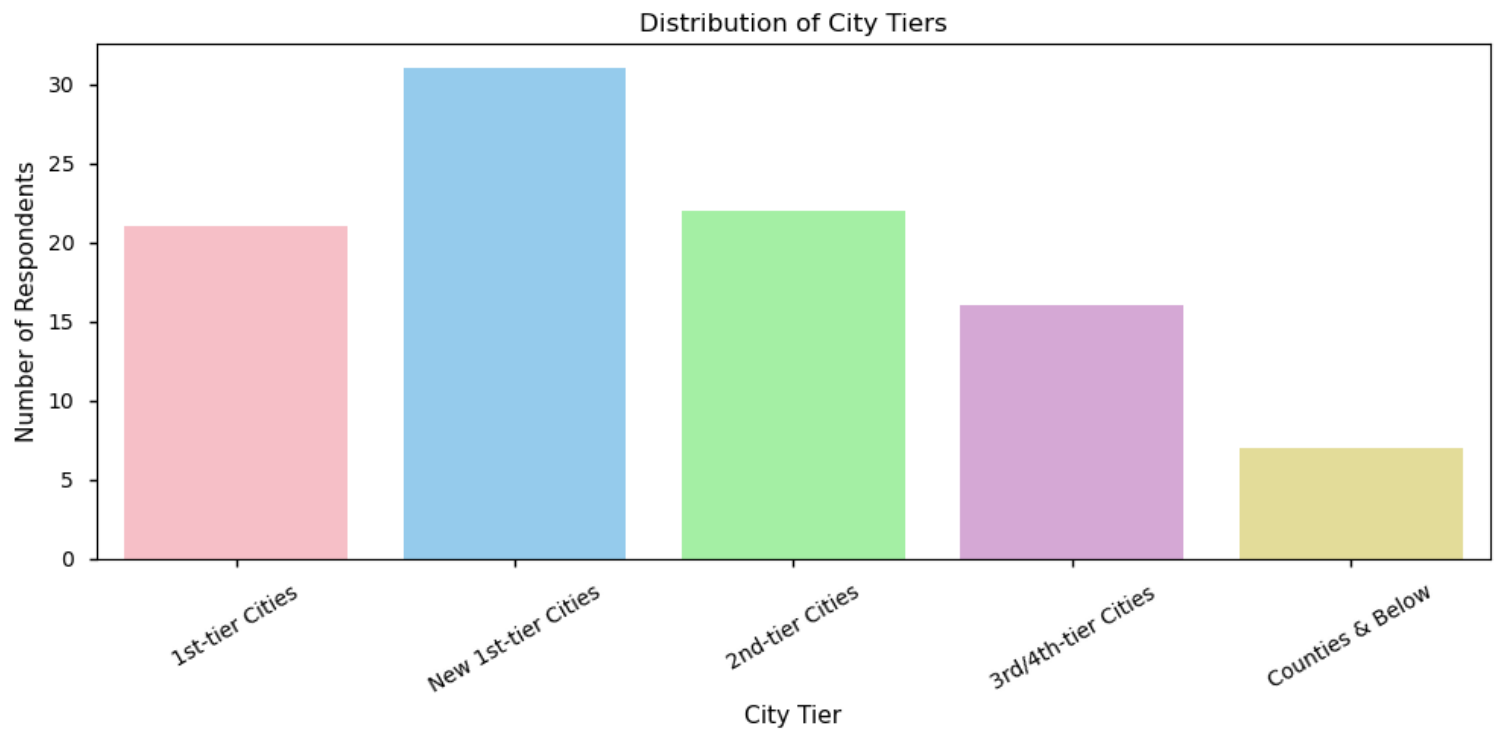
Distribution of City Tiers
Number of Respondents
0
5
10
15
20
25
30
1st-tier Cities
New 1st-tier Cities
2nd-tier Cities
3rd/4th-tier Cities
Counties & Below
City Tier

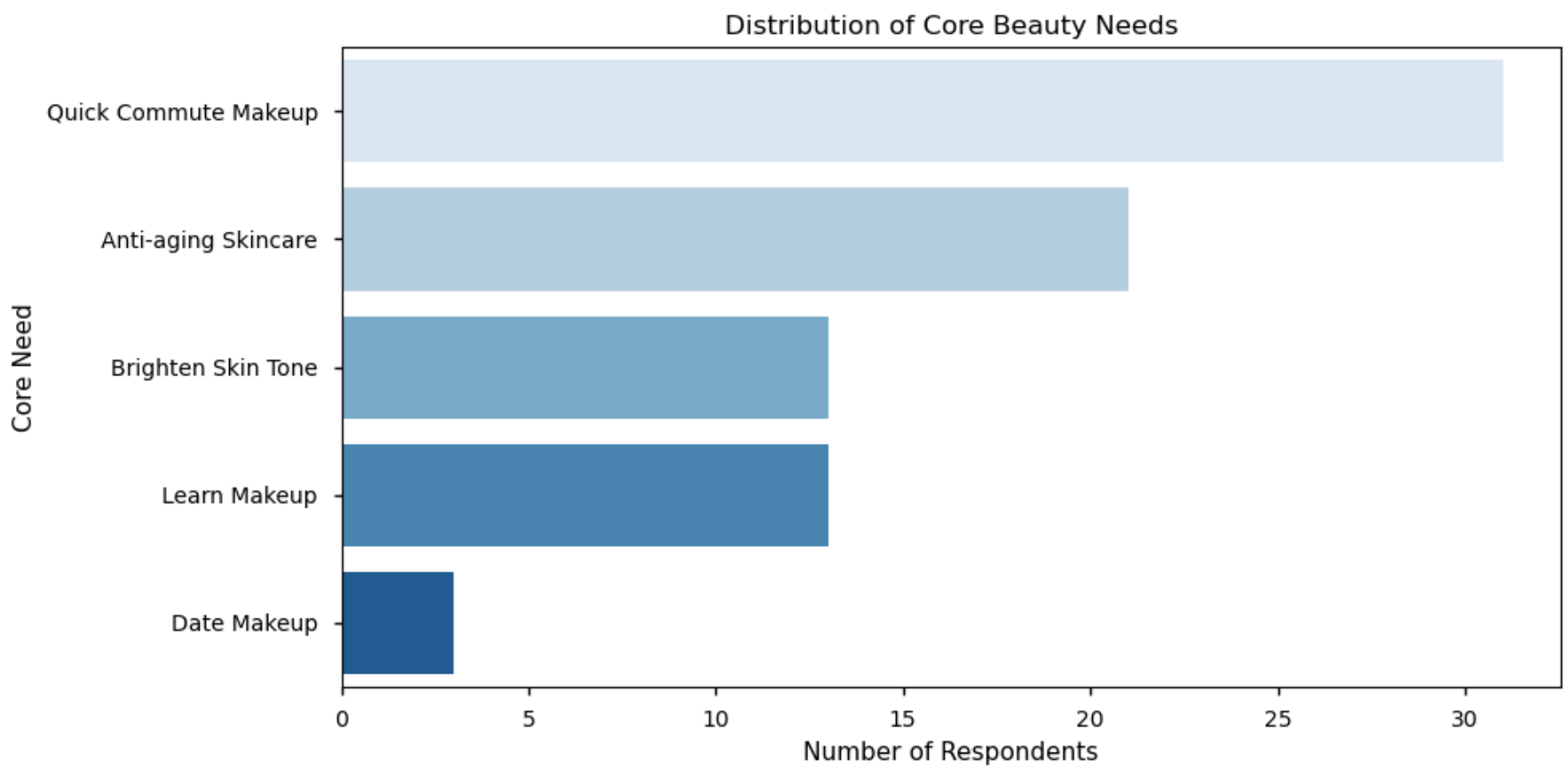
Distribution of Core Beauty Needs
Core Need
Quick Commute Makeup
Anti-aging Skincare
Brighten Skin Tone
Learn Makeup
Date Makeup
0
5
10
15
20
25
30
Number of Respondents

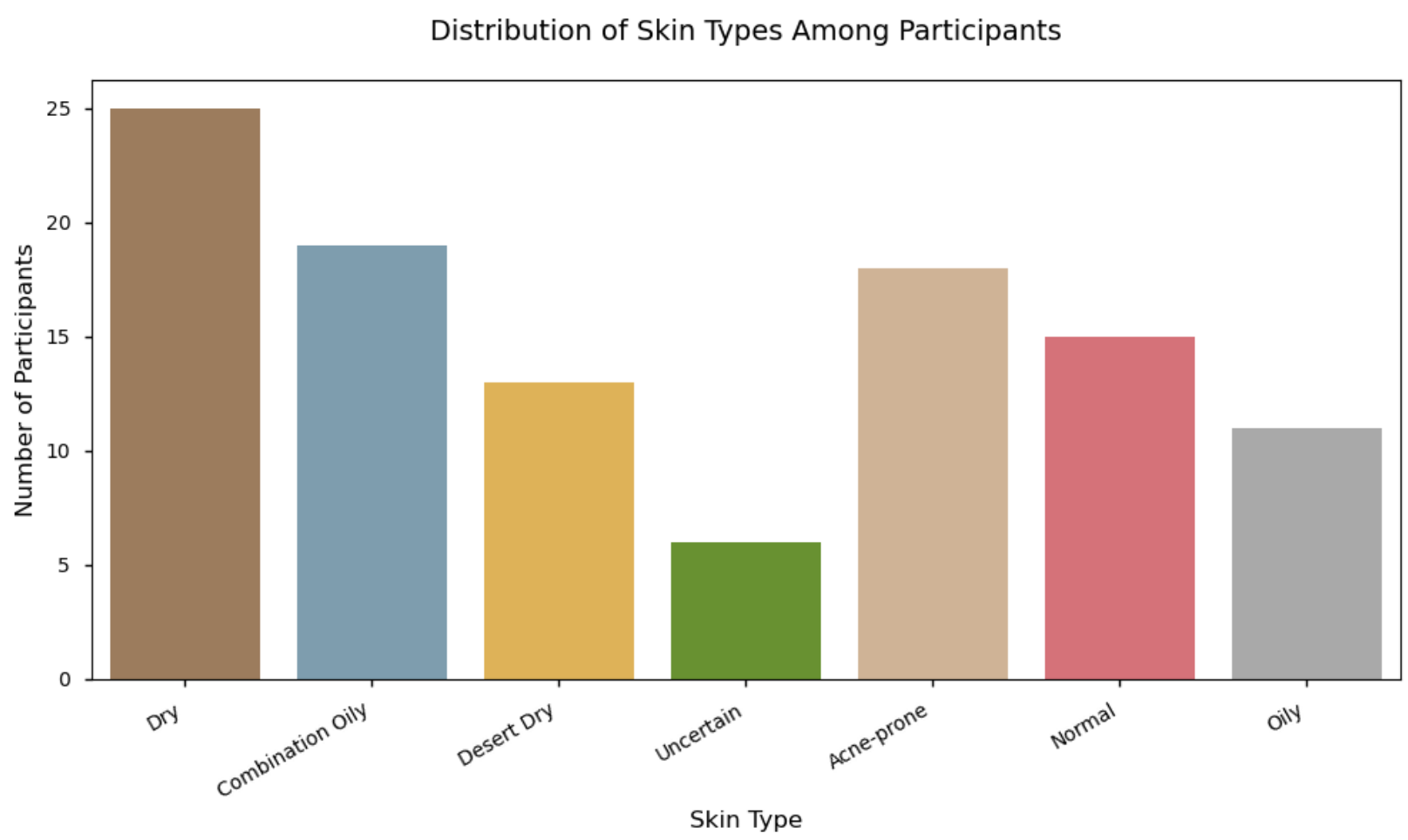
Distribution of Skin Types Among Participants
Number of Participants
0
5
10
15
20
25
Dry
Combination Oily
Desert Dry
Uncertain
Acne-prone
Normal
Oily
Skin Type